\documentclass[11pt]{article}

\usepackage[utf8]{inputenc}
\usepackage[T1]{fontenc}
\usepackage[margin=1in, bottom=0.3in]{geometry}
\usepackage{hyperref}
\usepackage{url}
\usepackage{booktabs}
\usepackage{amsfonts}
\usepackage{amsmath}
\usepackage{amssymb}
\usepackage{nicefrac}
\usepackage{graphicx}
\usepackage[numbers]{natbib}
\usepackage{fancyhdr}
\usepackage{xcolor}
\usepackage[figuresonly,nomarkers,nolists]{endfloat}
\usepackage{enumitem}
\usepackage{titlesec}
\titlespacing*{\section}{0pt}{2.3ex plus 0.7ex minus 0.2ex}{2.3ex plus 0.2ex}
\titlespacing*{\subsection}{0pt}{2.2ex plus 0.7ex minus 0.2ex}{1.0ex plus 0.2ex}
\titlespacing*{\subsubsection}{0pt}{2.2ex plus 0.7ex minus 0.2ex}{1.0ex plus 0.2ex}
\titleformat{\paragraph}[runin]{\normalfont\normalsize\bfseries}{}{0pt}{}
\titlespacing*{\paragraph}{0pt}{2.2ex plus 0.7ex minus 0.2ex}{0.5em}

\usepackage{amsmath,amsfonts,bm}

\def\1{\bm{1}}

\DeclareMathAlphabet{\mathsfit}{\encodingdefault}{\sfdefault}{m}{sl}
\SetMathAlphabet{\mathsfit}{bold}{\encodingdefault}{\sfdefault}{bx}{n}

\newcommand{\Wseed}{W_{\text{seed}}}

\usepackage{todonotes}
\usepackage{xstring}
\definecolor{bordeaux}{HTML}{800020}
\makeatletter
\DeclareRobustCommand{\pdfmarkupcomment}[3][]{%
  #2%
  \begingroup
    \def\pmc@name{yellow}%
    \IfSubStr{#1}{color=green}{\def\pmc@name{green}}{}%
    \edef\pmc@bg{\pmc@name!30}%
    \def\pmc@fg{bordeaux}%
    \@ifundefined{@captype}{%
      \todo[inline, backgroundcolor=\pmc@bg, textcolor=\pmc@fg]{#3}%
    }{%
      {\textcolor{bordeaux}{[EDIT: #3]}}%
    }%
  \endgroup
}
\makeatother

\graphicspath{{figures/}}

\makeatletter
\renewcommand{\@seccntformat}[1]{\csname the#1\endcsname\hspace{0.4em}}
\makeatother

\title{A Little Rank Goes a Long Way: Random Scaffolds with LoRA Adapters Are All You Need}

\author{
  Hananel Hazan$^{1,*}$ \quad Yanbo Zhang$^{1}$ \quad Benedikt Hartl$^{1,2}$ \quad Michael Levin$^{1,3}$ \\[6pt]
  $^{1}$Allen Discovery Center at Tufts University, Medford, Massachusetts, USA \\
  $^{2}$Institute for Theoretical Physics, TU Wien, Wien, Austria \\
  $^{3}$Wyss Institute for Biologically Inspired Engineering, \\
  Harvard University, Boston, Massachusetts, USA
}

\date{}

\begin{document}
\twocolumn[%
  \vspace*{-1.5em}
  \maketitle
  \thispagestyle{empty}
  \vspace{-3.5em}

  \begin{abstract}
  How many of a neural network's parameters actually encode
  task-specific information?
  We investigate this question with \textbf{LottaLoRA}$^{\ddagger}$,
  a training paradigm in which every backbone
  weight is drawn at random and frozen; only low-rank LoRA adapters are
  trained. Across nine benchmarks spanning diverse architecture
  families---from single-layer classifiers to 900\,M-parameter
  Transformers---low-rank adapters over frozen random backbones recover
  96--100\% of fully trained performance while training only 0.5--40\% of
  the parameters. The task-specific signal therefore occupies a subspace
  orders of magnitude smaller than the full parameter count suggests.
  Three mechanistic findings underpin this result:
  (1)~the frozen backbone is actively exploited when static---the learned
  scaling~$\beta$ remains strictly positive across all architectures---but
  when the scaffold is destabilized, the optimizer silences it and the LoRA
  factors absorb all task information;
  (2)~the frozen backbone is preferable but interchangeable---any random
  initialization works equally well, provided it remains fixed
  throughout training; and
  (3)~the minimum LoRA rank at which performance saturates estimates the
  intrinsic dimensionality of the task, reminiscent of the number of
  components retained in Principal Component Analysis (PCA).
  The construction is formally analogous to Reservoir Computing
  unfolded along the depth axis of a feedforward network.
  Because the backbone is determined by a random seed alone, models can be
  distributed as adapters plus seed---a footprint that grows with task
  complexity, not model size, so that storage and memory savings compound
  as architectures scale.
  \end{abstract}

  \medskip
  \noindent\textbf{Keywords:}
  Reservoir Computing,
  Low-Rank Adaptation,
  Large Language Models,
  Parameter-Efficient Training,
  Random Networks,
  Hyperdimensional Computing

  \vspace{25pt}
]%
{\renewcommand{\thefootnote}{\fnsymbol{footnote}}\footnotetext[1]{Corresponding author: \href{mailto:hananel@hazan.org.il}{hananel@hazan.org.il}}}
{\renewcommand{\thefootnote}{\fnsymbol{footnote}}\footnotetext[3]{A portmanteau of ``LoRA'' and ``a lotta,'' alluding to the Lottery Ticket Hypothesis~\cite{frankle2019lottery}.}}
\setlength{\dbltextfloatsep}{6pt}

\section{Introduction}
\label{sec:intro}

As neural networks grow larger, so does the cost of training them.
Low-Rank Adaptation (LoRA)~\cite{hu2022lora} has become the dominant method
for reducing this cost at fine-tuning time---freeze a pre-trained backbone, inject small
trainable matrices, and 
fine-tune the network on a new task.
The frozen
weights in that setting encode rich semantic knowledge accumulated over
pre-training; the low-rank adapters merely 
redirect the network's behavior toward the target task.

LoRA is more expressive than this framing suggests: a product of two
rank-$r$ matrices can, given sufficient rank, represent any weight matrix.
The adapters can therefore encode the full connectivity of
every layer from scratch.
This observation connects to intrinsic dimensionality findings~\cite{aghajanyan2021intrinsic} and the Lottery Ticket Hypothesis~\cite{frankle2019lottery}: task-relevant structure may occupy a surprisingly small subspace of the full weight space.
The bottleneck dimension~$r$ of the LoRA factorization 
(referred to as the \emph{rank} throughout this paper, since it upper-bounds the matrix rank of the learned update~$BA$) 
is the only structural degree of freedom, reducing the number of trainable variables per layer by orders
of magnitude.

Independently, the Reservoir Computing literature has long established that randomly
initialized, fixed dynamical systems can serve as powerful feature extractors---provided
only that a trained readout channels their dynamics toward the
task~\cite{jaeger2001echo,maass2002real,hazan2012topological}. This raises a natural question: if a frozen random
network already computes, can LoRA serve as a low-dimensional controller that steers
the random projections of this substrate toward the task at each layer?
\emph{What if the backbone weights are never pre-trained at all---what if they are simply random?}
A related principle has been demonstrated in spiking neural networks, where
fixed random weights paired with trained synaptic delays suffice for
classification~\cite{hazan2022memory}, suggesting that the learned component
of a network can be far smaller than conventionally assumed.

Here, we show that they can be: across nine benchmarks, low-rank adapters over frozen random backbones recover 96--100\% of fully trained performance under matched architectures and training protocols, while training only 0.5--40\% of the parameters. This recovery metric measures proximity to the fully trained ceiling, not absolute efficiency; Section~\ref{sec:cost} analyzes the full compute picture. The results suggest that pre-training, while beneficial, is not strictly necessary.

The implication extends beyond parameter efficiency. If a frozen random
backbone suffices, then the vast majority of a network's parameters are
\emph{scaffolding}---structurally necessary but carrying no learned
information. The task-specific signal concentrates in a low-dimensional
subspace whose size reflects the problem, not the architecture. This
reframes the relationship between model size and capability: the number of
parameters measures the capacity of the scaffold, while the LoRA rank
measures the complexity of the task itself.

\begin{figure}[t]
  \centering
  \includegraphics[width=\linewidth]{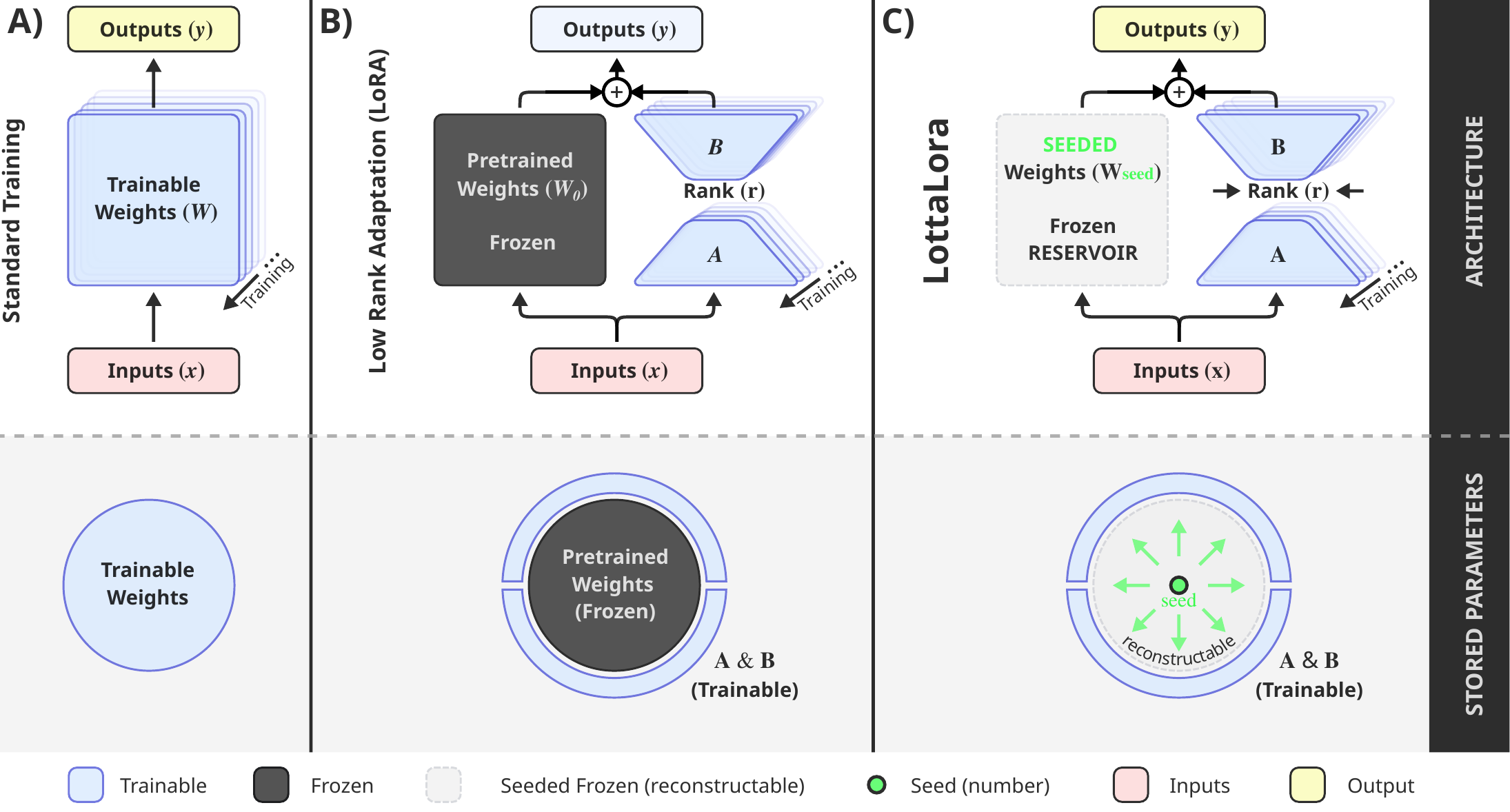}
  \caption{\textbf{LottaLoRA replaces pre-trained weights with seeded reservoirs.}
  Three parameterization strategies for a single network layer.
  \textbf{(a)}~A conventional dense layer stores and trains all $m \times n$
  weights.
  \textbf{(b)}~Low-Rank Adaptation (LoRA) freezes a pre-trained weight matrix
  $W_0$ and learns only two small factors $A \in \mathbb{R}^{m \times r}$ and
  $B \in \mathbb{R}^{r \times n}$; the stored parameters are $W_0$ plus the
  adapters.
  \textbf{(c)}~LottaLoRA (this work) replaces $W_0$ with a random matrix
  $\Wseed$ generated from a fixed seed, so the backbone is
  never pre-trained and need not be stored: only the seed and the
  low-rank factors $A$, $B$ are saved.  The distributable footprint
  therefore shrinks from the full weight matrix to a single integer
  plus two rank-$r$ factors.}
  \label{fig:model}
\end{figure}

Figure~\ref{fig:model} illustrates the three strategies.
This construction is formally analogous to Reservoir Computing~\cite{jaeger2001echo,maass2002real}, though it departs from the classical framework in key respects developed in Section~\ref{sec:rc}. In our formulation the
reservoir unfolds along the depth axis in feedforward architectures and along
the temporal axis in recurrent ones, and the LoRA adapter replaces the linear
readout with a low-dimensional feedback controller.
To demonstrate the power of this approach, we investigate a progression of
experiments---from a single linear layer on MNIST through reservoir RNNs,
graph networks, CNNs, Decision Transformers, Vision Transformers, and
NLP sentiment classification with DistilBERT, up to
900\,M-parameter Transformers on WikiText-103. Across all tested
architectures, we find that LottaLoRA is competitive with full-weight
training, given sufficient adapter rank, while training only a small fraction
of the parameters.
A key mechanistic finding is that the role of the scaffold depends on its
stability: when static, the optimizer retains the backbone contribution
($\beta > 0$, with magnitude varying by architecture); when the scaffold is resampled during training, the
optimizer drives $\beta$ toward zero, pushing all task information into the
LoRA factors alone. The scaffold's specific values are interchangeable, but
its fixedness is essential.

These findings support a central hypothesis: \emph{the minimum LoRA rank at
which the adapter matches fully trained performance provides an estimate of
the task's intrinsic dimensionality}---reminiscent of the number of principal
components retained in PCA.
Because the frozen backbone is fully determined by a random seed, the
only artifact that must be saved, transmitted, or updated is the compact
LoRA factors.  At the 900\,M-parameter scale this yields a $21\times$
reduction in distributable size versus fp16, $6\times$ versus 4-bit
quantization, and a 36\% reduction in peak GPU memory---savings that
grow monotonically with model size (Section~\ref{sec:cost}).

\section{Related Work}
\label{sec:related}

\paragraph{Reservoir Computing.}
The idea of exploiting a randomly initialized, fixed dynamical system as a
feature extractor originates in the Reservoir Computing (RC) literature. Echo
State Networks~\cite{jaeger2001echo} and Liquid State
Machines~\cite{maass2002real} both demonstrate that a large,
randomly connected recurrent network can serve as a universal temporal feature
extractor provided that only its linear readout is trained. The key condition
is that the reservoir dynamics satisfy the Echo State Property (ESP)---the requirement that the effect of initial conditions vanishes over time---typically enforced by constraining the spectral radius of the recurrent weight matrix to be less than one. LottaLoRA builds upon
this philosophy. In feedforward architectures, we re-interpret the reservoir as unfolding along the depth
dimension rather than along time, replacing temporal recurrence
with layer-to-layer propagation. In our recurrent benchmarks (Section~\ref{sec:broader}), the reservoir operates in the classical temporal sense.

\paragraph{Frozen and partially frozen Transformers.}
Prior work~\cite{shen2021reservoir} has shown that interspersing randomly
initialized, permanently frozen layers with fully trainable Transformer~\cite{vaswani2017attention}
layers maintains or even improves performance on machine translation and
masked language modeling, while accelerating wall-clock convergence. A
complementary line of work~\cite{zhong2024algorithmic} demonstrates that a
randomly initialized decoder-only Transformer, with only embedding layers
optimized, can already perform a variety of algorithmic tasks. LottaLoRA extends both directions to their extreme: the entire backbone
is frozen, and the trainable components can be restricted to low-rank adapters alone.

\paragraph{Parameter-efficient adaptation.}
LoRA~\cite{hu2022lora} proposes to fine-tune a pre-trained model by freezing
its weights and
adding trainable low-rank corrections to each linear layer. In the
standard setting, the frozen weights encode rich semantic knowledge; the
LoRA factors steer that knowledge toward a new objective. LottaLoRA asks what
happens when the frozen weights encode no prior knowledge at all. The
adapters must therefore serve a different function: rather than steering
pre-trained representations, they provide a per-layer correction path that stabilizes the random projections
of the frozen backbone and channels its signals toward the downstream task,
as formalized in Section~\ref{sec:rc}.
A parallel direction~\cite{sarkar2024eggroll} applies rank-$r$ low-rank structure to the perturbations of gradient-free evolution strategies, showing that the same rank bottleneck that enables parameter-efficient fine-tuning also enables efficient gradient estimation at hyperscale.

\paragraph{Intrinsic dimensionality.}
The hypothesis that neural network optimization operates in a low-dimensional
subspace has been explored in several settings. Prior
work~\cite{aghajanyan2021intrinsic,li2018measuring} has shown that
fine-tuning large models can be effective even when restricted to a
random low-dimensional subspace of the full parameter space. LottaLoRA pushes
this observation further: by training \emph{from scratch} rather than
fine-tuning, we test whether the intrinsic dimensionality of a task can be
captured by a low-rank adapter over a random backbone that carries no prior
knowledge whatsoever.
Concurrently, EGGROLL~\cite{sarkar2024eggroll} demonstrates that structuring evolution-strategy perturbations as rank-$r$ matrices is sufficient to train billion-parameter language models, providing independent evidence that effective updates are confined to a low-dimensional subspace even when gradient information is unavailable.

\paragraph{Hyperdimensional Computing.}
A complementary perspective comes from Hyperdimensional Computing
(HDC)~\cite{kanerva2009hyperdimensional,kleyko2023survey}.
HDC represents data as high-dimensional random vectors (typically thousands of
dimensions) and manipulates them with simple elementwise operations such as
binding and bundling. Because near-orthogonality is the norm in high
dimensions, these representations are inherently robust to noise and
hardware-level quantization.
While Reservoir Computing also exploits fixed random projections, it
fundamentally relies on recurrent dynamics to create a temporal state space.
LottaLoRA's frozen backbone is a purely feedforward structure, making
HDC---which operates on static high-dimensional random
representations---the more precise analogy.
LottaLoRA's frozen scaffold plays an analogous role, and the
empirical finding that sign-binarized backbones (each weight replaced by its sign, $\pm1$) achieve comparable task performance to Gaussian initialization
(Section~\ref{sec:q2}) echoes HDC's core noise-tolerance property.
In both cases, high-dimensional random representations concentrate around
near-orthogonality, so that even aggressive per-element quantization (here,
sign binarization) preserves the geometric structure the downstream learner
relies on. The
seed-based reconstruction of the backbone further parallels HDC's approach, in
which the high-dimensional random space is specified rather than learned.

\paragraph{Converging evidence.}
These converging lines of evidence---from RC, frozen Transformers, parameter-efficient adaptation, intrinsic dimensionality, and Hyperdimensional Computing---suggest that the degrees of freedom needed to solve a task may occupy a far smaller subspace than the full parameter space. We now formalize this intuition by describing LottaLoRA, a training paradigm in which the frozen random backbone provides a high-dimensional substrate and the low-rank subspace specifies how to channel its representational capacity toward the task.

\section{Method}
\label{sec:method}

Let $f_\theta$ denote a neural network with $n$ layers and hidden
dimension~$d$, where $\theta$ comprises all parameters.
In standard LoRA~\cite{hu2022lora}, the frozen weight matrix $W_0$ encodes
knowledge from pre-training; in LottaLoRA, we replace it with $\Wseed$ drawn
from a fixed random distribution (e.g., a zero-mean Gaussian with variance
matched to the standard initialization heuristic for the given depth and
width) and never updated.
The parameter vector~$\theta$ is partitioned into fixed random weights
$\{\Wseed^{(i)}\}_{i=1}^{n_\mathrm{layers}}$ and a trainable subset~$\Theta$
(defined below).
We refer to the frozen backbone~$\Wseed$ as the \emph{scaffold}, since it
provides fixed structural support---a high-dimensional substrate for the adapter---but
carries no learned, task-specific information.

For each linear layer with fixed random weight matrix
$\Wseed \in \mathbb{R}^{d_\mathrm{out} \times d_\mathrm{in}}$, we introduce a
LoRA adapter~$BA$ where $B \in \mathbb{R}^{d_\mathrm{out} \times r}$ and
$A \in \mathbb{R}^{r \times d_\mathrm{in}}$, together with a trainable
scalar $\beta \in \mathbb{R}$ and a fixed scaling hyperparameter $\alpha > 0$.
The effective forward computation at each such layer
(Figure~\ref{fig:layer_diagram}c) is
\begin{equation}
  \label{eq:randlora}
  h_\mathrm{out} = \beta\, \Wseed\, h_\mathrm{in} \;+\; \frac{\alpha}{r}\,BA\, h_\mathrm{in}\,,
\end{equation}
where $h_\mathrm{in}$ and $h_\mathrm{out}$ are the input and output
activations, respectively.
The scaling factor $\alpha/r$ follows the standard LoRA
convention~\cite{hu2022lora}: $\alpha$ is a fixed hyperparameter that controls
the relative magnitude of the adapter path (set to $\alpha{=}1$ for MNIST and
$\alpha{=}16$ for Transformer experiments; see
Appendix~\ref{app:engineering}).
At fixed~$\alpha$ with increasing~$r$, the per-rank contribution is
attenuated, stabilizing training at high ranks.
Although we present Equation~\eqref{eq:randlora} for fully connected layers,
the decomposition applies to any layer whose forward pass is governed by a
weight matrix---attention projections ($W_Q$, $W_K$, $W_V$, $W_O$),
convolutional kernels (reshaped to two dimensions), recurrent gates, and graph
message-passing operators---as demonstrated across nine benchmarks spanning
diverse architecture families in Section~\ref{sec:experiments}.

The full set of trainable parameters is
$\Theta = \{\beta_i, B_i, A_i\}_{i=1}^{n_\mathrm{layers}}$ together with
any task-specific head (e.g., token embeddings and language model head for
Transformers, or a classification layer for MNIST).
When LayerNorm is present, its parameters are also trainable and included
in~$\Theta$; not all experiments use LayerNorm (see
Appendix~\ref{app:engineering} for details).

The adapter operates as follows: $A$ projects the current state
into a low-dimensional signal and $B$ maps it back into the main pathway,
providing a learned correction to the frozen backbone at each layer. The
scalar~$\beta$ provides per-layer amplitude control over the backbone
contribution: it is initialized to~$1.0$ and left unconstrained.
Empirically, $\beta$ remains strictly positive across all tested
architectures, indicating that the optimizer never fully discards the
backbone contribution (see Appendix~\ref{app:engineering} for the
empirical $\beta$ distribution).

Because the backbone weights are never subject to gradient updates, they can
be initialized in any format convenient for the target hardware---sparse,
binary, ternary, or any other low-precision scheme---and the adapter will
learn to channel the resulting representations equally well
(Section~\ref{sec:q2}).

\paragraph{Seed-based reconstruction.}
Because~$\Wseed$ is fully determined by the random seed, architecture
specification, and initialization distribution, the backbone can be
reconstructed on any device without transmitting the weight matrices
themselves.
The distributable artifact of a LottaLoRA model is therefore:
(1)~the random seed~$s$;
(2)~the architecture configuration (layer count, hidden dimensions, layer
types);
(3)~the initialization specification (distribution family, variance
schedule);
(4)~the LoRA state dictionary
$\{A_i, B_i, \beta_i\}_{i=1}^{n_\mathrm{layers}}$; and
(5)~any task-specific head parameters (embeddings, classification layers).
Note that the head itself can also be parameterized as a frozen random
layer with LoRA adapters rather than a fully trainable layer, further
compressing the distributable artifact (see
Appendix~\ref{app:mnist_output} for ablations).
Reconstruction proceeds by initializing the framework's pseudorandom number
generator (PRNG) with~$s$ and drawing~$\Wseed$ in layer order using the specified
distribution.
The PRNG algorithm must be recorded alongside the seed, since different
frameworks (PyTorch, JAX, NumPy) use different generators; reproducibility
is guaranteed within a specified framework and version. All experiments
in this work use PyTorch~2.11.0 with CUDA~12.8.
Appendix~\ref{app:algorithm} provides framework-agnostic pseudocode for the
full procedure; Appendix~\ref{app:design_choices} summarizes which components
are required versus optional.

\section{Reservoir Computing Analogy: Where the Connection Holds and Diverges}
\label{sec:rc}

A classical Echo State Network (ESN)~\cite{jaeger2001echo} evolves a hidden
state over time as
\begin{equation}
  \label{eq:esn}
  h_{t+1} = \sigma\!\left(W_R\, h_t + U\, x_t\right),
\end{equation}
where $\sigma$ is a pointwise nonlinearity, $x_t$ is the external input, $W_R$ and~$U$ are fixed random matrices, and only the linear
readout~$C$ in $y_t = C h_t$ is trained.

In LottaLoRA, treating the layer index~$l$ as the propagation dimension, the
hidden state~$h_l$ propagates through the depth of the network as
\begin{equation}
  \label{eq:randlora_depth}
  h_{l+1} = \sigma\!\left(\beta\, \Wseed\, h_l + \tfrac{\alpha}{r}\,B\, z_l\right),
  \qquad z_l = A\, h_l \in \mathbb{R}^r.
\end{equation}
This is formally identical to the ESN update~\eqref{eq:esn} with the depth-wise propagation of Equation~\eqref{eq:randlora_depth} mapping to
$W_R = \beta \Wseed$, the injection matrix replaced by~$(\alpha/r)\,B$, and the
exogenous input~$x_t$ replaced by the endogenous low-dimensional
signal~$z_l = A\,h_l$ (Table~\ref{tab:rc_lora}).
LottaLoRA is RC unfolded along the spatial depth axis of a
network rather than along the time axis.

\begin{table*}[ht]
  \caption{LottaLoRA maps onto Reservoir Computing through a depth--time correspondence.}
  \label{tab:rc_lora}
  \centering
  \begin{tabular*}{\textwidth}{@{\extracolsep{\fill}}lll@{}}
    \toprule
    \textbf{Concept} & \textbf{RC (time)} & \textbf{LottaLoRA (depth)} \\
    \midrule
    Propagation axis & Time step $t$          & Layer index $l$ \\
    System state     & Hidden state $h_t$     & Layer activation $h_l$ \\
    Fixed dynamics   & Reservoir $W_R$        & Frozen backbone $\beta \Wseed$ \\
    Injection matrix & Input matrix $U$       & Scaled up-projection $(\alpha/r)\,B$ \\
    Drive signal     & External input $x_t$   & Projection $z_l = A\,h_l$ \\
    Trainable output & Readout $C$            & Head, embeddings, LoRA \\
    \bottomrule
  \end{tabular*}
\end{table*}

Classical RC theory makes three claims about systems with fixed random
dynamics and a trained readout:
(i)~the reservoir's specific weight values should not matter, only their
statistical properties;
(ii)~the reservoir must remain fixed during training, because a plastic
reservoir invalidates the readout's learned mapping;
(iii)~higher-dimensional reservoirs should yield better performance, as the
random projections span a richer feature space.
All three predictions are supported by our experiments in
Sections~\ref{sec:q2}--\ref{sec:q4}: 22 initialization families produce
statistically indistinguishable accuracy (prediction~i); resampling the
scaffold during training collapses performance significantly
(prediction~ii); and the loss gap narrows steadily with increasing backbone
size (prediction~iii). In all three cases, LottaLoRA behaves as RC theory
predicts, supporting the view that the RC framing accurately describes the
method's dynamics.

The analogy does have limits, however.
LottaLoRA departs from classical RC in two respects: the backbone~$\Wseed$ is
not constrained to satisfy the ESP, and the signal propagates
along the depth axis of a feedforward network rather than along
time---so there is no recurrence and no temporal dynamics.
Our empirical results show that the depth-wise reservoir functions effectively even without enforcing classical ESP constraints.
Characterizing whether spectral-radius conditions and approximation
guarantees of temporal RC transfer to this setting---perhaps via
fan-in scaling and LayerNorm
(Appendix~\ref{app:engineering})---is a natural next step.
Nonetheless, the LoRA adapter at each layer provides a parallel correction
path: $A$ projects the state into a low-dimensional signal and~$B$ maps it
back into the main pathway. This per-layer correction can both counteract
unstable dynamics in the backbone and amplify useful ones within the
frozen backbone, even when~$\Wseed$ violates the
ESP~\cite{jaeger2001echo}.
The trainable~$\beta$ provides a further degree of freedom to attenuate or
amplify the reservoir contribution as needed.
The meta-LoRA experiment (Section~\ref{sec:q3}) provides direct evidence:
at sufficiently high rank, the adapter partially recovers performance even
when the scaffold is resampled during training---suggesting that with enough
capacity, the adapter can compensate for an unstable backbone, though at the
cost of departing from the RC regime.

The RC perspective also illuminates the seed-gating result
(Figure~\ref{fig:seed_gating}): in classical RC, distinct
reservoirs paired with the same readout produce distinct input--output
mappings, because the reservoir geometry determines which features are
linearly accessible.  Analogously, a single shared LottaLoRA adapter paired
with different backbone seeds~$s$ produces different classification
behavior---each seed realizes a different reservoir geometry, exposing a
different subspace of the input to the trained readout.
The points of divergence from RC suggest a complementary connection to
Hyperdimensional Computing~\cite{kanerva2009hyperdimensional}. The
distribution robustness across 22 initialization families
(Section~\ref{sec:q2}) indicates that the scaffold's role is
geometric---providing a high-dimensional space to navigate---rather than
dynamic, which aligns with HDC's framing of random projections as an
intrinsically useful representational substrate. The seed-gated task
specialization (Figure~\ref{fig:seed_gating}) further echoes HDC's binding
operation, in which combining two high-dimensional representations produces a
distinct third.

\section{What Do the Adapter and the Frozen Backbone Each Contribute?}
\label{sec:experiments}

We evaluate LottaLoRA through a series of experiments, each building on the
previous one. We begin by asking whether the scaffold is necessary at all
(Section~\ref{sec:q1}), then test whether its specific initialization
distribution matters (Section~\ref{sec:q2}), and ask whether the
scaffold must remain fixed during training (Section~\ref{sec:q3}).
We then extend the architecture survey to recurrent reservoirs
(Section~\ref{sec:broader}) and test generalization across architecture
families spanning reinforcement learning, graph prediction, and vision
(Section~\ref{sec:additional}), before scaling to large Transformer
language models (Section~\ref{sec:q4}).

\subsection{Does the scaffold contribute beyond the adapter's own capacity?}
\label{sec:q1}

To probe whether the frozen scaffold contributes beyond what the adapter can achieve alone, we
evaluate LottaLoRA on MNIST digit classification under three conditions:
\emph{normal-scaffold} ($\Wseed \sim \mathcal{N}(0, \sigma^2)$, $\sigma =
0.1$), \emph{zero-scaffold} ($\Wseed = 0$, removing the backbone entirely),
and \emph{no-scaffold} evaluation (trained with scaffold, then $\Wseed$
removed after training). For reference, a linear classifier on MNIST
reaches approximately 85\% accuracy; anything below that threshold is not
learning nonlinear structure. Table~\ref{tab:mnist_q1} summarizes the
results.

\begin{table}[ht]
  \caption{MNIST test accuracy (\%) under three backbone conditions. Mean
    over 5 seeds $\pm$ std, ReLU activation, ranks 1/2/4/8.
    Normal-$\Wseed$ numbers are from the
    main evaluation protocol (30 runs, range shows min--max across seeds);
    zero-$\Wseed$ and no-backbone from the
    supplementary protocol (15 runs each, ReLU; see Appendix~\ref{app:mnist}).
    Rank~1 falls below the ${\sim}85\%$ linear-classifier threshold for
    zero-$\Wseed$ and no-backbone, indicating insufficient capacity for
    nonlinear structure at that rank.}
  \label{tab:mnist_q1}
  \centering
  \begin{tabular}{lccc}
    \toprule
    & \multicolumn{3}{c}{Test Accuracy (\%)} \\
    \cmidrule(lr){2-4}
    Rank & $\Wseed = 0$ & Normal-$\Wseed$ & No-$\Wseed$ \\
    \midrule
    1 & $82.9 \pm 4.9$ & 85.1--89.3 & $82.6 \pm 5.6$ \\
    2 & $91.0 \pm 1.9$ & 90.6--92.7 & $90.8 \pm 2.1$ \\
    4 & $94.5 \pm 1.0$ & 93.6--95.4 & $94.4 \pm 1.0$ \\
    8 & $96.3 \pm 0.5$ & 95.4--96.8 & $96.3 \pm 0.6$ \\
    \midrule
    Full & \multicolumn{3}{c}{98.1--98.4} \\
    \bottomrule
  \end{tabular}
\end{table}

\begin{figure*}[ht]
  \centering
  \includegraphics[width=\linewidth]{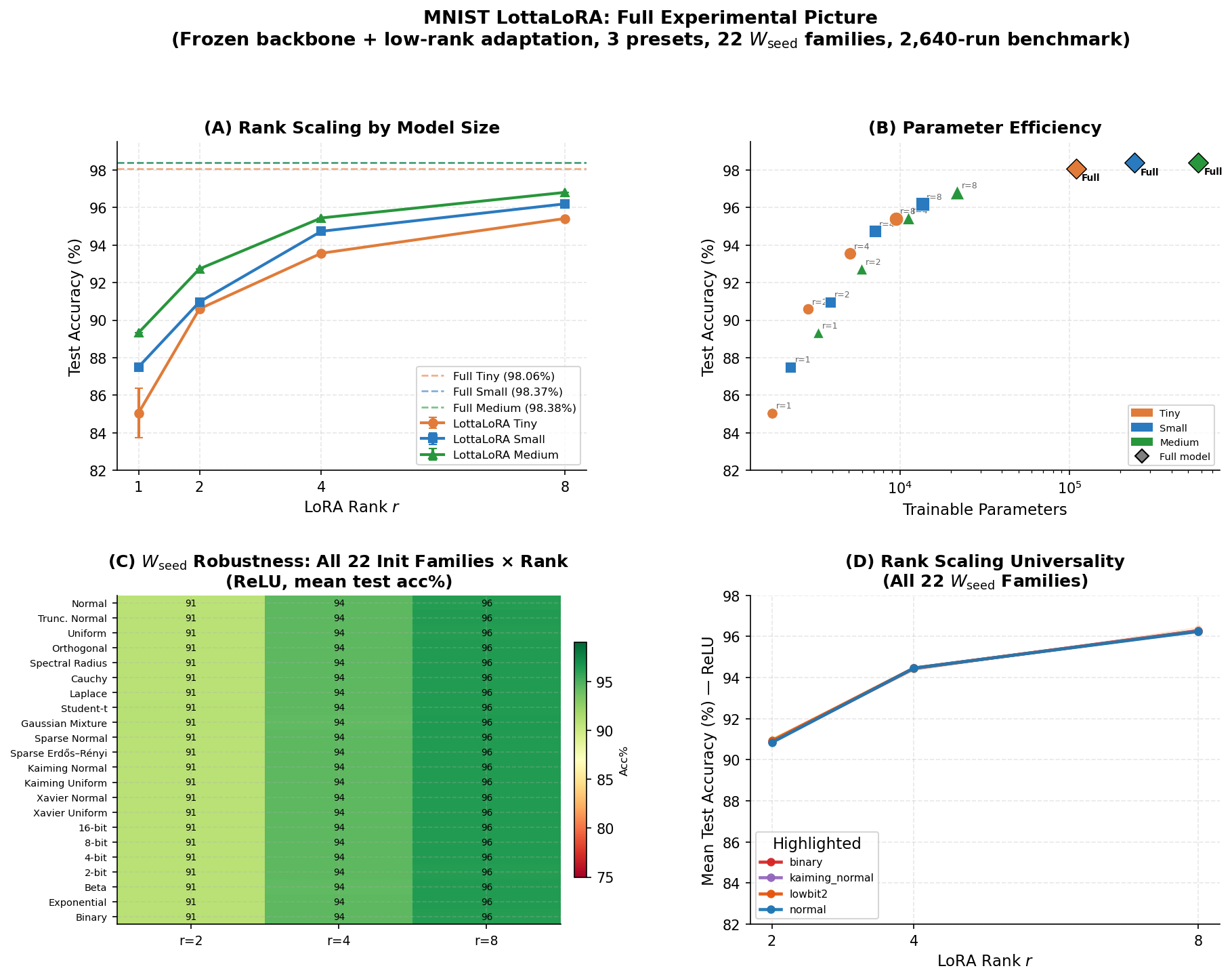}
  \caption{\textbf{MNIST: LottaLoRA accuracy scales monotonically with LoRA rank, closing the gap to fully trained baselines.}
    \textbf{(A)}~Accuracy scales monotonically with LoRA rank across three
    model sizes, closing the gap to fully trained baselines (dashed);
    the medium preset (4 layers, widths 512--64; see
    Appendix~\ref{app:mnist} for all presets) reaches 96.8\% at rank~8
    with only 3.65\% of the parameters of the fully trained counterpart.
    \textbf{(B)}~Parameter efficiency shows that each LottaLoRA configuration
    (circles, squares, triangles) attains comparable accuracy to its
    fully trained counterpart (diamonds) using one to two orders of
    magnitude fewer trainable parameters.
    \textbf{(C)}~All 22 $\Wseed$ initialization families (aggregated
    across tiny, small, and medium presets) exceed 95\% at rank~8,
    indicating that on MNIST the specific distribution has negligible
    effect on performance.
    \textbf{(D)}~Accuracy-vs-rank curves for all 22 families converge
    tightly, showing that the scaffold is interchangeable on this task.}
  \label{fig:mnist_panel}
\end{figure*}

At rank~2 and above, the zero-scaffold condition
matches the accuracy of the
normal-scaffold condition within fractions of a percentage point. On MNIST, the scaffold contributes negligibly to the learned
representation---the LoRA factors carry the task-specific signal. By
rank~4 all three conditions cross the 94\% mark, indicating that nonlinear
task structure is captured within the low-rank factors.
Figure~\ref{fig:mnist_panel}(A) plots accuracy against LoRA rank for three
model sizes: accuracy increases monotonically with rank and the gap to fully
trained baselines shrinks with model size. Panel~(B) shows that each
LottaLoRA configuration achieves comparable accuracy to its fully trained
counterpart using one to two orders of magnitude fewer trainable
parameters.

\begin{figure*}[ht]
  \centering
  \includegraphics[width=\linewidth]{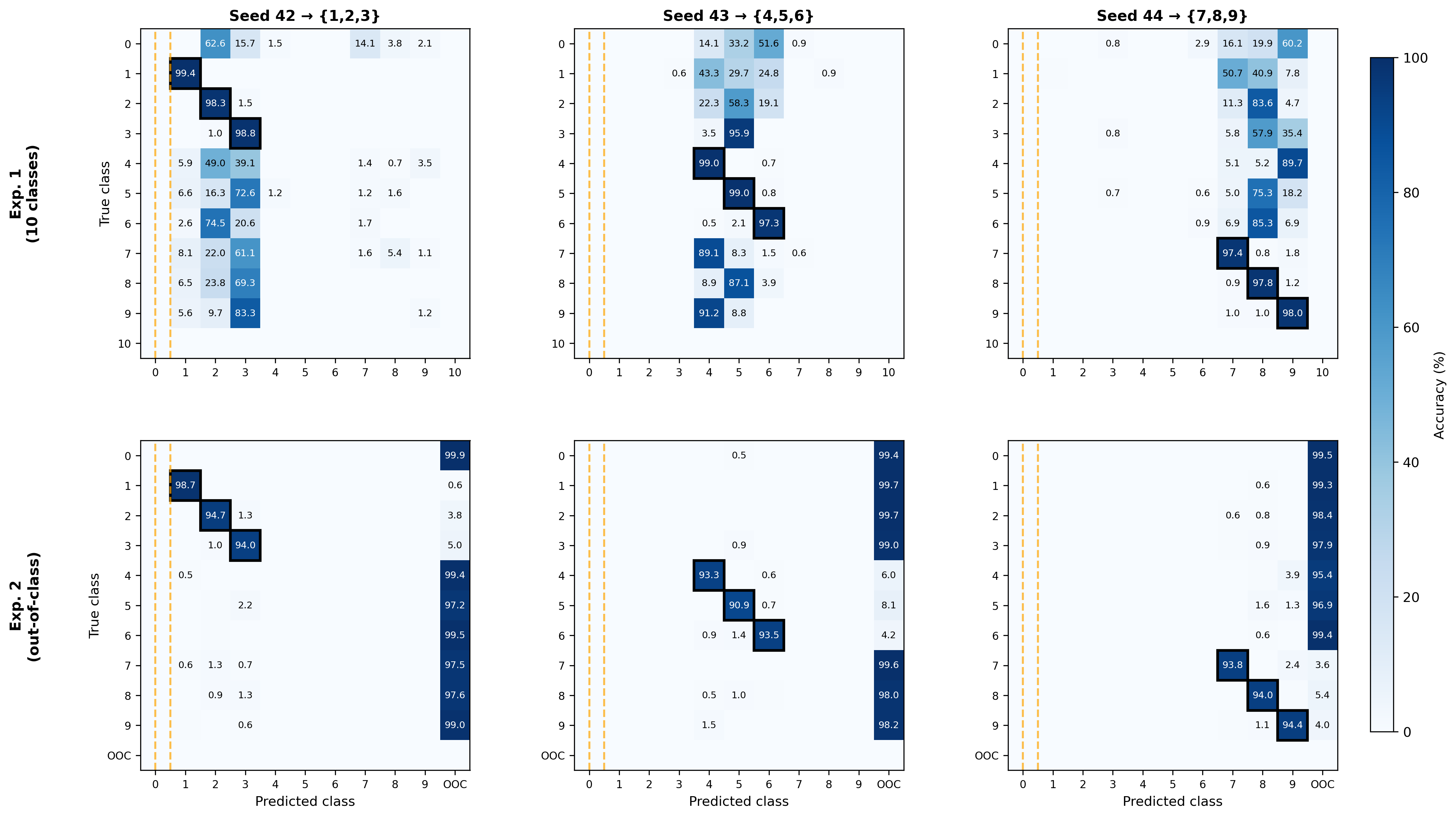}
  \caption{\textbf{A single shared adapter produces seed-gated task
    specialization with out-of-class rejection.}
    One LoRA adapter is trained across three disjoint MNIST label partitions
    ($\{1,2,3\}$, $\{4,5,6\}$, $\{7,8,9\}$), each paired with a distinct
    backbone seed~$s$. Columns show seeds 42, 43, 44; cells show row-normalized
    test accuracy (\%); black rectangles mark assigned classes; dashed orange
    columns highlight digit~0 (excluded from all partitions).
    \textbf{Top row (Experiment~1, 10 classes):} each seed activates 97--99\%
    accuracy on its assigned digits while non-assigned digits are confidently
    misclassified into the in-partition classes.
    \textbf{Bottom row (Experiment~2, out-of-class):} non-assigned digits are
    mapped to an explicit out-of-class label (OOC), achieving 93--99\% accuracy
    on assigned digits and 95--99\% OOC classification on non-assigned digits.
    Digit~0 is consistently mapped to OOC under all seeds.
    Configuration: medium preset, rank~4, 150 training epochs.}
  \label{fig:seed_gating}
\end{figure*}

\paragraph{Seed-gated task specialization (polycomputing).}
The decoupling of adapter and scaffold enables a striking form of
polycomputing---multiple computations sharing the same physical substrate. We partition the MNIST digit classes into three
disjoint subsets---$\{1,2,3\}$, $\{4,5,6\}$, and $\{7,8,9\}$---and train a
\emph{single shared} LoRA adapter across all three subsets simultaneously,
cycling through all three backbone seeds within each epoch: seed $s{=}42$ with labels
$\{1,2,3\}$, $s{=}43$ with $\{4,5,6\}$, and $s{=}44$ with $\{7,8,9\}$.
The adapter parameters are identical for all three tasks; only the seed that
reconstructs~$\Wseed$ differs at inference time.
Figure~\ref{fig:seed_gating} shows the resulting per-digit test accuracy
under two evaluation protocols. In Experiment~1 (10-class), each seed
activates 97--99\% accuracy on its assigned digits while non-assigned
digits are confidently misclassified into in-partition classes.
In Experiment~2, non-assigned digits are mapped to an explicit
out-of-class (OOC) label: the model achieves 93--99\% on assigned digits
and classifies 95--99\% of non-assigned digits as OOC, demonstrating
that seed-gating enables both task specialization and out-of-class
rejection from a single shared adapter. Digit~0, deliberately excluded
from all partitions, is consistently mapped to OOC under every seed.
This extreme specialization arises because each seed produces a different
random scaffold~$\Wseed$; the shared adapter channels each realization
differently, producing task-appropriate behavior only when the correct
scaffold is present. Although demonstrated here on MNIST, the mechanism
depends on the seed-adapter coupling, not on the task.

Taken together, these results show that on MNIST the LoRA factors alone
suffice to solve the task, but that a static scaffold---when
present---is actively incorporated into the computation. The seed-gating
result demonstrates this most directly: a single adapter produces entirely
different functions depending on which scaffold it is paired with,
confirming that the adapter and scaffold jointly determine the network's
behavior.

\subsection{Does a random scaffold help, and does its initialization distribution matter?}
\label{sec:q2}

Section~\ref{sec:q1} showed that the LoRA adapter can solve MNIST even
without a scaffold. Here we ask a different question: if a random scaffold
is present, does it matter \emph{how} it is initialized?
We test this by drawing~$\Wseed$ from 22 different distributions and measuring
whether the choice affects performance.

\paragraph{Distribution robustness.}
We tested 22 different $\Wseed$ distribution families on the MNIST medium
preset (four hidden layers, $d{=}512, 256, 128, 64$)---ranging from
standard continuous distributions to binary and sparse extremes
(see Appendix~\ref{app:mnist} for the full list and
definitions)---and measured accuracy at each rank.
Table~\ref{tab:w0_robust} summarizes the results.
At rank~8, every initialization family stays above 95\% accuracy, with a
worst-to-best spread of at most 0.30\,percentage points (pp).
Figure~\ref{fig:mnist_panel}(C) visualizes this as a heatmap across all
families and ranks; panel~(D) of the same figure plots accuracy against rank
for each family, showing that all 22 curves converge tightly.
On MNIST, the initialization distribution has negligible effect on
performance---the scaffold's specific values are interchangeable. This does
not mean the scaffold is unused: under all 22 static distributions, the
learned $\beta$ remains strictly positive (Section~\ref{sec:q3}), confirming that
the optimizer exploits the backbone regardless of how it is initialized.
We demonstrate this robustness across 22 initialization families on MNIST; extending it to larger-scale tasks is a natural next step (see Section~\ref{sec:discussion}).

\begin{table}[ht]
  \caption{MNIST test classification accuracy (\%) across 22~$\Wseed$
    initialization families. Each family is evaluated over 15 seeds
    (3 architecture sizes---tiny, small, medium---$\times$ 5 seeds;
    see Appendix~\ref{app:mnist} for details). Range shows worst-to-best family mean.}
  \label{tab:w0_robust}
  \centering
  \begin{tabular}{rccc}
    \toprule
    Rank & Mean (\%) & Range (\%) & Spread (pp) \\
    \midrule
    Full & 98.03 & 97.91--98.21 & 0.30 \\
    $r{=}2$ & 90.88 & 90.84--91.03 & 0.19 \\
    $r{=}4$ & 94.45 & 94.33--94.48 & 0.14 \\
    $r{=}8$ & 96.29 & 96.24--96.44 & 0.19 \\
    \bottomrule
  \end{tabular}
\end{table}

\paragraph{Low-bit quantization.}
On MNIST, binary~$\Wseed$ (entries from
$\{-1/\sqrt{d},\allowbreak +1/\sqrt{d}\}$) and 2-bit quantized~$\Wseed$
achieve test accuracy on par with Gaussian initialization (96.4\% and 96.5\%
respectively at rank~8 vs 96.2\% for Gaussian). On this task, the
scaffold can be initialized with a single bit per weight without accuracy
loss.

\subsection{Must the scaffold be static?}
\label{sec:q3}

To determine whether the scaffold must remain fixed or whether its values
can change during training, we resample~$\Wseed$ while training the LoRA
factors continuously. We call this ablation \emph{Meta-LoRA}: the
scaffold~$\Wseed$ is resampled from its initialization distribution at a fixed
cadence while LoRA factors are trained continuously. We test three
resampling schedules of increasing aggressiveness: (1)~once per epoch,
(2)~at each optimizer step, cycling through $k{=}2$ different scaffolds per
update (gradients accumulated across both), and (3)~within each batch,
splitting it into $k{=}4$ sub-batches each processed with a different~$\Wseed$.
Table~\ref{tab:meta_ablation} summarizes the results.

\begin{table}[ht]
  \caption{Resampling the scaffold during training degrades performance progressively with resampling frequency (MNIST, medium preset). Standard = static $\Wseed$.}
  \label{tab:meta_ablation}
  \centering
  \begin{tabular}{lccc}
    \toprule
    Schedule & $r{=}2$ & $r{=}4$ & $r{=}8$ \\
    \midrule
    Standard (static) & 92.61 & 95.53 & 96.80 \\
    Epoch & 67.96 & 89.00 & 95.53 \\
    Batch ($k{=}2$) & 42.46 & 54.94 & 91.19 \\
    Microbatch ($k{=}4$) & 41.49 & 47.67 & 56.29 \\
    \bottomrule
  \end{tabular}
\end{table}

Performance collapses under aggressive resampling (Table~\ref{tab:meta_ablation}): at rank~2, microbatch resampling drops accuracy
from 92.6\% to 41.5\%---a 51\,pp loss.

The learned $\beta$ scalars reveal whether the backbone is exploited or
silenced. Under static scaffolds, $\beta$ remains strictly positive---the optimizer
retains the backbone as a computational substrate (magnitude varies by
architecture: median ${\approx}0.91$ for Transformers and CfC reservoirs,
${\approx}0.99$ for ViT, and ${\approx}0.48$ for Decision Transformers). Under aggressive
resampling, $\beta$ collapses toward zero at high rank: the optimizer
silences the unreliable scaffold and the LoRA factors must carry all task
information alone. The scaffold provides a stable reference frame against
which the LoRA factors can learn; resampling destroys that frame, analogous
to changing the reservoir mid-training in classical RC.

Table~\ref{tab:meta_ablation} shows a rank-dependent recovery pattern.
At high rank, even aggressive resampling partially recovers performance
(91.2\% at rank~8 under batch resampling), while at low rank performance
collapses entirely. This is consistent with a compression interpretation:
a low-rank adapter has limited capacity, so it must rely on the scaffold as
a stable 
reference frame
to encode task-relevant structure efficiently.
When the scaffold is resampled, the adapter cannot rely on it and must
instead absorb all task information into its own trainable parameters---but
at low rank it lacks the capacity to do so. A high-rank adapter, by
contrast, has enough degrees of freedom to encode the task independently,
making the scaffold less critical.
This suggests a ``sweet spot'': at intermediate ranks, the adapter benefits
from the reservoir's fixed dynamics; at high ranks, the adapter can
represent the task independently. This also explains why higher ranks close
the gap to full training even under the fixed-scaffold condition: the
adapter progressively internalizes more of the task structure rather than
depending on the scaffold's structure. Note that under resampling, the
adapter faces an additional burden beyond learning the task itself---it must
also learn to be robust to scaffold changes, which consumes capacity that
would otherwise serve the task. The resampling experiment thus doubles as a
probe to identify the minimum rank at which the LoRA factors alone are
sufficient: the rank at which resampled performance recovers to match the
fixed-scaffold condition marks the point where the adapter's capacity
exceeds the task's requirements. This rank provides an estimate of the
number of trainable degrees of freedom needed to represent the task (each
rank contributes $d_\mathrm{in} + d_\mathrm{out}$ parameters per layer).

The scaffold must be static: its fixedness is the necessary condition for
the adapter to co-adapt to its structure.

\subsection{Time-Series Prediction with Reservoir RNNs}
\label{sec:broader}

The MNIST experiments above apply LottaLoRA to feedforward
architectures, where the reservoir unfolds along depth. Here we test
whether the same principle works in the classical RC setting: a frozen
random \emph{recurrent} network with LoRA adapters replacing the trained
readout. This directly connects LottaLoRA to its RC roots, where the
reservoir unfolds along time.

\paragraph{CfC reservoir on PhysioNet 2012 ICU mortality.}
To test LottaLoRA on a real-world recurrent task, we apply it to ICU
mortality prediction on PhysioNet 2012~\cite{silva2012predicting}---a
binary classification benchmark with severe class imbalance (11.7:1).
The backbone is a frozen Closed-form Continuous-time network
(CfC~\cite{hasani2021liquid,hasani2022closed}) with hidden size~256 and a
2-layer backbone of 64~units each, initialized from a random seed.  LoRA adapters replace the trained weight matrices
in each recurrent layer; the backbone remains frozen throughout.
Performance is measured by AUROC (area under the ROC curve; higher is
better).  We compare against a fully trained CfC baseline using
identical architecture and hyperparameters
(Appendix~\ref{app:cfc_physionet}).

\begin{table}[ht]
  \caption{\textbf{CfC reservoir on PhysioNet 2012 ICU mortality}
    ($h{=}256$, backbone 64~units, 2~layers): AUROC and trainable parameter count by LoRA
    rank.  LottaLoRA uses a frozen random CfC backbone; Full CfC is
    fully trained from scratch.  Mean $\pm$ std over 5~seeds.
    \%Trainable is relative to the Full CfC parameter count
    (92{,}930).}
  \label{tab:cfc_physionet}
  \centering
  \small
  \begin{tabular}{lccc}
    \toprule
    Rank & AUROC & \% & \%Trainable \\
    \midrule
    $r{=}1$ & $0.832 \pm 0.005$ &  3{,}482 &  3.7\% \\
    $r{=}2$ & $0.835 \pm 0.005$ &  5{,}292 &  5.7\% \\
    $r{=}4$ & $0.834 \pm 0.002$ &  8{,}912 &  9.6\% \\
    $r{=}8$ & $0.833 \pm 0.001$ & 16{,}152 & 17.4\% \\
    \cmidrule{2-4}
    Full    & $0.836 \pm 0.002$ & 92{,}930 & 100\%  \\
    \bottomrule
  \end{tabular}
\end{table}

\begin{figure}[ht]
  \centering
  \includegraphics[width=0.85\linewidth]{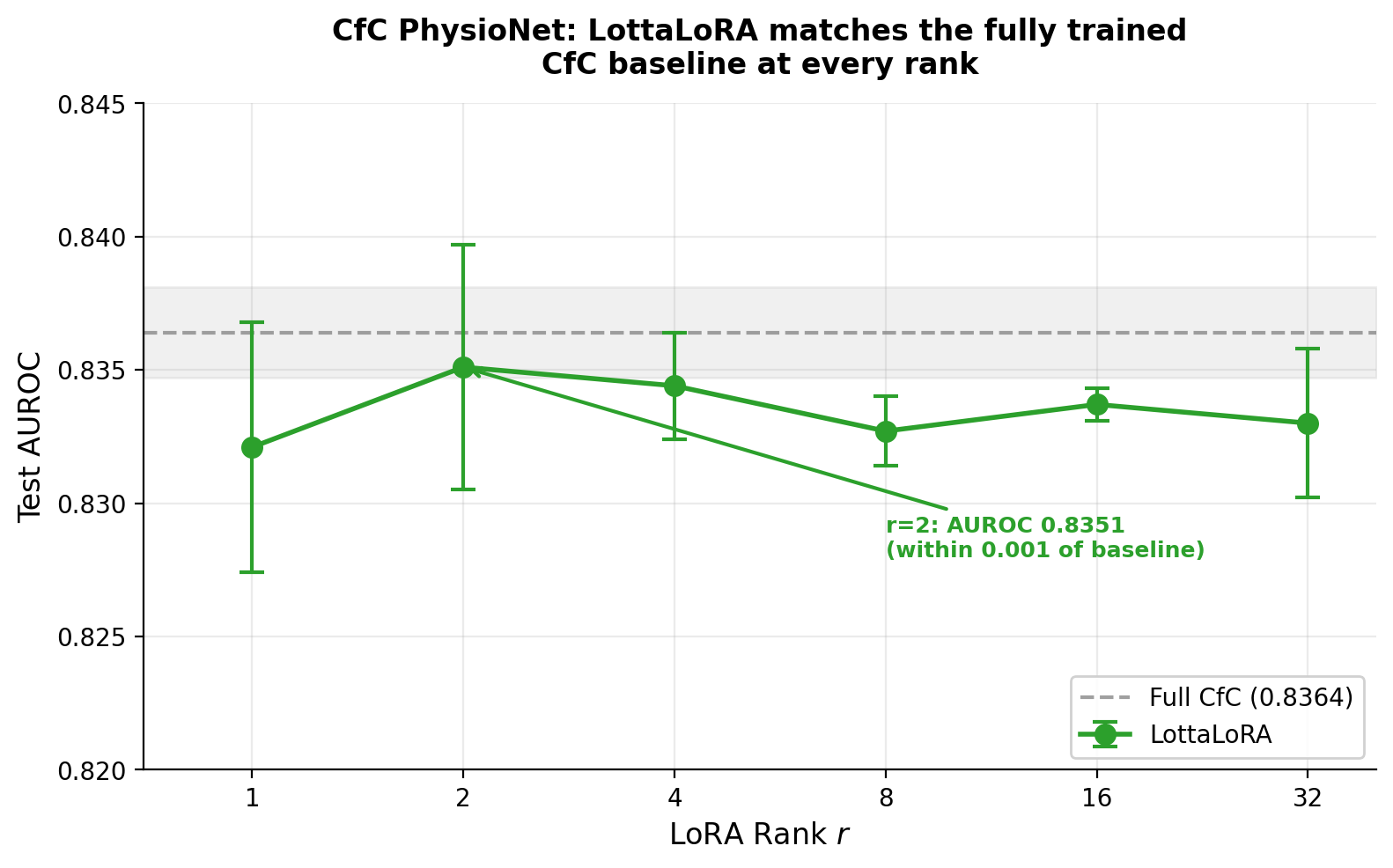}
  \caption{\textbf{LottaLoRA AUROC saturates at rank~2 on PhysioNet
    2012 ICU mortality.}  Mean $\pm$ std AUROC over 5~seeds at ranks
    1--32.  Dashed line: fully trained CfC baseline (0.836).  Rank~1
    recovers 99.5\% of baseline with 3.7\% of trainable parameters.}
  \label{fig:cfc_physionet_rank}
\end{figure}

Table~\ref{tab:cfc_physionet} shows that LottaLoRA with rank~1
achieves AUROC~$= 0.832 \pm 0.005$, recovering 99.5\% of the fully
trained CfC baseline ($0.836 \pm 0.002$) while using only 3{,}482
trainable parameters---3.7\% of the baseline's 92{,}930.  Performance
saturates at rank~2 (Figure~\ref{fig:cfc_physionet_rank}) with no
further gain at higher ranks, confirming low intrinsic dimensionality
for this clinical prediction task.

However, because the published CfC architecture (92{,}930~parameters) is
substantially overparameterized for this binary-classification task
(3{,}200~samples, 37~features), the flat saturation curve may reflect
architectural excess rather than genuinely low intrinsic dimensionality.
To test this, we scaled the CfC hidden size and backbone width by factors of
$0.125\times$, $0.25\times$, and $0.5\times$ and repeated the rank sweep
(5~seeds each; Appendix~\ref{app:cfc_physionet}).
At $0.125\times$ (2{,}210~parameters), rank~1 recovers only
$93.7\%$ of the scaled baseline (AUROC~$0.778 \pm 0.037$), and saturation
shifts to $r{=}4$; at $0.25\times$ and above, rank~1 already recovers
${\geq}98.8\%$.  Meanwhile, the fully trained baseline improves only
from 0.831 to 0.836 across a $42\times$ parameter increase
(Table~\ref{tab:cfc_size_sweep}),
confirming that the architecture is far larger than the task requires.
The saturation rank therefore reflects an interaction between task complexity
and model capacity, not task complexity alone.

\begin{table}[ht]
  \caption{\textbf{Reducing CfC capacity shifts rank saturation rightward.}
    AUROC (mean $\pm$ std, 5~seeds) at each size scale and LoRA rank.
    At $0.125\times$, rank~1 falls to $93.7\%$ recovery; rank~4 restores
    $99.8\%$.  Fully trained baselines (``Full'') improve only
    0.831--0.836 across a $42\times$ parameter range.}
  \label{tab:cfc_size_sweep}
  \centering
  \small
  \begin{tabular}{rrccc}
    \toprule
    Scale & $h$ & Rank & AUROC & Recovery \\
    \midrule
    $0.125\times$ & 32 & Full & $0.831 \pm 0.003$ & --- \\
    & & $r{=}1$ & $0.778 \pm 0.037$ & 93.7\% \\
    & & $r{=}4$ & $0.829 \pm 0.005$ & 99.8\% \\
    \cmidrule{1-5}
    $0.25\times$ & 64 & Full & $0.832 \pm 0.002$ & --- \\
    & & $r{=}1$ & $0.822 \pm 0.009$ & 98.8\% \\
    & & $r{=}2$ & $0.831 \pm 0.002$ & 100.0\% \\
    \cmidrule{1-5}
    $0.5\times$ & 128 & Full & $0.833 \pm 0.001$ & --- \\
    & & $r{=}1$ & $0.827 \pm 0.004$ & 99.3\% \\
    & & $r{=}4$ & $0.831 \pm 0.002$ & 99.8\% \\
    \cmidrule{1-5}
    $1\times$ & 256 & Full & $0.836 \pm 0.002$ & --- \\
    & & $r{=}1$ & $0.832 \pm 0.005$ & 99.5\% \\
    \bottomrule
  \end{tabular}
\end{table}

\begin{figure}[ht]
  \centering
  \includegraphics[width=\linewidth]{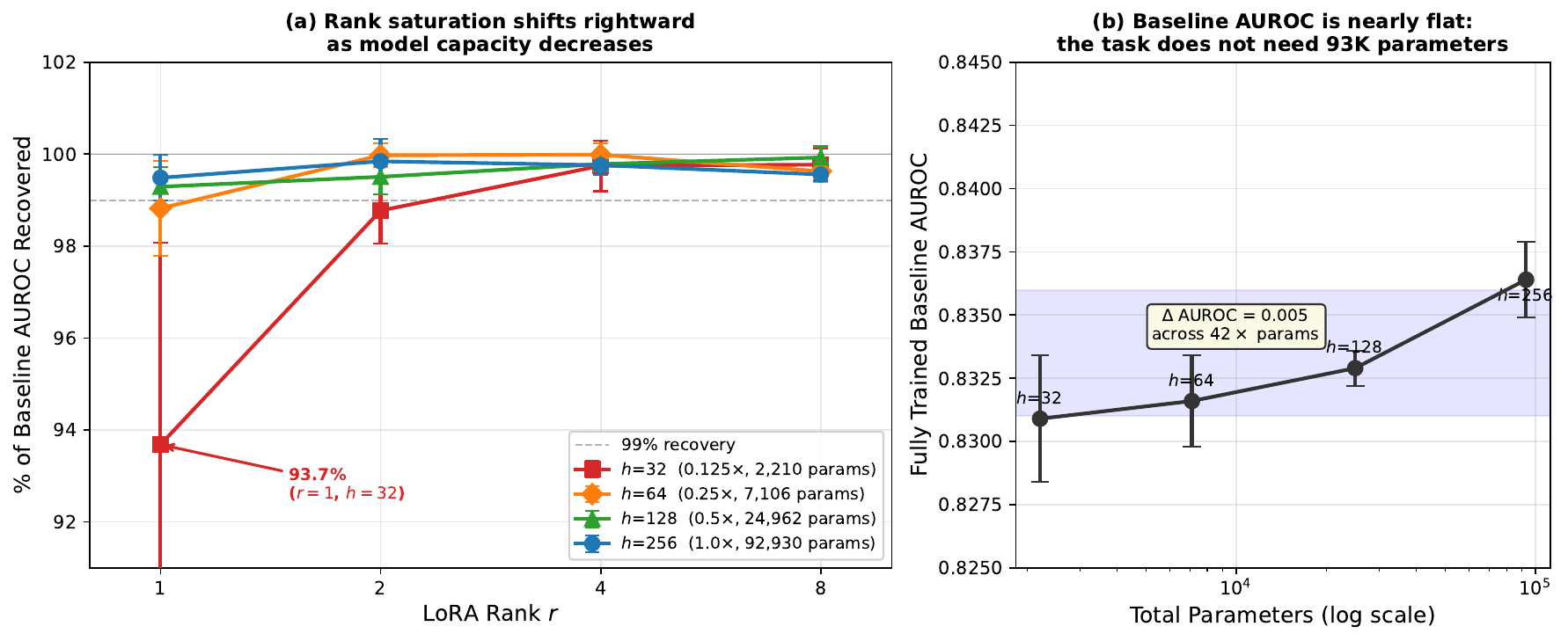}
  \caption{\textbf{Overparameterization masks rank saturation on CfC
    PhysioNet.}  \textbf{(a)}~LottaLoRA recovery (normalized to each
    scale's fully trained baseline) vs.\ LoRA rank at four CfC sizes.
    At $0.125\times$ ($h{=}32$, 2{,}210~parameters), rank~1 recovers
    only 93.7\% and saturation shifts to $r{=}4$; at larger scales
    rank~1 already exceeds 98.8\%.
    \textbf{(b)}~Fully trained baselines improve only from 0.831 to
    0.836 across a $42\times$ parameter increase, confirming that the
    published CfC architecture is far larger than this task requires.
    See Appendix~\ref{app:cfc_physionet} for the full
    overparameterization analysis.}
  \label{fig:cfc_size_sweep}
\end{figure}

On real clinical time-series, LottaLoRA operates in the regime it
inherits from RC: a frozen recurrent backbone with a trained readout.
At the published architecture scale, rank-1 adapters match the fully
trained CfC on ICU mortality prediction; the size-reduction ablation
(Figure~\ref{fig:cfc_size_sweep}) shows that this flat saturation
reflects architectural excess, not genuinely low task dimensionality.
A randomly initialized frozen backbone provides
sufficient temporal mixing without task-adaptive training of its weights.

\subsection{Generalization Across Architecture Families}
\label{sec:additional}

A central claim of this work is that the low-rank subspace phenomenon is
architecture-general, not an artifact of a specific model family. To test
this, we apply LottaLoRA to six additional domains spanning offline
reinforcement learning, convolutional image classification, graph-level
and node-level graph prediction, fine-grained image classification,
and natural language sentiment classification
(Table~\ref{tab:broader_prelim}). In each case, LoRA adapters replace the
trainable weight matrices of the frozen backbone, following the same
$W_\mathrm{eff} = \beta\,\Wseed + (\alpha/r)\,BA$ formulation.
\textbf{CIFAR-10 Plain-20 CNN}~\cite{he2016deep}: 10-class image
classification using a 20-layer convolutional network
(3~seeds; Appendix~\ref{app:cifar10}).
\textbf{Decision Transformers}~\cite{chen2021decision}: offline RL on
HalfCheetah-expert-v2, with a rank sweep and two scaffold stability
conditions (5~seeds; Appendix~\ref{app:dt}).
\textbf{OGBG-MolHIV}~\cite{hu2020ogb}: graph-level molecular property
prediction (binary classification) using a GIN~\cite{xu2019gin} (strictly more expressive than GCN~\cite{kipf2017gcn})
(5~seeds; Appendix~\ref{app:molhiv}).
\textbf{OGBN-Arxiv}~\cite{hu2020ogb}: node-level citation classification
using a GCN~\cite{kipf2017gcn}
(10~seeds; Appendix~\ref{app:ogbn_arxiv}).
\textbf{Vision Transformer (ViT)}~\cite{dosovitskiy2021image}: fine-grained
image classification on Flowers-102~\cite{nilsback2008flowers}, using
ViT-Base/16 with an ImageNet-1k pre-trained backbone
(5~seeds; Appendix~\ref{app:vit}).
\textbf{IMDB Sentiment Classification}~\cite{maas2011learning}: binary
sentiment classification using a DistilBERT~\cite{sanh2019distilbert} backbone
(4~seeds; Appendix~\ref{app:imdb}).

\begin{table*}[ht]
  \caption{LottaLoRA is competitive across diverse architectures with 0.5--40\%
    of trainable parameters. All comparisons use matched-protocol training;
    IMDB and ViT use pre-trained backbones, others train from scratch.
    For DT, lower MSE is better. See appendices for full
    details per benchmark.}
  \label{tab:broader_prelim}
  \centering
  \small
  \begin{tabular}{@{\extracolsep{0pt}}p{4.2cm}cccccc@{}}
    \toprule
    Benchmark & Rank & Metric Test & LottaLoRA & Baseline & \#seeds & \% trainable \\
    \midrule
    CIFAR-10 Plain-20   & 16 &  acc         & $87.70 \pm 0.26$\%   & $90.81 \pm 0.20$\%   & 3  & 38.3\% \\
    OGBG-MolHIV (GIN)   & 16 &  ROC-AUC    & $0.7562 \pm 0.0158$  & $0.7755 \pm 0.0060$  & 5  & 10.9\% \\
    OGBN-Arxiv (GCN)    & 32 &  acc        & $70.11 \pm 0.21$\%   & $71.86 \pm 0.22$\%   & 10 & 35.6\% \\
    DT (small)          & 32 & Val MSE$\downarrow$ & $0.0271 \pm 0.0017$ & $0.0272 \pm 0.0017$ & 5  & 37.5\% \\
    DT (large)          &  8 & Val MSE$\downarrow$ & $0.0285 \pm 0.0015$ & $0.0281 \pm 0.0019$ & 5  & 1.87\% \\
    ViT Flowers-102     &  1 & Top-1 acc       & $94.53 \pm 0.35$\%   & $96.83 \pm 0.08$\%   & 5  &  1.19\% \\
    IMDB (DistilBERT)   &  8 &  acc         & $85.12 \pm 0.57$\%   & $85.69 \pm 0.44$\%   & 4  & 0.48\% \\
    \bottomrule
  \end{tabular}
\end{table*}

Six findings stand out. First, LottaLoRA is competitive across all six
domains: on CIFAR-10 (Plain-20 CNN) it recovers 96.6\% of baseline accuracy
at rank~16 with 38.3\% of the parameters (Table~\ref{tab:broader_prelim});
on Decision Transformers at sufficient rank ($r{=}32$, small scale),
LottaLoRA matches the fully trained baseline
(Val~MSE 0.0271 vs 0.0272; paired difference not significant) while
training 37.5\% of the parameters, and at extreme compression
($r{=}8$, large scale, 1.87\% trainable) it trails by only 1.2\% relative MSE;
on ViT Flowers-102 it recovers 97.6\% of baseline
accuracy with 1.19\% of the parameters; and on OGBG-MolHIV it recovers
97.5\% of baseline ROC-AUC with 10.9\% of the parameters.
Second, the Decision Transformer sweep confirms the resampling collapse
observed on MNIST (Section~\ref{sec:q3}) on a second architecture: resampling
the scaffold each epoch drops D4RL online score from 61\% to 4\% at rank~1,
with recovery following the same rank-dependent pattern
(Appendix~\ref{app:dt}).
Third, the OGBG-MolHIV result demonstrates generalization to graph-level
molecular property prediction: LottaLoRA at $r{=}16$ achieves ROC-AUC
0.7562, recovering 97.5\% of the fully trained baseline (0.7755) while
training only 10.9\% of the parameters. Notably, the LottaLoRA score
matches the published OGB GIN baseline
($0.7558 \pm 0.0140$~\cite{hu2020ogb}), meaning a frozen-backbone model
with ${\sim}$200\,K trainable parameters performs on par with a fully
trained GIN with ${\sim}$1.8\,M parameters.
Fourth, the ViT result quantifies the value of a learned reservoir:
replacing ImageNet-1k weights with random noise costs ${\sim}$40\,pp,
showing that while LottaLoRA works with random backbones, learned
reservoirs provide a substantial advantage on complex vision tasks.
Fifth, OGBN-Arxiv extends graph generalization from graph-level
(MolHIV, GIN) to node-level classification (GCN): LottaLoRA at $r{=}32$
recovers 97.6\% of the fully trained baseline ($70.11\%$ vs $71.86\%$)
while training 35.6\% of the parameters (Table~\ref{tab:broader_prelim}).

Sixth, IMDB sentiment classification extends generalization to NLP: LottaLoRA
at $r{=}8$ recovers 99.3\% of full fine-tuning accuracy ($85.12\%$ vs
$85.69\%$) while training only 0.48\% of the parameters---the lowest
trainable ratio of any benchmark in this paper
(Table~\ref{tab:broader_prelim}; Appendix~\ref{app:imdb}).

Across all nine benchmarks (MNIST, CfC PhysioNet, CIFAR-10, OGBG-MolHIV, OGBN-Arxiv,
Decision Transformer, ViT, IMDB, WikiText---each representing a distinct architecture), a
consistent pattern emerges: increasing LoRA rank yields decreasing marginal gains,
and different tasks saturate at different ranks.

\subsection{Scaling to Transformers}
\label{sec:q4}

The preceding experiments establish the scaffold's role across feedforward,
recurrent, and graph architectures; we now ask whether the method scales to
modern language models.
We compare full-parameter training and LottaLoRA on decoder-only
Transformer language models at five backbone scales (3\,M to
900\,M total parameters) on WikiText-103~\cite{merity2016pointer},
training from scratch under a single-epoch compute-optimal budget
following the Chinchilla protocol~\cite{hoffmann2022training}.
Neither condition uses a pre-trained checkpoint, so the comparison
isolates the value of full weight optimization against random frozen
weights steered by low-rank adapters
(Appendix~\ref{app:quantitative}).

\begin{table*}[ht]
  \caption{Transformer scaling on WikiText-103: full training versus LottaLoRA $r{=}8$ at each
    backbone scale. Internal trainable ratio (``Ratio (\%)'') counts only transformer-internal
    parameters (attention projections and feedforward layers, excluding token embeddings and the language-model output head). Best loss per scale in \textbf{bold}; $\Delta$ is the gap LottaLoRA$\;-\;$Full.}
  \label{tab:scaling_summary}
  \centering
  \begin{tabular}{lcccc}
    \toprule
    Scale & Full loss & LottaLoRA $r{=}8$ loss & Ratio (\%) & $\Delta$ \\
    \midrule
    3M   & \textbf{5.007} & 5.639 & 7.68 & +0.63 \\
    30M  & \textbf{3.757} & 4.681 & 1.39 & +0.92 \\
    300M & \textbf{3.252} & 4.191 & 0.51 & +0.94 \\
    600M & \textbf{3.185} & 4.058 & 0.40 & +0.87 \\
    900M & \textbf{3.156} & 3.950 & 0.31 & +0.79 \\
    \bottomrule
  \end{tabular}
\end{table*}

Under the same sample budget, LottaLoRA does not yet surpass full training
at any scale, but the gap narrows steadily from $+$0.94\,nats at 300\,M to $+$0.79\,nats
at 900\,M (Table~\ref{tab:scaling_summary})---while training fewer than 0.5\% as many internal parameters as the full model.
The narrowing trend is consistent with the RC
literature~\cite{jaeger2001echo,maass2002real}:
a fixed random reservoir must be sufficiently
high-dimensional before its random projections span the structure needed for
effective learning, and the trajectory suggests convergence at larger scales.

\begin{figure}[ht]
  \centering
  \includegraphics[width=0.75\linewidth]{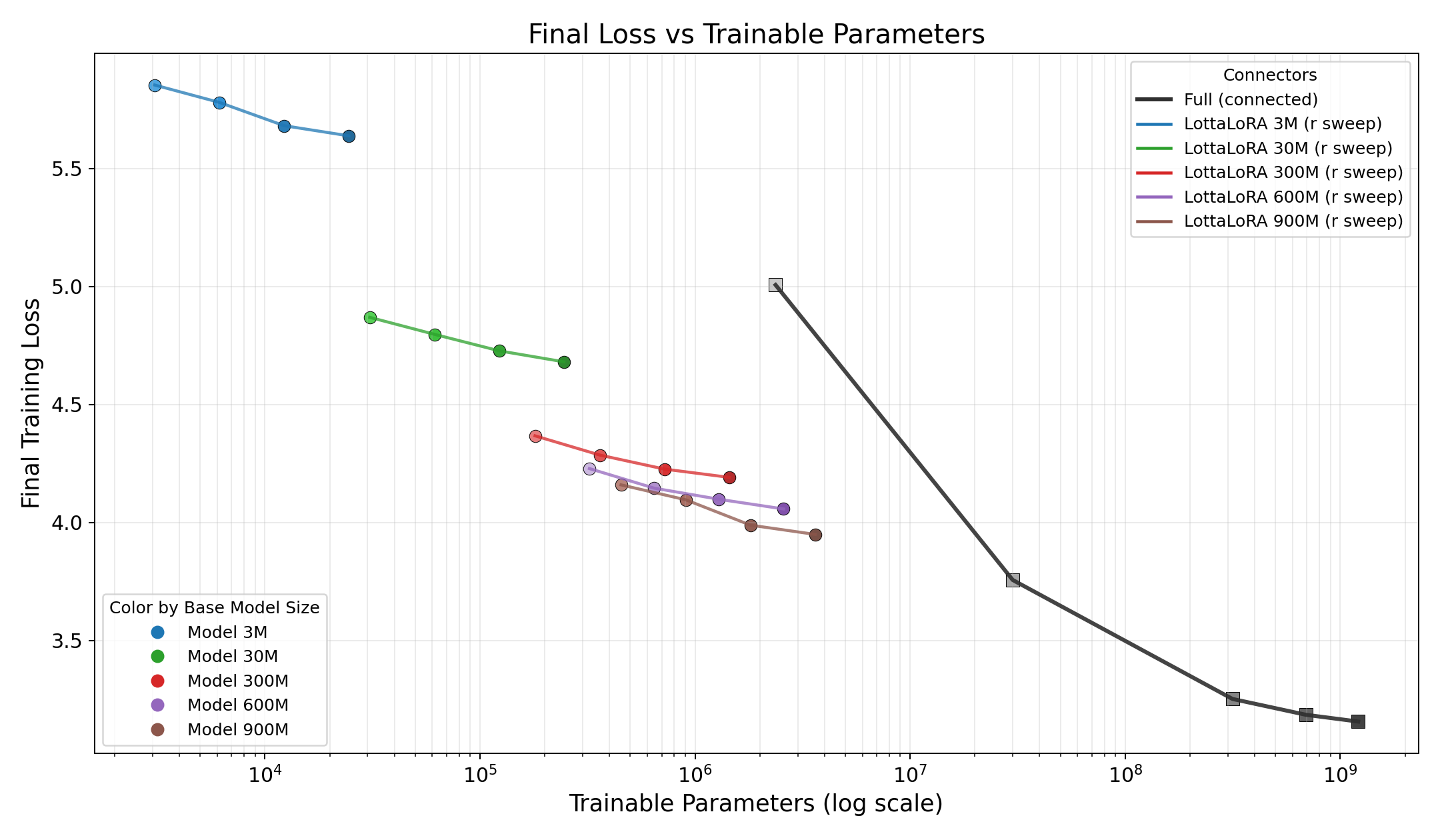}
  \caption{\textbf{A large frozen backbone with few LoRA parameters outperforms a small fully trained model.}
    Each colored curve shows LottaLoRA at one backbone scale across ranks;
    gray squares show fully trained baselines. At 900\,M, rank-8 LottaLoRA
    (3.6\,M trainable) achieves loss~3.950, while the fully trained 3\,M
    model (320\,K trainable) reaches only~5.007.}
  \label{fig:param_vs_loss}
\end{figure}

These same-scale comparisons (each row of Table~\ref{tab:scaling_summary})
measure efficiency at matched forward-pass cost: both conditions use the same architecture with equal total parameter count,
but LottaLoRA freezes the backbone and trains only a tiny fraction of its
internal weights via LoRA adapters.
A complementary view compares models at matched \emph{trainable parameter
count} (Figure~\ref{fig:param_vs_loss}). Consider two ways to spend a budget of
roughly 3\,M trainable parameters: (a)~train all weights of a 3\,M-parameter
model (loss~5.007), or (b)~distribute them as rank-8 LoRA adapters over a
900\,M frozen backbone (loss~3.950). Option~(b) achieves a loss more than 1.0 lower than option~(a), despite
training a comparable number of parameters. This shows that the frozen
backbone---even though random---provides a useful computational substrate:
a small adapter over a large random scaffold extracts more from the same
parameter budget than a small fully trained model can.
This is the central practical result: the frozen random backbone acts as a
computational amplifier, converting a fixed parameter budget into
substantially lower loss.

\begin{figure}[ht]
  \centering
  \includegraphics[width=0.90\linewidth]{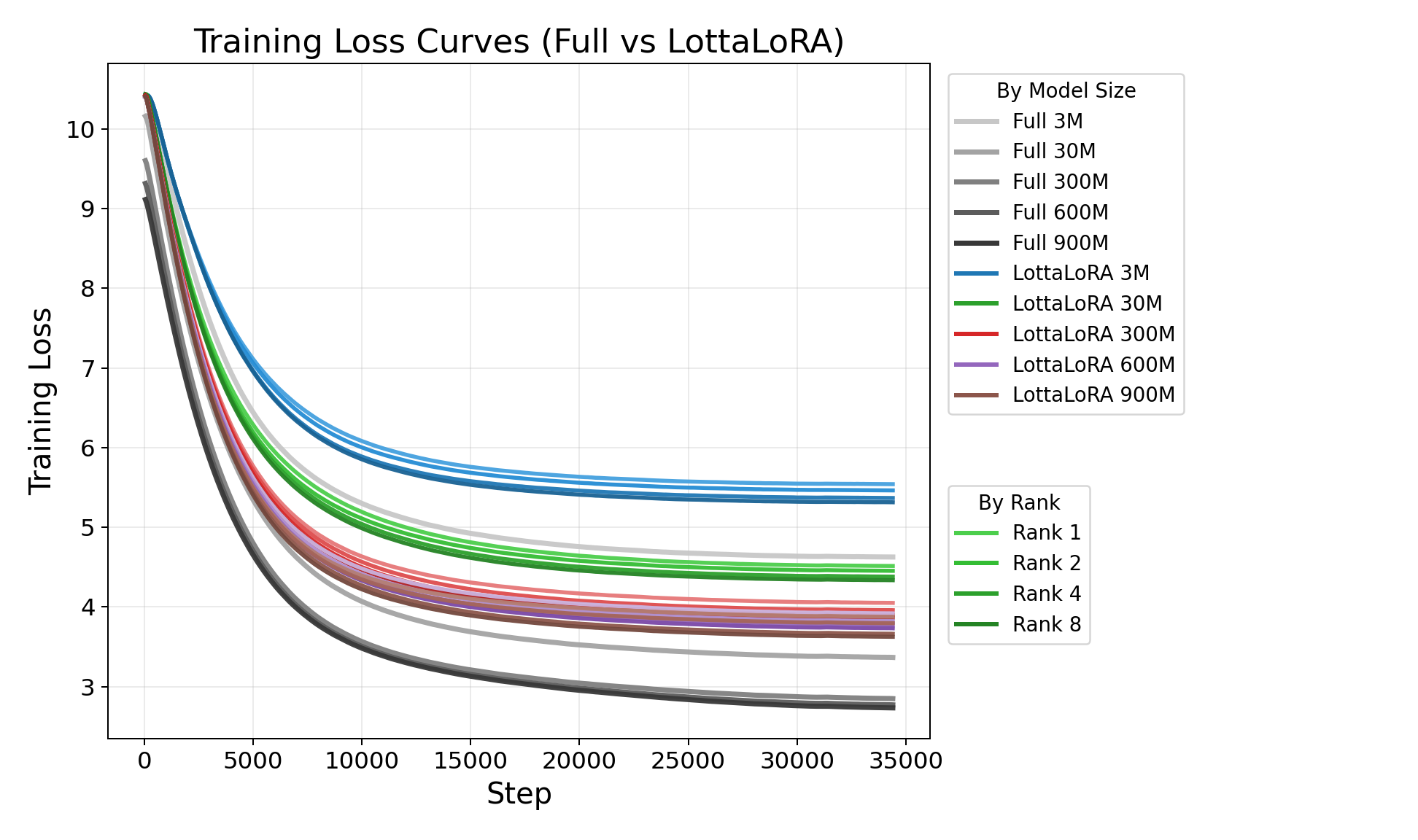}
  \caption{\textbf{LottaLoRA narrows the gap to full training as backbone size
    increases.} Training loss curves on WikiText-103 at five scales (3\,M to
    900\,M); colored curves show LottaLoRA (hue encodes scale, lightness
    encodes rank), grayscale shows fully trained baselines. At 900\,M, the
    best LottaLoRA run (rank~8) reaches 3.950 vs 3.156 for full training,
    while training fewer than 0.5\% of the internal parameters.}
  \label{fig:loss_curves}
\end{figure}

The gap between LottaLoRA and full training narrows steadily with scale
(Figure~\ref{fig:loss_curves}),
consistent with the RC prediction that larger random reservoirs provide
richer projections. At 900\,M parameters, a random frozen backbone with
rank-8 adapters closes to within 0.79\,nats of full training while
optimizing fewer than 0.5\% of the internal parameters. Moreover, at
matched trainable parameter count, the 900\,M LottaLoRA model
(loss~3.950) substantially outperforms a fully trained 3\,M model
(loss~5.007), demonstrating that the frozen backbone amplifies the value
of each trainable parameter.

\section{Computational Cost Analysis}
\label{sec:cost}

The results in Section~\ref{sec:experiments} show a large reduction in
trainable parameter count: at large backbone scales, LottaLoRA narrows the gap to full training
while using fewer than 0.5\% of the internal trainable parameters of a
same-scale fully trained model.
A natural question is whether this parametric advantage translates into a
practical reduction in training cost, measured in floating-point operations (FLOPs) and memory.
We analyze this question by deriving closed-form estimates for both quantities,
comparing LottaLoRA against full-parameter training first on conventional GPU hardware
and then on dedicated accelerators.
The central finding is that on conventional hardware the savings are real but
modest in absolute terms; the full benefit of the parametric advantage is realized
only when the frozen backbone is executed on application-specific integrated circuit (ASIC) accelerators designed for
fixed-weight computation.
The analysis follows a physicist-style approximation that retains only the
dominant terms.

\paragraph{FLOPs.}
For a linear layer $y = Wx$, a forward pass requires $\sim 2d_\mathrm{in}
d_\mathrm{out}$ FLOPs.
The backward pass adds another $\sim 2d_\mathrm{in}d_\mathrm{out}$ to compute
input gradients ($dx = W^\top dy$) and a further $\sim 2d_\mathrm{in}
d_\mathrm{out}$ to compute weight gradients ($dW = dy\, x^\top$).
In full-parameter training all three terms apply to every parameter, giving
approximately $6MN$ total FLOPs for $M$ training tokens and $N$ total
parameters (Equation~\eqref{eq:flop_full}):
\begin{equation}
  \label{eq:flop_full}
  F_\mathrm{full} \approx 6MN.
\end{equation}
In LottaLoRA, frozen parameters incur forward and input-gradient costs but not
weight-gradient costs, since gradients must still flow through them to reach
the LoRA adapters.
Denoting trainable and frozen parameter counts by $N_\mathrm{tr}$ and
$N_\mathrm{fr} = N - N_\mathrm{tr}$, respectively, the LottaLoRA training FLOPs
are (Equation~\eqref{eq:flop_lrlm}):
\begin{equation}
  \label{eq:flop_lrlm}
  F_\mathrm{LottaLoRA} \approx M\!\left(4N_\mathrm{fr} + 6N_\mathrm{tr}\right)
  = M\!\left(4N + 2N_\mathrm{tr}\right).
\end{equation}
The ratio of LottaLoRA to full-training FLOPs is therefore
\begin{equation}
  \label{eq:flop_ratio}
  \frac{F_\mathrm{LottaLoRA}}{F_\mathrm{full}}
  = \frac{2}{3} + \frac{N_\mathrm{tr}}{3N}.
\end{equation}
For a LLaMA-like architecture~\cite{touvron2023llama}, $N \approx 12nd^2$ while
$N_\mathrm{LoRA} \approx 18nrd$, so $N_\mathrm{tr}/N \sim r/d \to 0$ as the
model scales at fixed rank $r$.
The FLOPs ratio therefore converges to $2/3$: on conventional GPU hardware,
freezing the backbone saves at most one third of training compute.

\paragraph{Memory.}
Training memory splits into three components: (i)~weight storage,
(ii)~optimizer states and gradients for trainable parameters, and
(iii)~activations.
Under a standard mixed-precision setup (bf16 weights and gradients, Adam with
fp32 first and second moments and fp32 master weights), each trainable
parameter requires approximately $2 + 2 + 8 + 4 = 16$ bytes, while each
frozen parameter requires only $2$ bytes for storage.
Denoting memory in bytes (Equations~\eqref{eq:mem_lrlm}--\eqref{eq:mem_full}):
\begin{align}
  \label{eq:mem_lrlm}
  \mathrm{Mem}_\mathrm{LottaLoRA}^{\mathrm{param+opt}}
  &\approx 2N + 14N_\mathrm{tr}, \\
  \label{eq:mem_full}
  \mathrm{Mem}_\mathrm{full}^{\mathrm{param+opt}}
  &\approx 16N,
\end{align}
giving a ratio of
\begin{equation}
  \label{eq:mem_ratio}
  \begin{split}
  \frac{\mathrm{Mem}_\mathrm{LottaLoRA}^{\mathrm{param+opt}}}
       {\mathrm{Mem}_\mathrm{full}^{\mathrm{param+opt}}}
  &= \frac{1}{8} + \frac{7}{8}\,\frac{N_\mathrm{tr}}{N} \\
  &\longrightarrow \frac{1}{8}
  \quad\text{as } N_\mathrm{tr}/N \to 0.
  \end{split}
\end{equation}
The dominant memory saving therefore comes from eliminating optimizer states
for frozen parameters, approaching an $8\times$ reduction when $N_\mathrm{tr}
\ll N$.
Activation memory is comparable in both settings because gradients must still
pass through the frozen backbone; it does not contribute meaningfully to the
advantage unless gradient checkpointing is employed.

\paragraph{Empirical GPU measurements.}
Table~\ref{tab:gpu_memory} reports measured peak GPU memory and training
throughput for our WikiText-103 scaling runs on an H200 GPU.
At 900\,M parameters, LottaLoRA uses 19.1\,GB versus 29.8\,GB for full
training---a 36\% reduction in peak memory---consistent with the theoretical
prediction of Equation~\eqref{eq:mem_ratio}.
Training throughput (steps per second) is identical at 300\,M and above,
consistent with the FLOPs analysis: the frozen backbone still performs a full
forward and input-gradient pass, so wall-clock time is not reduced.
The memory saving is available on current hardware; the throughput saving
requires dedicated support for frozen-weight acceleration (see below).

\begin{table*}[ht]
  \caption{Measured peak GPU memory and training throughput
    on an H200 GPU (WikiText-103, batch size 32, 128-token blocks, bf16 mixed
    precision, single epoch = 34{,}429 steps).
    Memory reduction grows with scale as the frozen backbone dominates.}
  \label{tab:gpu_memory}
  \centering
  \small
  \begin{tabular}{lrrrrr}
    \toprule
    Scale & Full (GB) & LottaLoRA $r{=}8$ (GB) & Mem.\ red. & Tput.\ ratio \\
    \midrule
    3\,M   &  1.6 &  1.6 &  1\% & 0.62$\times$ \\
    30\,M  &  2.8 &  2.5 & 10\% & 0.73$\times$ \\
    300\,M & 10.7 &  7.7 & 28\% & 1.00$\times$ \\
    600\,M & 19.3 & 13.0 & 33\% & 1.00$\times$ \\
    900\,M & 29.8 & 19.1 & 36\% & 1.00$\times$ \\
    \bottomrule
  \end{tabular}
\end{table*}

\paragraph{Implications on conventional hardware.}
The ratios in Equations~\eqref{eq:flop_ratio} and~\eqref{eq:mem_ratio} assume the
same total parameter count $N$ in both settings.
The results in Section~\ref{sec:experiments}, however, indicate that LottaLoRA
requires substantially more total parameters than a fully trained
model to reach comparable loss, consistent with the RC
literature: a fixed random reservoir must be considerably larger than a
trained network to achieve comparable expressive power, because the random
projection is not adapted to the task.
At $\alpha N$ total parameters (where $\alpha \gg 1$), the LottaLoRA training FLOPs become approximately
$\alpha N \times (2/3)$, comparable to or worse than full-parameter
training at the target scale $N$.
At inference time, where only a forward pass is required, both regimes incur
approximately $2N$ FLOPs at equal parameter count; the larger LottaLoRA model
therefore yields no inference-time advantage either.
On conventional GPU hardware, a hardware-agnostic comparison does not favor
LottaLoRA in terms of raw compute cost, whether at training or inference.

\subsection{Precision and sparsity of the frozen backbone}
\label{sec:precision}
Because $\Wseed$ is never updated, it can be stored in any reduced-precision
format without penalty. On MNIST (2{,}640 runs, 22 initialization families),
binary $\Wseed$ ($\{-1/\sqrt{d}, +1/\sqrt{d}\}$) and 2-bit quantized $\Wseed$
match Gaussian initialization at rank~8 with no statistically significant
accuracy difference. On IMDB sentiment classification (408-run sweep; see also
Appendix~\ref{app:imdb} for rank sensitivity),
lowbit2 $\Wseed$ is the best-performing variant at ranks~4 and~8. The backbone
can be reduced to a \emph{single bit per weight} with zero degradation.
A binary $\Wseed$ replaces every multiply-accumulate with an add-or-subtract
operation, reducing the arithmetic cost of the backbone forward pass by
roughly $2\times$ on hardware supporting efficient integer
GEMM~\cite{rastegari2016xnor}. At 900\,M parameters, binary storage
requires ${\sim}107$\,MB versus ${\sim}1.7$\,GB at fp16---a $16\times$
reduction that is realizable on commodity GPUs without custom silicon.
This precision tolerance is the empirical prerequisite for the ASIC
argument that follows.

\paragraph{ASIC acceleration.}
The picture changes qualitatively on dedicated accelerators.
A binary backbone is not merely a compressed weight matrix---it is a fixed
sign pattern that replaces floating-point multiply-accumulate with
integer add/subtract, the simplest possible arithmetic primitive.
The frozen backbone performs exactly the same (binary) matrix operations at
every training step and every inference call; the weights never change.
This is precisely the regime in which ASIC accelerators excel: a circuit
designed for a specific, fixed set of operations executes them at far
lower energy and latency than a general-purpose GPU performing the same FLOPs.
A recent survey~\cite{li2024llminference} of LLM inference across CPU, GPU,
field-programmable gate array (FPGA), ASIC, and processing-in-memory platforms demonstrates that ASIC and
FPGA implementations achieve substantially higher energy efficiency---measured
in tokens per joule---than GPU baselines at comparable throughput.
Large-scale inference for commercial language
services is increasingly served by dedicated accelerators, driven by the
economics of sustaining high query volumes at low cost.
In such a setting, the frozen computation in LottaLoRA is not merely cheaper by the
factor of $2/3$ identified above but potentially by one to two orders of
magnitude, depending on the degree of hardware specialization.
Dedicated hardware for neural network inference demonstrates the scale of
this efficiency gain empirically.
Google's Tensor Processing Unit (TPU v1), an ASIC designed specifically for
fixed-weight matrix operations, achieves 30--80$\times$ better
performance-per-watt than contemporary CPUs and GPUs on production ML
workloads~\cite{jouppi2017tpu}; the key enabler is that fixed weights can be
kept in on-chip SRAM and the dataflow hardcoded into the circuit.
LottaLoRA's frozen backbone is precisely this regime: the backbone matrix
multiplications are identical at every forward pass and therefore amenable
to the same weight-stationary ASIC optimization.
The trainable LoRA factors ($A$, $B$, $\beta$) constitute a small
reconfigurable overlay that fits entirely in on-chip buffers, mirroring the
architectural decomposition of emerging fixed-model accelerators into a
hardwired backbone path and a lightweight adapter
path~\cite{li2024llminference}.
The parametric advantage of LottaLoRA---a vastly reduced trainable parameter count
at scale---thus translates into a practical cost advantage primarily in
hardware-co-designed settings where fixed-weight computation can be offloaded
to dedicated circuits.

\paragraph{Device Variability Becomes a Feature: Implications for Neuromorphic Hardware.}
Beyond digital accelerators, LottaLoRA's insensitivity to backbone values
(Section~\ref{sec:q2}) suggests compatibility with computing substrates in
which fixed random connectivity arises naturally.
In analog crossbar arrays, device-to-device variation in memristive
conductances produces weight matrices that are effectively random and
difficult to reprogram; in neuromorphic chips, threshold variability and
synaptic noise introduce similar stochastic structure.  Our results---in
particular the finding that binary and ternary backbones match
continuous-valued ones (Section~\ref{sec:precision})---indicate that such
hardware-intrinsic randomness could serve directly as the frozen scaffold,
with only the compact LoRA adapter implemented in a precise, reconfigurable
digital pathway.
This heterogeneous architecture---an imprecise analog backbone paired with a
small digital adapter---mirrors the LottaLoRA decomposition and may offer a
route toward integrating low-rank adaptation with spiking and analog
classifiers, where the physical substrate provides the high-dimensional
random projection and the trainable adapter provides the task-specific
correction.

\paragraph{Distributable model size: low-rank compression as an alternative to quantization.}
Open-weight large language models are today routinely distributed in
quantized form (GPTQ, AWQ, GGUF) to reduce storage and inference cost.
Quantization achieves this by irreversibly reducing the bit-width of every
learned parameter, trading precision for size. LottaLoRA's structure
suggests an alternative compression pathway that operates on a fundamentally
different axis. Because the frozen backbone is fully determined by a
pseudorandom seed, the distributable footprint of a LottaLoRA model is the
seed plus the compact low-rank adapter---with no precision loss whatsoever.

Table~\ref{tab:dist_size} quantifies this advantage.
At 900\,M parameters with $r{=}8$, the distributable artifact (seed plus
LoRA factors plus embeddings in fp16) is 109\,MB---$6\times$ smaller
than 4-bit quantization (GGUF) and $21\times$ smaller than fp16.
The advantage grows with scale because backbone parameters---which dominate
at large sizes---are replaced by a single seed, while the adapter size grows
only as $O(r \cdot d)$.
At small scales (3\,M, 30\,M), embeddings dominate the distributable size
and the advantage is negligible; at 300\,M and above, the backbone dominates
and seed-based reconstruction provides substantial compression.
The achievable compression ratio is ultimately governed by the task's intrinsic dimensionality (Section~\ref{sec:hypothesis}): lower $r^*$ means a smaller adapter and greater compression, independent of the architecture's total parameter count.

\begin{table*}[ht]
  \caption{Distributable model size at each scale.
    ``fp16'' stores all parameters in half precision;
    ``4-bit'' uses group-quantized 4-bit format (GGUF-style, with fp16
    scale factors per 32-weight group);
    ``LottaLoRA'' stores only the random seed (8\,B), LoRA adapters, and
    token embeddings in fp16.
    Compression ratios are relative to fp16.}
  \label{tab:dist_size}
  \centering
  \small
  \begin{tabular}{lrrrrr}
    \toprule
    Scale & fp16 (MB) & 4-bit (MB) & LottaLoRA (MB) & vs.\ fp16 & vs.\ 4-bit \\
    \midrule
    3\,M   &     5 &     1 &     4 &  1$\times$ & 0.3$\times$ \\
    30\,M  &    57 &    16 &    24 &  2$\times$ & 0.7$\times$ \\
    300\,M &   586 &   165 &   103 &  6$\times$ & 1.6$\times$ \\
    600\,M & 1{,}184 &   333 &   105 & 11$\times$ & 3.2$\times$ \\
    900\,M & 2{,}312 &   650 &   109 & 21$\times$ & 6.0$\times$ \\
    \bottomrule
  \end{tabular}
\end{table*}

If hardware or software infrastructure were developed to exploit this
structure (e.g.\ on-chip PRNG generation of backbone weights, or optimized
sparse-plus-low-rank kernels), the same seed-plus-adapter representation
could simultaneously reduce training cost (gradients flow only through the
adapter), accelerate inference (the backbone is generated rather than loaded
from DRAM), and shrink storage---advantages that quantization, which acts
only on the distributable artifact, cannot jointly provide.

\section{The Minimum-Rank Hypothesis: Toward a Measurement of Task Complexity}
\label{sec:hypothesis}

\begin{figure}[ht]
  \centering
  \includegraphics[width=0.95\linewidth]{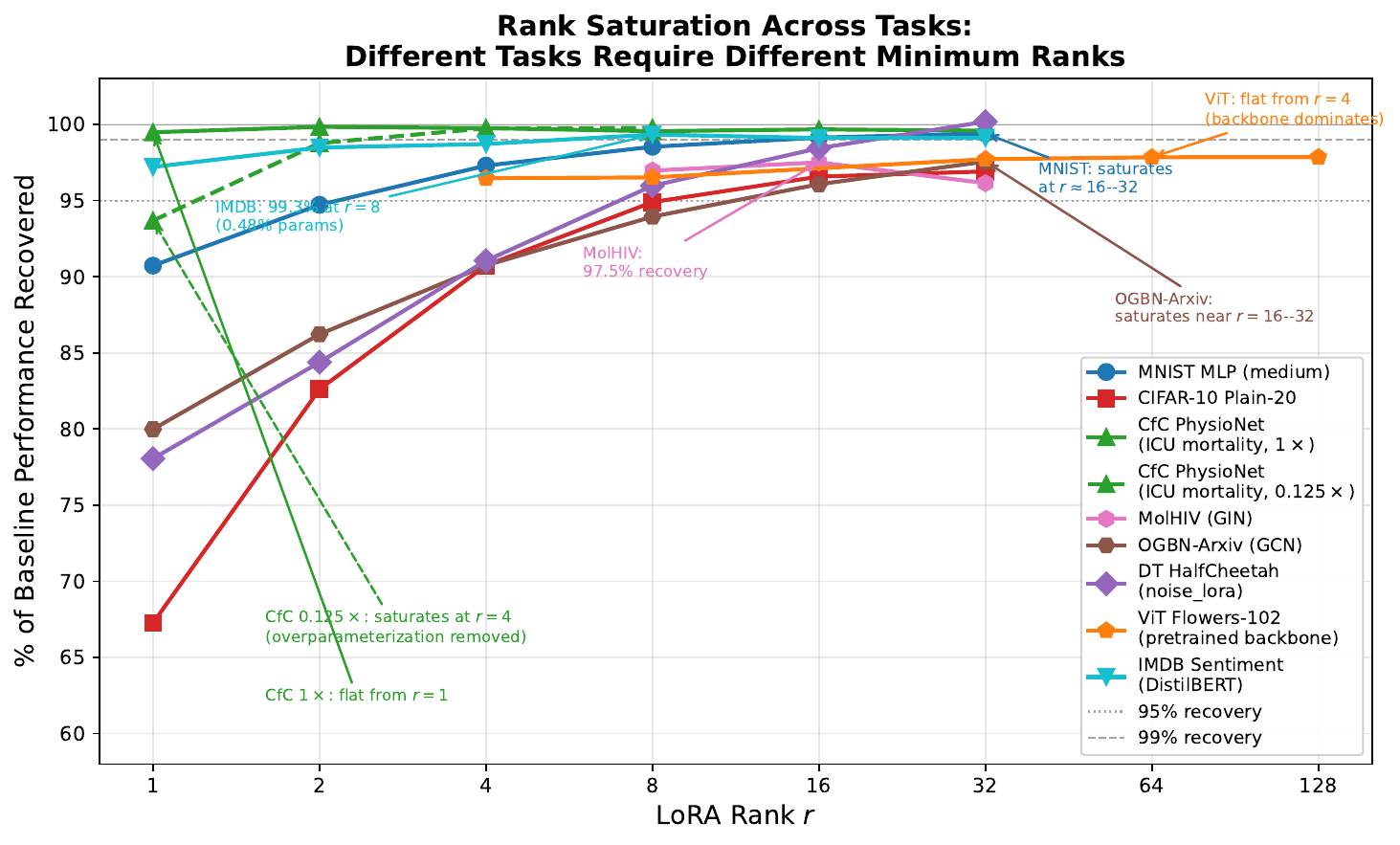}
  \caption{\textbf{Different tasks saturate at different minimum ranks, reflecting intrinsic dimensionality.}
    Each curve shows LottaLoRA
    performance (normalized as \% of fully trained baseline recovered)
    against LoRA rank. CfC PhysioNet (ICU mortality) is flat from $r{=}1$ at the
    published architecture scale; a size-reduction ablation
    (Table~\ref{tab:cfc_size_sweep}) shows that saturation shifts
    rightward when model capacity is reduced, indicating that
    overparameterization contributes to the flat curve;
    MNIST and CIFAR-10 saturate
    between $r{=}8$ and $r{=}32$; Decision Transformer (offline RL) reaches
    100\% recovery near $r{=}16$--$32$; ViT with a pre-trained backbone is
    nearly flat from $r{=}4$, indicating the backbone---not the adapter
    rank---limits performance. MolHIV recovers 97.5\% at $r{=}16$ on
    molecular property prediction.  OGBN-Arxiv (node classification)
    saturates near $r{=}16$--$32$, recovering 98.5\% of baseline.
    IMDB sentiment (DistilBERT) recovers 99.3\% at $r{=}8$ with 0.48\%
    trainable parameters.}
  \label{fig:rank_saturation}
\end{figure}

The experiments point to a striking pattern: the minimum LoRA rank required to solve a task correlates with task complexity. We hypothesize that this minimum rank estimates the task's intrinsic dimensionality---the effective number of degrees of freedom the problem requires \emph{relative to a given architecture}. Crucially, the necessity of the scaffold itself is rank-dependent. At very low ranks, the LoRA factors have limited capacity and benefit from the scaffold's fixed reference frame; at sufficiently high ranks, the adapter can represent the task independently of~$\Wseed$, as the resampling experiments in Section~\ref{sec:q3} suggest. The transition rank---where performance becomes insensitive to the scaffold---marks the point at which the adapter's capacity matches the task's intrinsic complexity. Consider the progression on MNIST (medium preset, 5 seeds): rank~1 reaches
89\%, which is near the performance of a linear classifier. Rank~2 crosses
into the nonlinear regime at 93\%. Rank~4 reaches 96\%, rank~8 reaches
97\%, and by rank~16--32 accuracy saturates at ${\sim}$98\%---within 0.6\,pp
of the fully trained baseline---with only 3.6\% of the parameters.
Figure~\ref{fig:rank_saturation} shows this pattern across eight tasks:
each curve saturates at a different rank, reflecting the task's intrinsic
complexity. Low-dimensional tasks (CfC PhysioNet ICU mortality) saturate
at rank~1 when the architecture is large relative to the task---a
size-reduction ablation confirms that saturation shifts to $r{=}4$ under
a $0.125\times$ model (Table~\ref{tab:cfc_size_sweep});
medium tasks (MNIST, CIFAR-10; Appendix~\ref{app:cifar10})
saturate between ranks~8 and~32; and on tasks where a pre-trained backbone
dominates (ViT Flowers-102), additional rank provides decreasing marginal
gains.

Consider the analogy to PCA: the number of
components needed to reconstruct a signal depends on the intrinsic
dimensionality of the data, not on the ambient dimension (i.e., the total number of parameters in the weight space). Similarly, in LottaLoRA the
minimum rank required to solve a task depends on the complexity of the function the
network must represent, not on the number of parameters in the architecture.
To make this precise, we define the \emph{minimum sufficient rank}
$r^*(\varepsilon)$ as the smallest rank~$r$ such that
\begin{equation}
  \label{eq:min_rank}
  \mathcal{L}_{\mathrm{LottaLoRA}}(r) \;\leq\; \mathcal{L}_{\mathrm{full}} + \varepsilon\,,
\end{equation}
where $\mathcal{L}_{\mathrm{LottaLoRA}}(r)$ is the loss achieved by
LottaLoRA at rank~$r$ and $\mathcal{L}_{\mathrm{full}}$ is the loss of a
fully trained baseline under the same training protocol.
This parallels the PCA criterion of retaining enough components to explain a
$(1{-}\varepsilon)$ fraction of total variance: in both cases, $\varepsilon$
controls the acceptable approximation gap, and the required number of
dimensions reflects the data's intrinsic complexity rather than the ambient
dimension.
The weight matrix $\Wseed$ provides the ambient dimension; the LoRA factors
$B$ and $A$ provide the principal subspace.

This definition connects to the intrinsic dimensionality literature.
Prior work measures the dimensionality of the fine-tuning subspace by
projecting into a random low-dimensional
space~\cite{aghajanyan2021intrinsic} or restricting optimization to random
subspaces~\cite{li2018measuring}.
LottaLoRA offers a complementary perspective: rather than projecting into an
\emph{arbitrary} random subspace, it trains a \emph{structured} low-rank
factorization over a frozen random backbone, so $r^*$ reflects the rank
structure of the task-relevant update rather than the dimensionality of an
unconstrained subspace.
Our results position $r^*$ as a complementary measure of intrinsic dimensionality; establishing formal relationships with these prior measures on shared benchmarks is a natural extension.

The computational and storage advantages of operating at reduced rank---including up to $8\times$ memory reduction and $21\times$ distributable-size compression---are analyzed in Section~\ref{sec:cost}; these hold independent of the minimum-rank hypothesis. The hypothesis itself has two consequences that go beyond cost savings:

\begin{enumerate}[leftmargin=0pt, itemindent=1.5em]
  \item \textbf{Task characterization.} The minimum rank at which LottaLoRA
    approaches full-training performance provides a task-specific measure of intrinsic
    dimensionality. Tasks that require high rank have high-dimensional
    solution spaces; tasks solvable at rank~1 (like Lorenz prediction) have
    low intrinsic dimensionality regardless of the model's parameter count.
  \item \textbf{Interpretability.} Because all task-specific degrees of
    freedom are concentrated in the low-rank factors $\{A_i, B_i\}$, the learned
    function is described by far fewer parameters than in a fully trained
    network. At rank~$r^*$, the number of parameters that must be examined
    to understand the model's behavior scales as $r^*(d_\mathrm{in} +
    d_\mathrm{out})$ per layer rather than $d_\mathrm{in} \times
    d_\mathrm{out}$---a reduction that may enable direct analysis of
    what the network has learned. The low-rank structure imposes a
    bottleneck that is not merely a compression convenience but an
    insight into the functional subspace the task actually requires,
    potentially advancing mechanistic understanding of neural network
    representations.
\end{enumerate}

Finding the minimum rank $r^*(\varepsilon)$ (Equation~\eqref{eq:min_rank}) provides an empirical estimate of task complexity under specific conditions: a fixed architecture family (Transformers, RNNs), a standard optimizer (SGD), and a training budget. Within Transformers, $r^*$ appears robust across initialization families (Section~\ref{sec:q2}), suggesting some architecture-independence. Our experiments suggest $r^*$ captures intrinsic task complexity within a fixed architecture family; characterizing its invariance across fundamentally different architectures (e.g., CNNs, attention-free models), alternative optimizers (Adam, second-order methods), and different training regimes is a natural extension.

\section{Discussion}
\label{sec:discussion}

\begin{figure}[ht]
  \centering
  \includegraphics[width=0.85\linewidth]{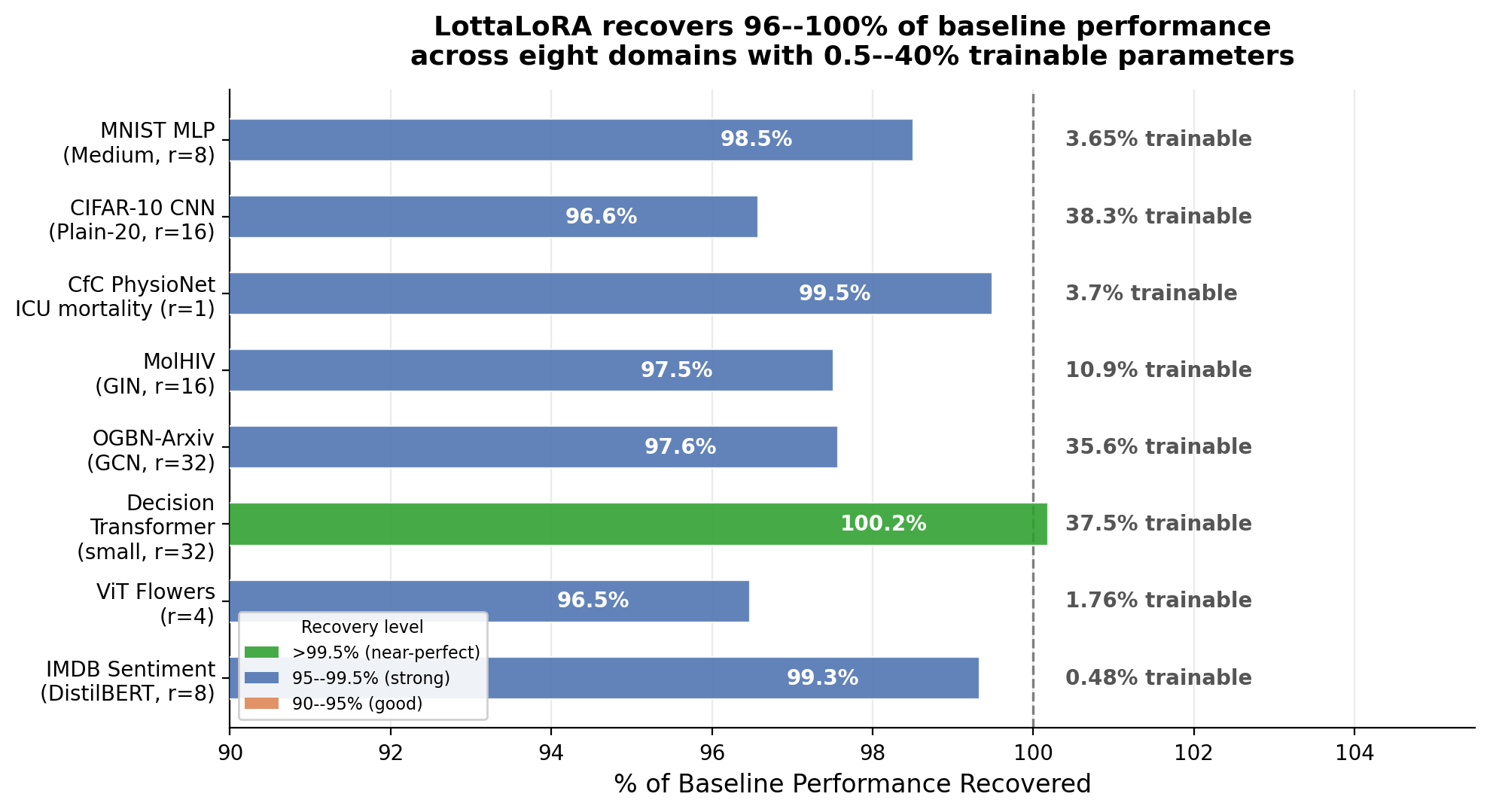}
  \caption{\textbf{LottaLoRA recovers 96--100\% of baseline performance across eight benchmarks.}
    Each bar shows the ratio of LottaLoRA to baseline performance (accuracy,
    $R^2$, or inverted MSE as appropriate); annotations give the trainable
    parameter ratio.  Data from Tables~\ref{tab:mnist_full},
    \ref{tab:cfc_physionet}, \ref{tab:broader_prelim},
    and~\ref{tab:imdb_full}.}
  \label{fig:hero_summary}
\end{figure}

We presented LottaLoRA, a training paradigm in which every backbone weight is
drawn at random and frozen while only low-rank LoRA adapters are trained.
Across nine benchmarks spanning feedforward, recurrent, convolutional, graph,
and Transformer architectures, LottaLoRA recovers 96--100\% of fully trained
performance while training 0.5--40\% of the parameters
(Figure~\ref{fig:hero_summary}). Three mechanistic findings emerged:
(1)~the frozen scaffold is actively exploited ($\beta > 0$ across all
architectures) yet its initialization distribution is
interchangeable---any of 22 tested families, from binary to Gaussian,
yields equivalent performance, provided the specific instance remains
frozen throughout training;
(2)~scaffold stability is necessary---resampling collapses performance by up
to 51\,pp; and (3)~the minimum LoRA rank at which performance saturates
varies by task, providing an empirical estimate of intrinsic dimensionality.
On language modeling at 900\,M parameters, LottaLoRA achieves comparable loss
while training fewer than 0.5\% of the internal parameters, reducing
optimizer-state memory by up to $8\times$ (Section~\ref{sec:cost}).

On WikiText-103, the gap to full training narrows from +0.94\,nats at
300\,M to +0.79\,nats at 900\,M parameters. On other tasks (chaotic
dynamics, time series, graph-level and node-level classification), LottaLoRA
matches or approaches fully trained baselines, while on vision tasks
pre-trained backbones remain substantially advantageous
(Section~\ref{sec:additional}).
These results challenge the assumption that fully optimized weights are
strictly necessary; a random frozen backbone provides a useful computational
substrate whose effectiveness is task and scale dependent.
Three elements support this interpretation:
(1)~the formal equivalence to Reservoir Computing (Section~\ref{sec:rc}),
(2)~the minimum-rank hypothesis (Section~\ref{sec:hypothesis}),
and (3)~systematic evaluation across multiple benchmarks and architectures (Section~\ref{sec:experiments}).

Two results stand out. On the CfC PhysioNet ICU mortality benchmark,
rank-1 adapters recover 99.5\% of the fully trained CfC baseline AUROC
with 3.7\% of its parameters. On OGBG-MolHIV,
LottaLoRA at $r{=}16$ matches the published OGB GIN
baseline~\cite{hu2020ogb} while training only 10.9\% of the parameters,
demonstrating generalization from sequential to graph-level molecular
property prediction; on OGBN-Arxiv, a complementary node-level task, it
recovers 97.6\% of baseline with 35.6\% of parameters
(Section~\ref{sec:additional}). Conversely, on ViT (Flowers-102),
replacing the learned backbone with random noise costs
${\sim}$40\,pp---this quantifies when reservoir quality matters most.

The meta-LoRA collapse (Section~\ref{sec:q3}) provides the sharpest evidence that the scaffold is structurally necessary.
Resampling the scaffold even once per epoch destroys performance---51\,pp on
MNIST---because modifying the reservoir invalidates the reference frame the
readout has learned to exploit~\cite{jaeger2001echo,maass2002real}. The
standard RC remedy---freeze the reservoir, train only the readout---is not a
simplifying assumption but a necessary one, and LottaLoRA inherits the same
constraint for the same reason.

The learned $\beta$ scalars remain strictly positive across all architectures, confirming
that the optimizer actively exploits the frozen backbone rather than silencing it
(Appendix~\ref{app:engineering}).
The adapter works \emph{through} the random weights---selectively amplifying useful
dynamics, much as the Lottery Ticket
Hypothesis~\cite{frankle2019lottery} posits that trainable subnetworks hide
within random initializations; in LottaLoRA, low-rank factorization replaces
binary masking as the selection mechanism. The seed-gated specialization
experiment (Figure~\ref{fig:seed_gating}) provides the strongest evidence:
adapter and scaffold jointly determine the network's function, an instance of
polycomputing~\cite{bongard2023polycomputing}. The two regimes are cleanly
separated by $\beta$: static scaffolds yield $\beta > 0$ (backbone exploited, with
architecture-dependent magnitude; Section~\ref{sec:q3}), resampled
scaffolds drive $\beta \to 0$ (backbone silenced).
This joint determination has a practical deployment consequence: because the
backbone is fully determined by a seed, the adapter and scaffold can be
distributed separately---the LoRA state dictionary released independently of
the reconstruction parameters (seed, distribution specification, framework
version)---enabling polycomputing in practice.
The adapter--scaffold coupling demonstrated by seed-gating resembles the
binding operation central to Hyperdimensional
Computing~\cite{kanerva2009hyperdimensional}: distinct backbone seeds produce
distinct high-dimensional geometries that the same adapter navigates
differently, analogous to how bound hypervectors produce dissimilar
representations recoverable by their components. The hardware implications of
LottaLoRA---tolerance of binary and ternary backbones, compatibility with
device variability in analog crossbar arrays---connect directly to HDC's
motivating use case of noise-tolerant in-memory computing.

This mechanistic picture has a practical consequence.
Because the frozen backbone is fully determined by a random seed,
adapters trained by different users against the same seed are
interchangeable. This enables a distribution paradigm in which a single
published seed defines a universal backbone and task-specific knowledge is
exchanged as compact adapter files---decoupling model distribution from
model training. Unlike standard LoRA, where adapters are tied to a specific
pre-trained checkpoint, LottaLoRA adapters are portable across any
deployment that reconstructs the same backbone from the same seed.

\paragraph{The architectural ceiling and the role of the optimizer.}
Does the residual gap between LottaLoRA and fully trained models reflect
fundamental limits of low-rank steering, or limitations of gradient-based
optimization? Our scaling results show that the gap between LottaLoRA and full training
narrows asymptotically with increasing rank: each additional rank recovers a
decreasing fraction of the remaining deficit. This implies that the
fully trained baseline is the asymptotic ceiling---the maximum performance
the architecture can extract from the data under a given optimizer---and that low-rank adapters
approach it from below with diminishing returns. A striking asymmetry
follows: full training reaches this ceiling by optimizing all~$N^{2}$
parameters, yet a low-rank adapter arrives at essentially the same point
with orders of magnitude fewer degrees of freedom. The surplus parameters in
a fully trained network therefore carry no task-specific function---on
simple tasks like MNIST, the $\Wseed{=}0$ ablation
(Section~\ref{sec:q1}) demonstrates that the task can be encoded entirely
in the low-rank factors. On more complex tasks (e.g., ViT on
Flowers-102), the scaffold's structure provides additional
representational capacity that the adapters exploit, but the
task-specific \emph{signal} still resides in the low-rank factors.

\paragraph{Adapter scaling narrows the gap further.}
Replacing the standard $\alpha/r$ LoRA scaling with rank-stabilized scaling
($\alpha/\!\sqrt{r}$; rsLoRA~\cite{kalajdzievski2024rank}) reduces validation
perplexity of the frozen-backbone configuration by 29\% at rank~32
at the 300\,M scale (from 159.8 to 113.9, three-seed mean $\pm$ std: $0.36$),
while standard scaling shows no improvement beyond rank~8.
DoRA~\cite{liu2024dora} adds only marginal benefit (${\sim}0.4$ perplexity
points) on top of rsLoRA.
These results indicate that a substantial portion of the residual gap is
attributable to suboptimal adapter scaling rather than a fundamental capacity
limit of the frozen backbone, and that the asymptotic ceiling identified above
is closer than standard LoRA training suggests.

\subsection{Limitations}
The present Transformer scaling results at four of five scales represent single-seed runs evaluated on training loss; multi-seed validation-perplexity evaluation at the 300\,M scale confirms the trend, but replication across all scales is needed for a definitive comparison.
The ASIC co-design argument (Section~\ref{sec:cost})---that a fixed random
backbone could be hardwired into silicon---rests on an empirically validated
foundation: binary and 2-bit backbones match full-precision performance
across benchmarks (Section~\ref{sec:precision}), establishing the algorithmic
precondition for fixed-weight acceleration.
The hardware realization itself is beyond the scope of this work; the algorithmic preconditions are established, and wall-clock throughput and energy measurements on dedicated accelerators are a natural next step.

\subsection{Future Directions}
Structured rank allocation across layers, explicit stability analysis of random frozen
dynamics, empirical energy measurements on dedicated hardware, and a formal
characterization of the relationship between minimum rank and task
complexity are all natural next steps.

A deeper question follows naturally from these results: do similar principles govern
computation in biological neural systems? During development, neural circuits
acquire random connectivity through noisy developmental processes; during learning,
specific synaptic pathways are strengthened~\cite{e24060819}. The minimum-rank framework suggests a testable analogy: biological circuits may face similar constraints, with high-dimensional developmental connectivity steered by low-dimensional plastic changes~\cite{thomas2021theoretical,gallego2017neural}.

A fundamental open question is why LottaLoRA consistently approaches but
never exceeds the fully trained baseline. The random backbone provides a
qualitatively different computational substrate; there is no a priori
guarantee that its ceiling must coincide with that of gradient-optimized
weights. One hypothesis is that the fully trained network already saturates
the architecture's representational capacity for a given task, leaving no
room for improvement regardless of the backbone---the ceiling is
architectural, not weight-dependent. An alternative is that gradient-based
optimization of the low-rank factors is itself the bottleneck: the adapter's
loss landscape over a random backbone may contain harder-to-navigate saddle
points or plateaus that prevent the optimizer from finding solutions the
architecture could in principle express. Disentangling architectural
saturation from optimization difficulty is a natural next step, with
implications for both adapter design and our understanding of what fully
trained networks actually learn.

A related finding concerns the relationship between model capacity and
minimum rank. On CfC PhysioNet, the published architecture
(92{,}930~parameters) yields $r^*{=}1$, yet reducing the model to
$0.125\times$ its original size shifts $r^*$ to~4
(Table~\ref{tab:cfc_size_sweep}).
The fully trained baselines themselves improve only from $0.831$ to $0.836$
across a $42\times$ parameter range, confirming substantial
overparameterization. This result indicates that $r^*$ measures task
complexity \emph{relative to a given architecture}: an overparameterized
model compresses the task into a lower-rank subspace than a right-sized one.
Controlling for model capacity is therefore necessary when comparing
$r^*$ across tasks, and a capacity-normalized variant of $r^*$ is a
natural next step.

\subsection{Conclusion}
LottaLoRA demonstrates that the degrees of freedom needed to solve a task
concentrate in a surprisingly small subspace: across nine benchmarks spanning
diverse architecture families,
low-rank adapters over frozen random backbones recover 96--100\% of fully
trained performance while training 0.5--40\% of parameters.
The specific values of most weights serve as scaffolding---structurally
necessary and actively exploited, yet interchangeable---implying that a
network's learned function lives in a low-dimensional subspace whose dimension
reflects the task, not the architecture.
This reframing opens a new perspective: the minimum rank
$r^*(\varepsilon)$ is not merely a compression knob but an empirical measure
of task complexity itself. If this view holds broadly, then the distributable
size of a neural network becomes a statement about the problem it solves,
not an engineering choice---and the experimental framework presented here
provides the empirical tools to make that measurement.

\section*{Acknowledgments}
This publication was made possible through the support of Grant 62212 from the 
John Templeton Foundation and a grant from Templeton World Charity Foundation, Inc. 
We also gratefully acknowledge support provided through a sponsored 
research agreement with Astonishing Labs. The opinions expressed in this publication 
are those of the author(s) and do not necessarily reflect the views of the John Templeton Foundation. 
The authors acknowledge the Tufts University High Performance Compute 
Cluster \url{https://it.tufts.edu/high-performance-computing} which was utilized 
for the research reported in this paper.

\bibliographystyle{unsrt}
\bibliography{references}

\clearpage
\onecolumn
\processdelayedfloats

\clearpage
\appendix

\section*{Use of Large Language Models}
Claude (Opus 4.6, Anthropic) and Gemini (3 Pro, Google) were used as
agentic coding assistants throughout this project: for experiment
orchestration (SLURM script generation, result intake pipelines), data
visualization, manuscript drafting and structural revision, and compilation
of technical details from the codebase into appendix material. These tools
were not used for research ideation or experimental design. All
AI-generated outputs---including code, prose, and technical
specifications---were reviewed, verified, and refined by the authors, who
take full responsibility for the accuracy and integrity of this work.

\section{Transformer Quantitative Comparison}
\label{app:quantitative}

All five Transformer scales use a LlamaForCausalLM architecture with
RMSNorm, rotary positional embeddings (RoPE), SiLU activations, and tied
word embeddings (vocabulary size 32{,}000). Table~\ref{tab:arch_specs}
specifies the architecture at each scale.

\begin{table}[ht]
  \caption{Transformer architecture specifications for the WikiText-103
    scaling experiments. All scales share the same tokenizer
    (TinyLlama, 32{,}000 vocabulary), RMSNorm, and RoPE.}
  \label{tab:arch_specs}
  \centering
  \begin{tabular}{lrrrrr}
    \toprule
    Scale & Layers & Hidden dim & Heads & MLP dim & Total params \\
    \midrule
    3\,M   &  6 &    64 &  4 &   192 &   2.4\,M \\
    30\,M  & 10 &   384 &  6 & 1{,}024 &  30.0\,M \\
    300\,M & 22 & 1{,}024 & 16 & 2{,}816 & 315.4\,M \\
    600\,M & 30 & 1{,}344 & 21 & 3{,}584 & 693.4\,M \\
    900\,M & 34 & 1{,}664 & 26 & 4{,}608 & 1{,}212.0\,M \\
    \bottomrule
  \end{tabular}
\end{table}

\noindent Training uses AdamW (learning rate $10^{-4}$, weight decay 0.1) with
cosine learning rate schedule and 200-step warmup. LoRA adapters (ranks
$r \in \{1, 2, 4, 8\}$, $\alpha{=}16$, dropout 0.0) are applied to all
attention projections ($W_Q$, $W_K$, $W_V$, $W_O$); the backbone is frozen
with a trainable per-layer $\beta$ scalar. Batch size is 32 with 128-token
blocks; each run trains for one epoch (34{,}429 steps).

Table~\ref{tab:quant} reports final training loss and trainable parameter
counts for every configuration.

\begin{table}[ht]
  \caption{Final training loss on WikiText-103 and parameter counts for full training and
    LottaLoRA across all backbone scales and ranks. ``Internal'' denotes
    transformer-internal trainable parameters. Ratio is the internal count as
    a percentage of the same-scale full baseline.}
  \label{tab:quant}
  \centering
  \small
  \begin{tabular}{llrrrr}
    \toprule
    Method & Rank & Total params & Internal trainable & Ratio (\%) & Final loss \\
    \midrule
    Full (3M)        & --  & 2,368,320     & 320,320        & 100.00 & 5.007 \\
    LottaLoRA (3M)    & 1   & 2,371,416     & 3,096          & 0.97   & 5.855 \\
    LottaLoRA (3M)    & 2   & 2,374,488     & 6,168          & 1.93   & 5.781 \\
    LottaLoRA (3M)    & 4   & 2,380,632     & 12,312         & 3.84   & 5.682 \\
    LottaLoRA (3M)    & 8   & 2,392,920     & 24,600         & 7.68   & 5.639 \\
    \midrule
    Full (30M)       & --  & 29,990,784    & 17,702,784   & 100.00 & 3.757 \\
    LottaLoRA (30M)   & 1   & 30,021,544    & 30,760         & 0.17   & 4.870 \\
    LottaLoRA (30M)   & 2   & 30,052,264    & 61,480         & 0.35   & 4.798 \\
    LottaLoRA (30M)   & 4   & 30,113,704    & 122,920        & 0.69   & 4.728 \\
    LottaLoRA (30M)   & 8   & 30,236,584    & 245,800        & 1.39   & 4.681 \\
    \midrule
    Full (300M)      & --  & 315,405,312   & 282,637,312  & 100.00 & 3.252 \\
    LottaLoRA (300M)  & 1   & 315,585,624   & 180,312        & 0.06   & 4.368 \\
    LottaLoRA (300M)  & 2   & 315,765,848   & 360,536        & 0.13   & 4.286 \\
    LottaLoRA (300M)  & 4   & 316,126,296   & 720,984        & 0.26   & 4.227 \\
    LottaLoRA (300M)  & 8   & 316,847,192   & 1,441,880      & 0.51   & 4.191 \\
    \midrule
    Full (600M)      & --  & 693,370,944   & 650,362,944  & 100.00 & 3.185 \\
    LottaLoRA (600M)  & 1   & 693,693,624   & 322,680        & 0.05   & 4.229 \\
    LottaLoRA (600M)  & 2   & 694,016,184   & 645,240        & 0.10   & 4.146 \\
    LottaLoRA (600M)  & 4   & 694,661,304   & 1,290,360      & 0.20   & 4.099 \\
    LottaLoRA (600M)  & 8   & 695,951,544   & 2,580,600      & 0.40   & 4.058 \\
    \midrule
    Full (900M)      & --  & 1,212,039,296 & 1,158,791,296 & 100.00 & 3.156 \\
    LottaLoRA (900M)  & 1   & 1,212,492,040 & 452,744       & 0.04   & 4.160 \\
    LottaLoRA (900M)  & 2   & 1,212,944,648 & 905,352       & 0.08   & 4.096 \\
    LottaLoRA (900M)  & 4   & 1,213,849,864 & 1,810,568     & 0.16   & 3.989 \\
    LottaLoRA (900M)  & 8   & 1,215,660,296 & 3,621,000     & 0.31   & 3.950 \\
    \bottomrule
  \end{tabular}
\end{table}

\section{MNIST Full Results}
\label{app:mnist}

\paragraph{Training configuration.}
All MNIST experiments use AdamW (learning rate $10^{-3}$, weight decay
$10^{-2}$), batch size 128, 20 epochs, and cosine annealing learning rate
schedule. Input images ($28 \times 28$) are normalized (mean $= 0.1307$,
std $= 0.3081$) and flattened. The train set is split 90/10 into
training/validation.

\paragraph{Architecture presets.}
Four fully connected architectures are tested, all with ReLU activations
and dropout 0.1:
\textbf{tiny}: 2 hidden layers ($d = 128, 64$);
\textbf{small}: 3 layers ($d = 256, 128, 64$);
\textbf{medium}: 4 layers ($d = 512, 256, 128, 64$);
\textbf{large}: 5 layers ($d = 1024, 512, 256, 128, 64$).
Input dimension is $784$ (flattened $28 \times 28$); output is a 10-class
softmax.

\paragraph{$\Wseed$ initialization families (22 total).}
Table~\ref{tab:w0_families} lists all 22 initialization families with
their definitions and default parameters. All families use fan-in scaling
($\sigma = 1/\sqrt{d_\mathrm{in}}$) unless the family defines its own
variance schedule (e.g., Kaiming, Xavier).

\begin{table}[ht]
  \caption{The 22~$\Wseed$ initialization families used in the distribution
    robustness experiment (Section~\ref{sec:q2}).}
  \label{tab:w0_families}
  \centering
  \small
  \begin{tabular}{lll}
    \toprule
    Family & Distribution & Key parameters \\
    \midrule
    \multicolumn{3}{l}{\textit{Standard}} \\
    Normal          & $\mathcal{N}(0, \sigma^2)$ & $\sigma = 1.0$ \\
    Trunc.\ Normal  & $\mathcal{N}(0, \sigma^2)$ clipped at $\pm 2\sigma$ & $\sigma = 1.0$ \\
    Uniform         & $\mathcal{U}(-a, a)$ & $a = 0.1$ \\
    Orthogonal      & Orthogonal matrix & gain $= 1.0$ \\
    \midrule
    \multicolumn{3}{l}{\textit{Scale-aware}} \\
    Kaiming Normal  & $\mathcal{N}(0, 2/\text{fan\_in}(1+a^2))$ & $a = \sqrt{5}$ \\
    Kaiming Uniform & $\mathcal{U}(-b, b)$, $b = \sqrt{6/\text{fan\_in}(1+a^2)}$ & $a = \sqrt{5}$ \\
    Xavier Normal   & $\mathcal{N}(0, 2/(\text{fan\_in}+\text{fan\_out}))$ & gain $= 1.0$ \\
    Xavier Uniform  & $\mathcal{U}(-b, b)$, $b = \sqrt{6/(\text{fan\_in}+\text{fan\_out})}$ & gain $= 1.0$ \\
    Spectral Radius & $\mathcal{N}(0, 1)$ rescaled to $\rho_{\max} = \rho$ & $\rho = 0.95$ \\
    \midrule
    \multicolumn{3}{l}{\textit{Heavy-tailed}} \\
    Cauchy          & $\mathrm{Cauchy}(0, s)$, clipped at $\pm 10s$ & $s = 0.1$ \\
    Laplace         & $\mathrm{Laplace}(0, b)$ & $b = 0.1$ \\
    Student-$t$     & $t_\nu$ & $\nu = 3$ \\
    \midrule
    \multicolumn{3}{l}{\textit{Mixture / Sparse}} \\
    Gaussian Mixture & $0.9\,\mathcal{N}(0, 0.05^2) + 0.1\,\mathcal{N}(0, 0.5^2)$ & --- \\
    Sparse Normal   & $(1-p)\,\mathcal{N}(0, \sigma^2) + p\,\delta_0$ & $p = 0.2$ \\
    Sparse Erd\H{o}s-R\'{e}nyi & Same as sparse normal & $p = 0.2$ \\
    \midrule
    \multicolumn{3}{l}{\textit{Other distributions}} \\
    Beta            & $\mathrm{Beta}(\alpha, \beta)$ scaled to $[-0.1, 0.1]$ & $\alpha = \beta = 2$ \\
    Exponential     & $\mathrm{Exp}(\lambda) - 1/\lambda$ (centered) & $\lambda = 10$ \\
    \midrule
    \multicolumn{3}{l}{\textit{Low-bit quantized}} \\
    16-bit          & $\mathcal{N}(0, 1)$ quantized to $2^{16}$ levels & --- \\
    8-bit           & $\mathcal{N}(0, 1)$ quantized to $2^{8}$ levels & --- \\
    4-bit           & $\mathcal{N}(0, 1)$ quantized to $2^{4}$ levels & --- \\
    2-bit           & $\mathcal{N}(0, 1)$ quantized to $2^{2}$ levels & --- \\
    Binary          & $\mathrm{sign}(\mathcal{N}(0, 1)) \in \{-1, +1\}$ & --- \\
    \bottomrule
  \end{tabular}
\end{table}

\subsection{Full vs LottaLoRA rank sweep (30 runs)}

\begin{table}[ht]
  \caption{MNIST full-vs-LottaLoRA comparison across three presets and
    multiple ranks. Medium preset values are means over 2 runs;
    tiny and small are from 2 runs (seed 42, normal $\Wseed$, scale 0.1).}
  \label{tab:mnist_full}
  \centering
  \small
  \begin{tabular}{lrrrr}
    \toprule
    Method & Test Acc (\%) & Trainable & Ratio (\%) \\
    \midrule
    Full\_tiny     & 98.06 & 109,386 & 100.00 \\
    Full\_small    & 98.37 & 242,762 & 100.00 \\
    Full\_medium   & 98.38 & 575,050 & 100.00 \\
    \midrule
    LottaLoRA\_tiny\_r1   & 85.06 & 1,756  & 1.49 \\
    LottaLoRA\_tiny\_r2   & 90.60 & 2,860  & 2.42 \\
    LottaLoRA\_tiny\_r4   & 93.56 & 5,068  & 4.28 \\
    LottaLoRA\_tiny\_r8   & 95.41 & 9,484  & 8.02 \\
    \midrule
    LottaLoRA\_small\_r1  & 87.51 & 2,269  & 0.93 \\
    LottaLoRA\_small\_r8  & 96.19 & 13,581 & 5.31 \\
    \midrule
    LottaLoRA\_medium\_r1  & 89.33 & 3,294  & 0.57 \\
    LottaLoRA\_medium\_r2  & 92.73 & 5,934  & 1.03 \\
    LottaLoRA\_medium\_r4  & 95.44 & 11,214 & 1.95 \\
    LottaLoRA\_medium\_r8  & 96.80 & 21,774 & 3.65 \\
    LottaLoRA\_medium\_r16 & 97.53 & 42,894 & 7.46 \\
    LottaLoRA\_medium\_r32 & 97.80 & 85,134 & 14.80 \\
    \bottomrule
  \end{tabular}
\end{table}

The four architecture presets differ in depth and width: \textbf{tiny} uses two hidden layers
($d{=}128, 64$), \textbf{small} three layers ($d{=}256, 128, 64$), \textbf{medium} four layers
($d{=}512, 256, 128, 64$), and \textbf{large} five layers ($d{=}1024, 512, 256, 128, 64$),
all with ReLU activations and dropout 0.1.
Gap shrinks with model size at fixed rank: $-2.65$\,pp (tiny) to $-1.58$\,pp (medium) at $r{=}8$,
consistent with the control-theoretic prediction that larger reservoirs are easier to steer
via low-rank feedback.

\subsection{$\Wseed$ distribution robustness (2{,}640 runs, 22 families)}

\begin{figure}[ht]
  \centering
  \begin{minipage}[t]{0.48\linewidth}
    \centering
    \includegraphics[width=\linewidth]{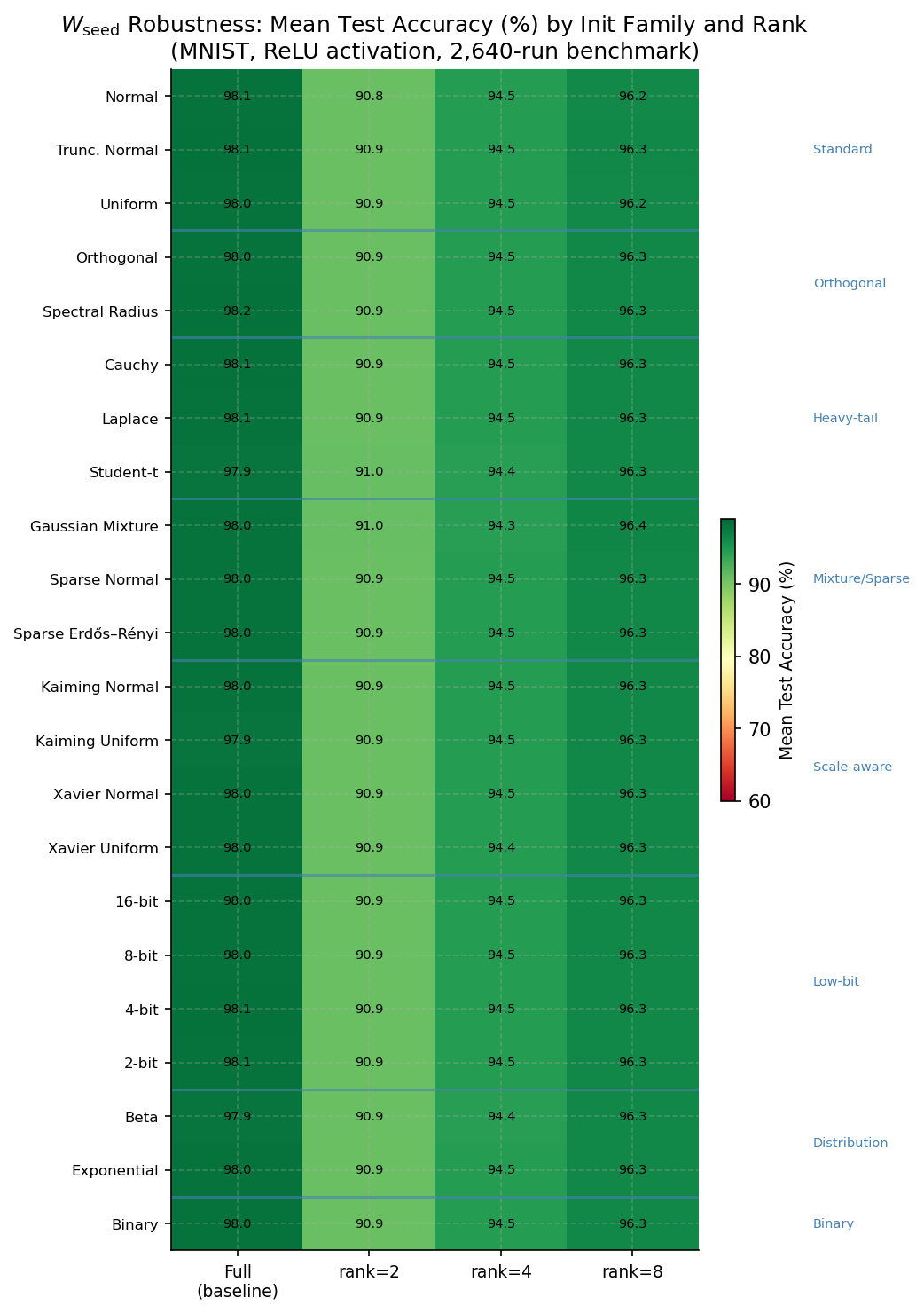}
  \end{minipage}\hfill
  \begin{minipage}[t]{0.48\linewidth}
    \centering
    \includegraphics[width=\linewidth]{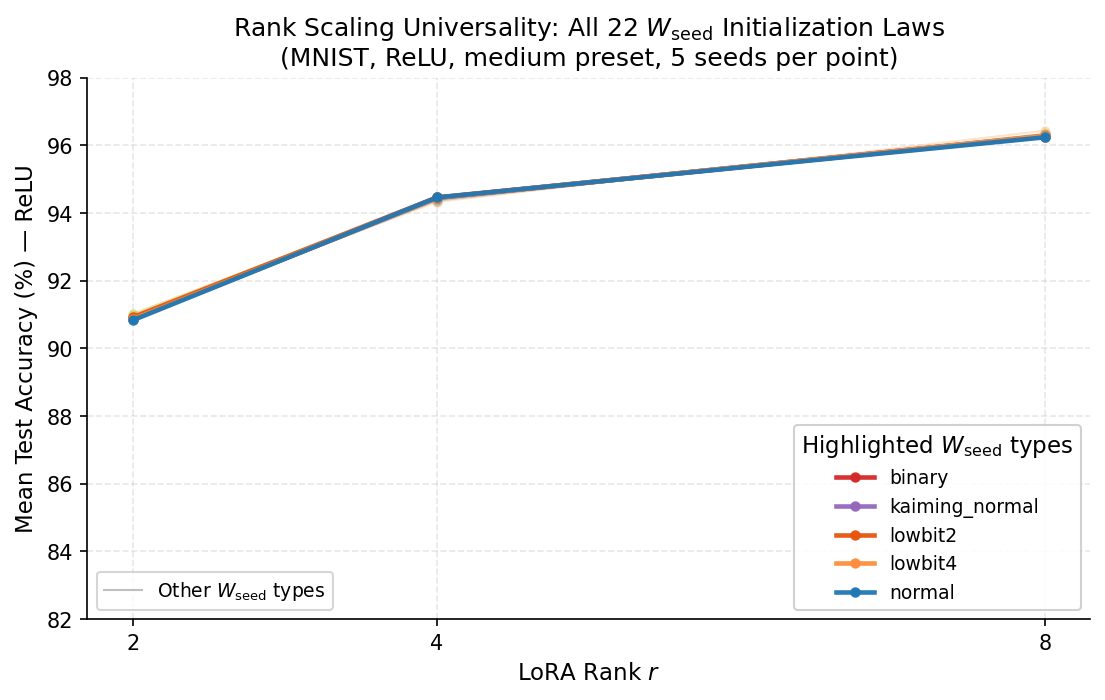}
  \end{minipage}
  \caption{\textbf{Every $\Wseed$ initialization family exceeds 95\% at rank~8 on MNIST.}
    \textbf{Left:} Every initialization family exceeds 95\% at rank~8,
    showing that the scaffold's specific distribution has negligible effect.
    \textbf{Right:} All 22 families converge tightly as rank increases,
    confirming that the scaffold is interchangeable on this task.}
  \label{fig:w0_robustness}
\end{figure}

Figure~\ref{fig:w0_robustness} reproduces the full heatmap and
rank-scaling curves for all 22 $\Wseed$ initialization families summarized in
Table~\ref{tab:w0_robust} of Section~\ref{sec:q2}. Each family was tested across four
LoRA ranks ($r \in \{2,4,8,\text{full}\}$) using the \textbf{medium} preset
(four hidden layers: $d{=}512, 256, 128, 64$), 5 seeds per family (120 runs per rank,
2{,}640 total).

\subsection{LoRA B-matrix initialization (32 runs)}
\label{app:mnist_binit}

Two B-matrix initialization strategies are compared: \textbf{zeros} (standard,
preserving the pre-trained trajectory at initialization) and
\textbf{kaiming\_uniform} (breaking symmetry).

\begin{table}[ht]
  \caption{MNIST test accuracy (\%) by LoRA B-init and rank (seed 42).
    Zeros dominates at rank~1; the gap closes at higher ranks.}
  \label{tab:mnist_binit}
  \centering
  \small
  \begin{tabular}{llrrrr}
    \toprule
    Preset & B-init & $r{=}1$ & $r{=}2$ & $r{=}4$ & $r{=}8$ \\
    \midrule
    tiny   & zeros   & 86.47 & 90.38 & 93.47 & 95.65 \\
    tiny   & kaiming & 84.12 & 90.03 & 92.73 & 95.53 \\
    \midrule
    small  & zeros   & 87.68 & 91.07 & 94.64 & 96.24 \\
    small  & kaiming & 75.23 & 89.93 & 93.27 & 96.03 \\
    \midrule
    medium & zeros   & 89.40 & 92.92 & 95.74 & 96.94 \\
    medium & kaiming & 70.59 & 90.19 & 94.38 & 96.96 \\
    \midrule
    large  & zeros   & 92.08 & 94.83 & 96.70 & 97.37 \\
    large  & kaiming & 61.31 & 92.10 & 95.73 & 97.29 \\
    \bottomrule
  \end{tabular}
\end{table}

Table~\ref{tab:mnist_binit} shows that at rank~1, kaiming causes severe underfitting that worsens with model size
($-2.35$\,pp for tiny, $-30.77$\,pp for large). By rank~8, both
initializations converge to within 0.2\,pp.  Under 3-seed robust evaluation
(varying the reservoir seed), kaiming shows \emph{higher} mean accuracy at
rank~8 (e.g.\ tiny: 85.2\% vs 60.0\%), suggesting that symmetry-breaking at
initialization produces solutions less dependent on the specific reservoir
realization.

\subsection{Output head configuration (48 runs)}
\label{app:mnist_output}

Three output-layer configurations are compared: \textbf{full} (standard
trainable linear head, 650~params for 10 classes), \textbf{lora} (LoRA-parameterized
output, 0~output params), and \textbf{lora\_bias} (LoRA output with 10~trainable bias
terms).

\begin{table}[ht]
  \caption{MNIST test accuracy (\%) by output head mode and rank (seed 42).
    Full output dominates at low rank; all modes converge by rank~4.}
  \label{tab:mnist_output}
  \centering
  \small
  \begin{tabular}{llrrrr}
    \toprule
    Preset & Output & $r{=}1$ & $r{=}2$ & $r{=}4$ & $r{=}8$ \\
    \midrule
    tiny   & full      & 86.46 & 90.38 & 93.47 & 95.65 \\
    tiny   & lora      & 75.74 & 87.96 & 92.70 & 95.40 \\
    tiny   & lora+bias & 77.83 & 88.30 & 92.90 & 95.47 \\
    \midrule
    small  & full      & 87.68 & 91.07 & 94.64 & 96.26 \\
    small  & lora      & 83.20 & 90.15 & 94.27 & 96.21 \\
    small  & lora+bias & 85.06 & 90.92 & 94.24 & 96.27 \\
    \midrule
    medium & full      & 89.40 & 92.92 & 95.68 & 96.94 \\
    medium & lora      & 87.08 & 93.01 & 95.09 & 97.24 \\
    medium & lora+bias & 88.65 & 93.10 & 94.75 & 97.06 \\
    \midrule
    large  & full      & 92.16 & 94.83 & 96.70 & 97.37 \\
    large  & lora      & 74.05 & 94.84 & 96.61 & 97.37 \\
    large  & lora+bias & 69.25 & 94.92 & 96.32 & 97.40 \\
    \bottomrule
  \end{tabular}
\end{table}

Table~\ref{tab:mnist_output} shows that at rank~1, replacing the full output head with a LoRA-parameterized version
costs 4--18\,pp (large model: catastrophic drop from 92.2\% to 74.1\% for
lora, 69.3\% for lora+bias). At rank~$\geq$4, all modes converge within
${\sim}1$\,pp. The parameter savings are modest (${\sim}650$ params, 3--7\%
of the trainable budget at rank~4), so the full output head remains the
recommended default.

Under robust evaluation, the full head also maintains significantly better
reservoir robustness at rank~$\geq$4 (tiny $r{=}8$: 60.0\% vs 32.1\%/26.7\%),
suggesting the standard output head helps generalize across reservoir
realizations.

\section{CfC Reservoir on PhysioNet 2012 ICU Mortality}
\label{app:cfc_physionet}

\paragraph{Dataset.}
PhysioNet 2012~\cite{silva2012predicting} is a real-world ICU benchmark
derived from the MIMIC-II database.  Each record consists of 37~clinical
variables (vital signs and lab measurements) recorded during the first
48~hours of an ICU stay; the prediction target is in-hospital mortality
(binary; positive rate ${\approx}8\%$, class imbalance 11.7:1).  We use
3{,}200~train / 800~test records.

\paragraph{Architecture.}
The backbone is a Closed-form Continuous-time network
(CfC~\cite{hasani2022closed}) with hidden size~256 and 2~backbone layers
of 64~units each (backbone parameter count: 91{,}264).  A linear readout maps the
final hidden state to a scalar logit.  For LottaLoRA, the backbone
weights are frozen at random initialization; LoRA adapters ($\Delta W =
BA$, rank~$r$) are inserted into each recurrent layer.  The readout and
learnable scalar $\beta$ are the only additional trainable parameters.

\paragraph{Training.}
All runs use AdamW (base learning rate $2{\times}10^{-3}$, exponential
decay $0.9$ per epoch, weight decay $4{\times}10^{-6}$, batch size~128)
for a fixed 100~epochs.  Positive-class weighting of~$11.69$ compensates
for the class imbalance.  Seeds: 42, 100, 123, 200, 300 (5~per rank).

\begin{figure}[ht]
  \centering
  \includegraphics[width=0.85\linewidth]{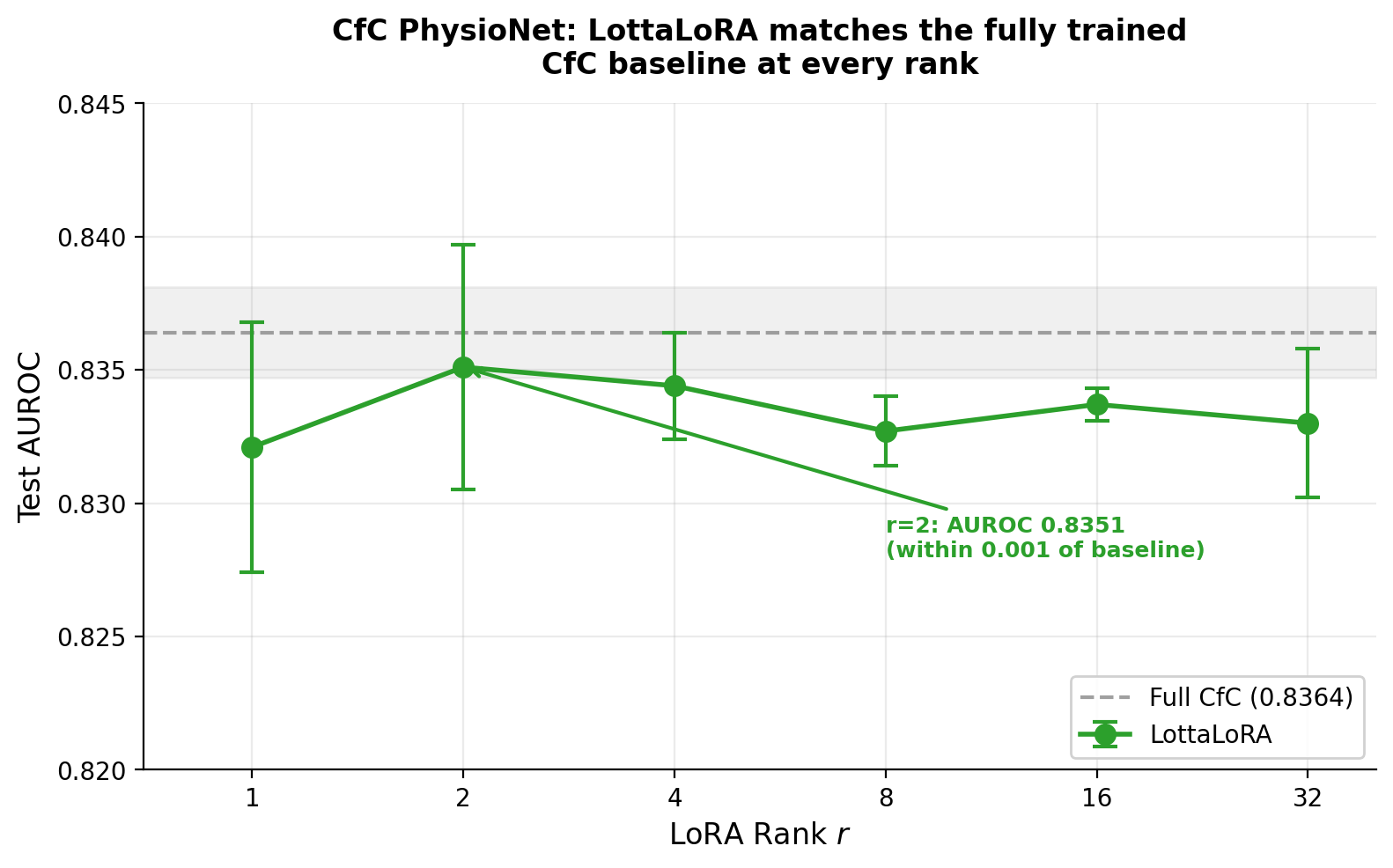}
  \caption{\textbf{LottaLoRA matches the fully trained CfC baseline on
    PhysioNet 2012 ICU mortality at every rank.}  Mean $\pm$ std AUROC
    over 5~seeds at ranks 1--32.  Dashed line: Full CfC baseline
    (AUROC~$= 0.836$).}
  \label{fig:cfc_physionet_main}
\end{figure}

\paragraph{Rank sensitivity.}
AUROC is flat from rank~1 onward (Table~\ref{tab:cfc_physionet_rank},
Figure~\ref{fig:cfc_physionet_main}); the gap between the best LottaLoRA
configuration ($r{=}2$, AUROC~$0.835$) and the full baseline
($0.836$) is $0.001$---within one standard deviation of either estimate.

\begin{table}[ht]
  \caption{CfC PhysioNet rank sensitivity: mean AUROC $\pm$ std over
    5~seeds.  \%Trainable is relative to the Full CfC parameter count
    (92{,}930).}
  \label{tab:cfc_physionet_rank}
  \centering
  \small
  \begin{tabular}{rccc}
    \toprule
    Rank & AUROC & \#Trainable params & \%Trainable \\
    \midrule
     1   & $0.8321 \pm 0.0047$ &  3{,}482 &  3.7\% \\
     2   & $0.8351 \pm 0.0046$ &  5{,}292 &  5.7\% \\
     4   & $0.8344 \pm 0.0020$ &  8{,}912 &  9.6\% \\
     8   & $0.8327 \pm 0.0013$ & 16{,}152 & 17.4\% \\
    16   & $0.8337 \pm 0.0006$ & 30{,}632 & 25.1\% \\
    32   & $0.8330 \pm 0.0028$ & 59{,}592 & 39.5\% \\
    \midrule
    Full CfC & $0.8364 \pm 0.0017$ & 92{,}930 & 100\% \\
    \bottomrule
  \end{tabular}
\end{table}

\section{OGBG-MolHIV Molecular Property Prediction}
\label{app:molhiv}

\paragraph{Setup.}
We use a Graph Isomorphism Network (GIN)~\cite{xu2019gin} with
GINEConv layers~\cite{hu2020ogb} that encode both atom and bond features,
hidden dimension 300, 5 layers, dropout 0.5, and skip connections for
layers 1--4. The standard OGB scaffold split is used (binary
classification: HIV activity prediction, 41{,}127 molecular graphs).
Training uses Adam (learning rate $1 \times 10^{-3}$) for 300 epochs with
batch size 64 and gradient clipping (norm 1.0). Class imbalance
(${\sim}4{:}1$ negative-to-positive ratio) is handled via
\texttt{pos\_weight} scaling in the binary cross-entropy loss.
In the LottaLoRA condition, all GIN MLP linear layers are frozen and
replaced with $W_\mathrm{eff} = \beta\,\Wseed + BA$; $\Wseed$ is
initialized with fan-in-scaled Gaussian noise
($\sigma = 1/\sqrt{d_\mathrm{in}} \approx 0.058$). The classification
head remains a plain \texttt{nn.Linear(300, 1)}, always fully trainable.
We sweep ranks $r \in \{8, 16, 32\}$. Results are averaged over 5 seeds
(42, 100, 123, 456, 789).

\paragraph{Results.}
Table~\ref{tab:molhiv_full} reports the full results.
The fully trained baseline achieves ROC-AUC $0.7755 \pm 0.0060$,
which closely matches the published OGB GIN+virtual-node result
($0.7707 \pm 0.0149$~\cite{hu2020ogb}), validating our implementation
despite not using a virtual node. LottaLoRA at $r{=}16$ achieves
$0.7562 \pm 0.0158$, recovering 97.5\% of baseline while training only
10.9\% of the parameters (203\,K of 1.86\,M). This score matches the
published OGB GIN baseline ($0.7558 \pm 0.0140$~\cite{hu2020ogb}),
demonstrating that a frozen-backbone GIN with low-rank adapters performs
on par with a fully trained GIN.

\begin{table}[ht]
  \caption{\textbf{LottaLoRA matches the published OGB GIN baseline on
    OGBG-MolHIV while training 10.9\% of the parameters.} Test ROC-AUC
    (5 seeds). Published GIN baseline:
    $0.7558 \pm 0.0140$~\cite{hu2020ogb}.}
  \label{tab:molhiv_full}
  \centering
  \small
  \begin{tabular}{lcccc}
    \toprule
    Method & Rank & Mean ROC-AUC & Std & Trainable \\
    \midrule
    Baseline (fully trained) & --- & 0.7755 & 0.0060 & 1{,}863{,}906 (100\%) \\
    LottaLoRA & $r{=}8$  & 0.7520 & 0.0194 & 131{,}416 (7.1\%) \\
    LottaLoRA & $r{=}16$ & \textbf{0.7562} & 0.0158 & 203{,}416 (10.9\%) \\
    LottaLoRA & $r{=}32$ & 0.7456 & 0.0152 & 347{,}416 (18.6\%) \\
    \bottomrule
  \end{tabular}
\end{table}

\begin{figure}[ht]
  \centering
  \includegraphics[width=0.75\linewidth]{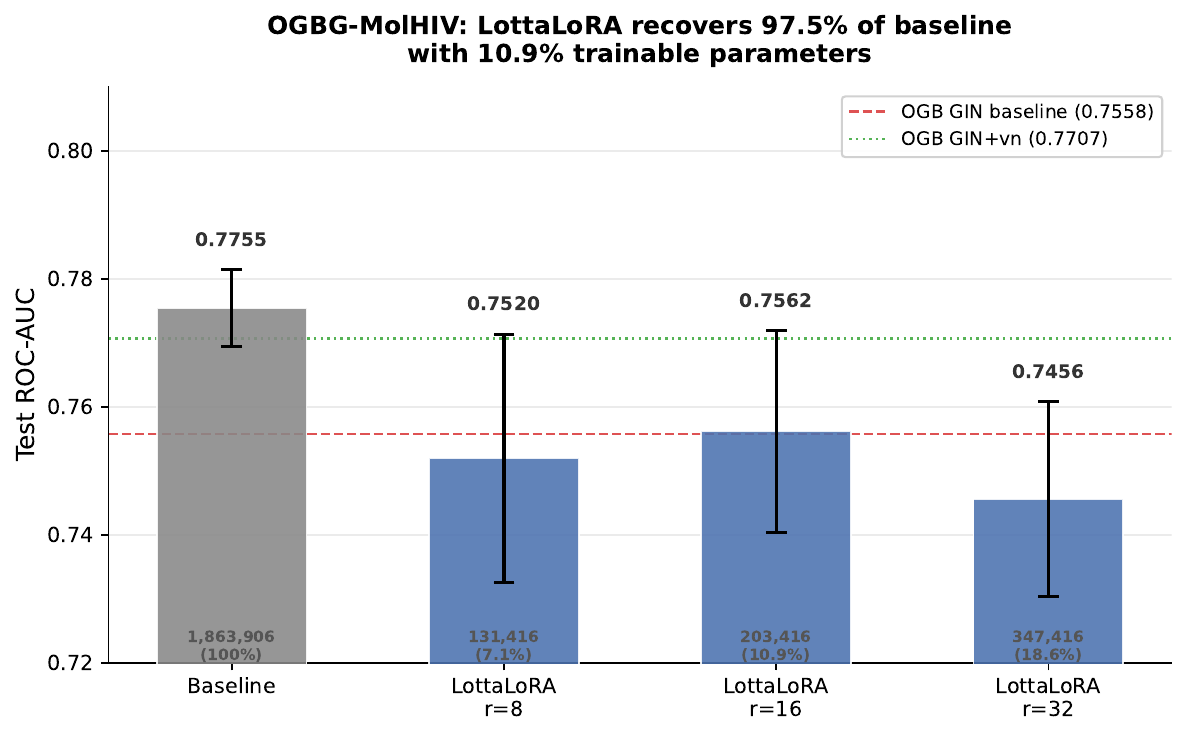}
  \caption{\textbf{LottaLoRA recovers 97.5\% of baseline ROC-AUC on
    molecular property prediction.} Test ROC-AUC on OGBG-MolHIV (5 seeds,
    error bars show $\pm 1$ std). Dashed lines mark published OGB
    baselines~\cite{hu2020ogb}: GIN (red) and GIN+virtual node (green).}
  \label{fig:molhiv}
\end{figure}

Figure~\ref{fig:molhiv} visualizes these results alongside the published
OGB baselines. Rank $r{=}16$ is the sweet spot: $r{=}8$ slightly
underperforms due to limited adapter capacity, while $r{=}32$ shows
decreasing marginal gains---and slight degradation, likely from overfitting the
small positive class
(${\sim}$20\% of training graphs). This non-monotonic pattern is consistent
with the minimum-rank hypothesis (Section~\ref{sec:hypothesis}): the
intrinsic dimensionality of molecular HIV activity prediction lies near
$r \approx 16$.

\section{OGBN-Arxiv Node Classification}
\label{app:ogbn_arxiv}

\paragraph{Setup.}
We replicate the exact OGB GCN baseline protocol~\cite{hu2020ogb}:
a Graph Convolutional Network (GCN)~\cite{kipf2017gcn} with 3 layers,
hidden dimension 256, BatchNorm, dropout 0.5, and the standard OGB
temporal split (40-class node classification on 169{,}343 nodes from
a citation graph). Training uses Adam (learning rate $1 \times 10^{-2}$,
weight decay 0) for 500 epochs. The published reference accuracy is
$71.74 \pm 0.29$\%.
In the LottaLoRA condition, all GCN linear layers are frozen and
replaced with $W_\mathrm{eff} = \beta\,\Wseed + BA$; $\Wseed$ is
initialized with fan-in-scaled Gaussian noise. The classification
head remains fully trainable. We sweep ranks
$r \in \{1, 2, 4, 8, 16, 32\}$. Results are averaged over 10 seeds.

\begin{figure}[ht]
  \centering
  \includegraphics[width=0.75\linewidth]{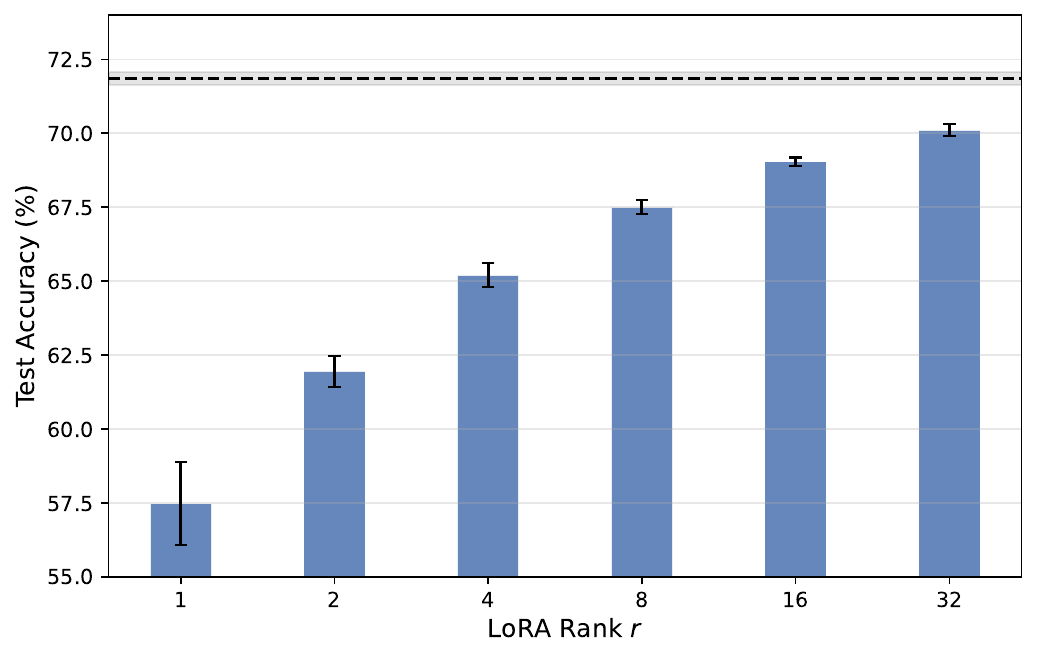}
  \caption{\textbf{LottaLoRA recovers 97.6\% of baseline GCN accuracy on
    OGBN-Arxiv node classification.} Test accuracy (10 seeds,
    error bars show $\pm 1$ std). Dashed line marks the fully trained
    GCN baseline ($71.86\%$).}
  \label{fig:ogbn_arxiv}
\end{figure}

\paragraph{Results.}
Table~\ref{tab:ogbn_arxiv_full} reports the full results.
Our fully trained baseline achieves $71.86 \pm 0.22$\% test accuracy,
closely matching the published OGB GCN baseline
($71.74 \pm 0.29$\%~\cite{hu2020ogb}), validating our reproduction.
LottaLoRA at $r{=}32$ achieves $70.11 \pm 0.21$\%, recovering 97.6\%
of baseline while training only 39{,}171 parameters (35.6\% of the
baseline's 110{,}120). Rank~1 degrades substantially (${\sim}57\%$),
consistent with the minimum-rank hypothesis: 40-class node
classification requires more adapter capacity than binary molecular
classification (MolHIV).
Figure~\ref{fig:ogbn_arxiv} visualizes the rank sweep.

\begin{table}[ht]
  \caption{\textbf{LottaLoRA recovers 97.6\% of baseline GCN accuracy on
    OGBN-Arxiv while training 35.6\% of the parameters.}
    Test accuracy (10 seeds). Published OGB GCN baseline:
    $71.74 \pm 0.29$\%~\cite{hu2020ogb}.}
  \label{tab:ogbn_arxiv_full}
  \centering
  \small
  \begin{tabular}{cccc}
    \toprule
    Rank & Acc (\%) & Std & \%\,Trainable \\
    \midrule
    Baseline & 71.86 & 0.22 & 100\% \\
    \midrule
    $r{=}32$ & \textbf{70.11} & 0.21 & 35.6\% \\
    $r{=}16$ & 69.04 & 0.14 & 18.3\% \\
    $r{=}8$  & 67.51 & 0.24 & 9.6\% \\
    $r{=}4$  & 65.21 & 0.40 & 5.3\% \\
    $r{=}2$  & 61.96 & 0.53 & 3.1\% \\
    $r{=}1$  & 57.48 & 1.40 & 2.0\% \\
    \bottomrule
  \end{tabular}
\end{table}

\section{Vision Transformer (Flowers-102)}
\label{app:vit}

All prior benchmarks in this paper operate on sequential or graph-structured
domains.  To assess whether the reservoir-control framework extends to dense
spatial representations, we apply LottaLoRA to a Vision Transformer
(ViT-Base/16)~\cite{dosovitskiy2021image} on Oxford
Flowers-102~\cite{nilsback2008flowers}, a 102-class fine-grained image
classification dataset (2{,}040 training images after merging the official
train and validation splits, 6{,}149 test images).

\paragraph{Setup.}
The backbone is initialized from an ImageNet-1k pre-trained
ViT-B/16 checkpoint.  In the LottaLoRA condition, the six projection matrices
per transformer block ($W_Q, W_K, W_V, W_O$ and the two MLP projections) are
replaced with frozen random noise matrices $\Wseed \sim \mathcal{N}(0, 1/d)$
plus trainable low-rank adapters $AB$ of rank $r$.  The patch embedding,
positional embedding, layer norms, and classification head remain trainable
in all conditions.  The baseline is standard full fine-tuning of all 85.9\,M
parameters.  Both conditions share the same training protocol: AdamW (learning rate
$2 \times 10^{-4}$, weight decay 0.05, batch size 64, gradient
accumulation 2 steps) with cosine learning rate schedule (10\% warmup),
label smoothing 0.1, and strong augmentation (RandAugment: 2 ops,
magnitude 9; Mixup $\alpha{=}0.2$, probability 0.8; CutMix
$\alpha{=}1.0$, probability 0.5; random erasing probability 0.25).
Evaluation uses EMA (decay 0.9997) and horizontal-flip TTA. Training runs
for 300 epochs with early stopping (patience 12). Pre-trained weights are
from torchvision (IMAGENET1K\_V1).

\paragraph{Results.}
Table~\ref{tab:vit} reports Top-1 accuracy (5 seeds per condition).  LottaLoRA
with $r{=}1$ (1.19\% trainable parameters) already reaches 94.53\%, recovering
\textbf{97.6\%} of the full fine-tuning baseline accuracy.  At $r{=}64$
(11.89\%), the gap narrows to 1.9\,pp.  All ranks from $r{=}1$ to $r{=}64$
fall within a 0.9\,pp band, confirming extreme rank insensitivity.

\begin{table}[ht]
  \caption{ViT-Base/16 on Flowers-102 (5 seeds, 300 epochs, pre-trained
    ImageNet-1k init).  LottaLoRA replaces attention and MLP projections
    with frozen random matrices plus low-rank adapters.  The random
    backbone condition uses a randomly initialized (non-pre-trained) ViT.}
  \label{tab:vit}
  \centering
  \small
  \begin{tabular}{lcccc}
    \toprule
    Method & Rank & Mean Top-1 & Std & Trainable Ratio \\
    \midrule
    Baseline (full fine-tune)  & ---  & 96.83\% & 0.08\% & 100\% \\
    \midrule
    LottaLoRA (pre-trained)      & 1    & 94.53\% & 0.35\% & 1.19\% \\
    LottaLoRA (pre-trained)      & 4    & 93.99\% & 0.28\% & 1.76\% \\
    LottaLoRA (pre-trained)      & 8    & 94.01\% & 0.35\% & 2.51\% \\
    LottaLoRA (pre-trained)      & 32   & 94.68\% & 0.22\% & 6.76\% \\
    LottaLoRA (pre-trained)      & 64   & 94.89\% & 0.16\% & 11.89\% \\
    \midrule
    LottaLoRA (random backbone) & 16   & 54.27\% & 0.72\% & 3.97\% \\
    LottaLoRA (random bk + LN)  & 16   & 55.56\% & 0.73\% & 4.15\% \\
    \bottomrule
  \end{tabular}
\end{table}

\begin{figure*}[ht]
  \centering
  \includegraphics[width=\linewidth]{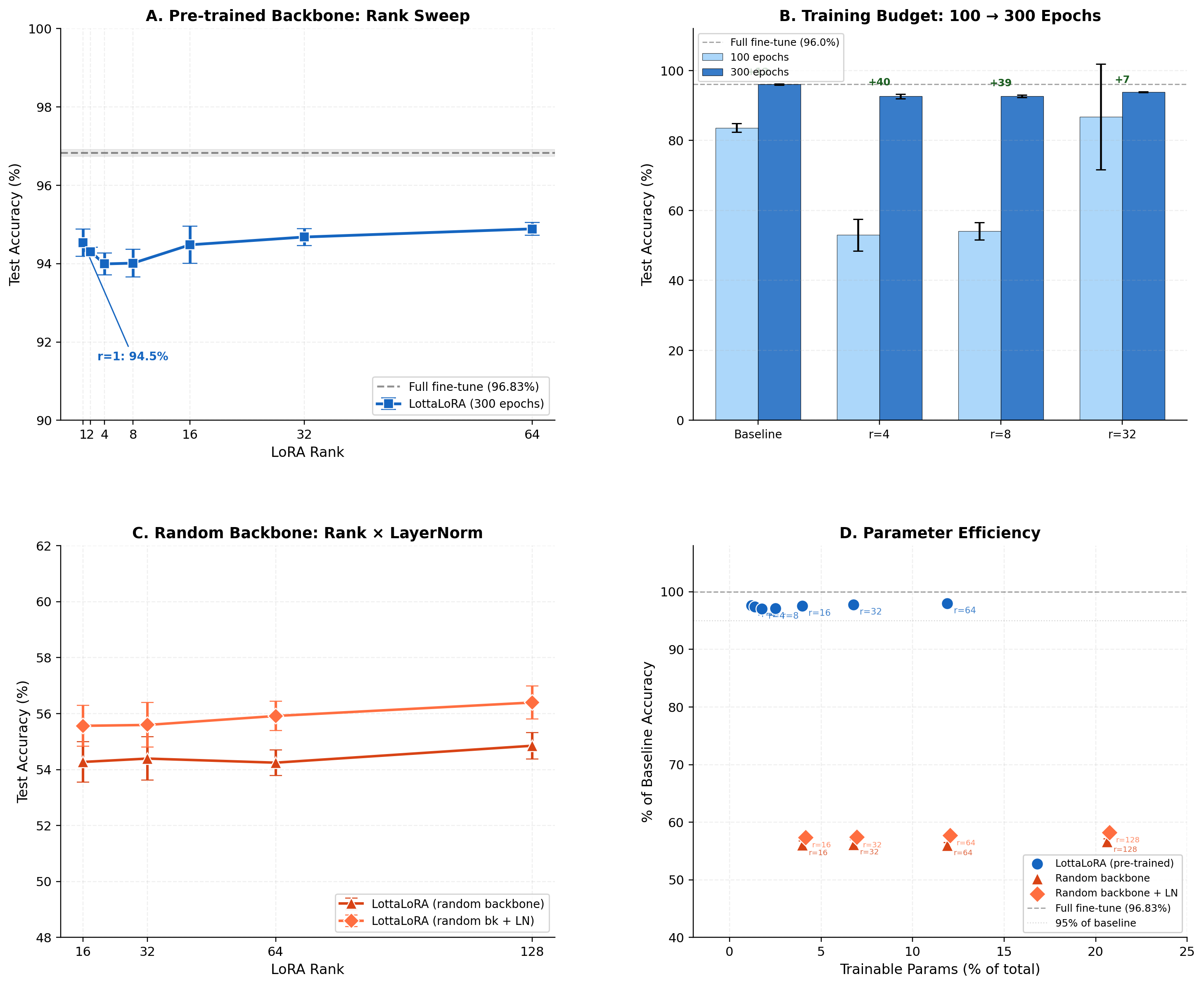}
  \caption{\textbf{Learned reservoir quality accounts for a ${\sim}$40\,pp advantage over random backbones.}
    \textbf{(A)}~Pre-trained backbone rank sweep: under extended training
    (300 epochs), LottaLoRA reaches 94--95\% across $r{=}1$--$64$, within
    1.9--2.8\,pp of full fine-tuning (dashed).  Even $r{=}1$ (1.19\%
    trainable) achieves 94.53\%.
    \textbf{(B)}~Training budget effect: extending from 100 to 300 epochs
    yields +39--40\,pp for LottaLoRA ranks and +12\,pp for the baseline.
    \textbf{(C)}~Random backbone with and without composed-path LayerNorm:
    LayerNorm adds +1.0--1.7\,pp across all ranks, but the random backbone
    ceiling remains ${\sim}$56\%, showing that learned reservoir
    quality accounts for a ${\sim}$40\,pp advantage.
    \textbf{(D)}~Parameter efficiency: at $r{=}1$ (1.19\% trainable
    parameters), LottaLoRA recovers 97.6\% of baseline accuracy; random
    backbone conditions cluster near 57\%, quantifying the information
    contributed by the frozen learned projection layers.}
  \label{fig:vit_rank}
\end{figure*}

\paragraph{Rank sensitivity.}
Accuracy is nearly flat across ranks: $r{=}1$ (94.53\%) to $r{=}64$
(94.89\%) spans only 0.9\,pp despite a $10\times$ increase in trainable
parameters (Figure~\ref{fig:vit_rank}A).  This contrasts with the
monotonic rank--accuracy relationship observed on MNIST and NCP tasks,
suggesting that on Flowers-102 the learned backbone already provides
features rich enough that even a single-rank adapter suffices.

\paragraph{Reservoir quality decomposition.}
Replacing the learned backbone with a randomly initialized one (holding
all other settings constant) drops accuracy from 94.01\% to 54.27\%---a
${\sim}$40\,pp gap (Figure~\ref{fig:vit_rank}C).  Adding composed-path
LayerNorm to the random backbone recovers only $+$1.3\,pp (55.56\%).  This
decomposition directly quantifies the \emph{reservoir quality} hypothesis:
the learned attention weights constitute a high-quality reservoir whose
structured representations are exploited by the low-rank controller, whereas a
random reservoir lacks the spatial inductive biases needed for fine-grained
visual classification.

\paragraph{Relation to published parameter-efficient methods.}
These results are not state of the art on Flowers-102.  Standard
LoRA~\cite{hu2022lora} achieves 96--98\% and Visual Prompt Tuning
(VPT)~\cite{jia2022vpt} exceeds 98\%, both by \emph{preserving} the full
pre-trained weight manifold and adding small learned corrections.
Importantly, those pre-trained backbones themselves required thousands of GPU
hours on ImageNet-scale data---resources that embody the very ``intelligence''
our experiment aims to localize.  Our goal is not to compete with methods
that inherit this investment intact, but to answer a different question:
\emph{how much of the learned reservoir structure is actually needed, and where
does it reside?}

LottaLoRA \emph{replaces} six dense projection matrices per block with
uncorrelated Gaussian noise, retaining only the patch embedding, positional
encoding, layer norms, and classification head.  Because these projections
are fully connected layers---precisely the layer type where LottaLoRA
excels across all benchmarks in this paper---the adapter can reconstruct
most of the functional pathway from scratch (Figure~\ref{fig:vit_rank}D).
The 2.8\,pp gap at $r{=}4$
quantifies the cost of discarding the learned attention structure: it is
the residual information that the low-rank controller cannot recover from a
noisy medium.
The dominant factor is therefore not the adapter but the \emph{reservoir}:
the structured spatial dynamics of a learned reservoir---analogous to the
neuroscience observation that cortical ``smart areas'' (e.g., visual
cortex) require far more developmental resources than relay nuclei,
even though the total information flow through both is comparable.

Across the full experimental arc, reservoir quality requirements form a
task-dependent spectrum.  On low-dimensional tasks (Lorenz: $R^2 = 0.9999$
at rank~1; MNIST: 96\% at rank~8), random reservoirs already provide
sufficient mixing and a learned reservoir is unnecessary.  On
high-dimensional tasks such as fine-grained 102-class visual classification
from 2{,}040 training images, the reservoir must encode spatial inductive
biases that random projections cannot supply.  This spectrum is itself a
contribution: it provides a principled framework for predicting \emph{when}
random initialization will suffice and when a learned reservoir remains
essential---a question that existing parameter-efficient methods
do not address because they never discard the pre-trained weights.

\section{IMDB Sentiment Classification}
\label{app:imdb}

All prior benchmarks in this paper operate on vision, time-series, graph, or
reinforcement learning domains.  IMDB sentiment classification extends
LottaLoRA to \emph{natural language processing}: binary classification of
movie reviews using a pre-trained
DistilBERT~\cite{sanh2019distilbert} backbone on the IMDB
dataset~\cite{maas2011learning} (25{,}000 training reviews,
25{,}000 test reviews).

\paragraph{Setup.}
The model is a 3-layer Transformer decoder (embedding dimension 256, 4 heads,
FFN multiplier~4) initialized from a DistilBERT tokenizer
(\texttt{bert-base-uncased}, max length 512).  In the LottaLoRA condition, all
attention projection matrices are replaced with fan-in-scaled frozen random
matrices $\Wseed \sim \mathcal{N}(0, 1/d_\mathrm{in})$ plus trainable low-rank
adapters of rank $r \in \{1, 2, 4, 8, 16, 32\}$.  LoRA is applied to
attention projections only (not FFN).  The classification head and all
bias/normalization parameters remain trainable.  The baseline is full
fine-tuning of all 10.3\,M parameters.  Both conditions share the same
training protocol: AdamW (learning rate $5 \times 10^{-4}$, weight decay 0.01,
batch size 32, gradient clipping 1.0) with 10\% linear warmup, LoRA dropout
0.05, model dropout 0.1, and early stopping (patience~7, $\delta = 10^{-4}$)
over a maximum of 30 epochs.  $\alpha = 2r$ throughout.  Each configuration is
evaluated over 4 seeds (42--45).  All runs executed on CPU (16 cores,
${\sim}$10 hours per run).

\begin{figure}[ht]
  \centering
  \includegraphics[width=0.95\linewidth]{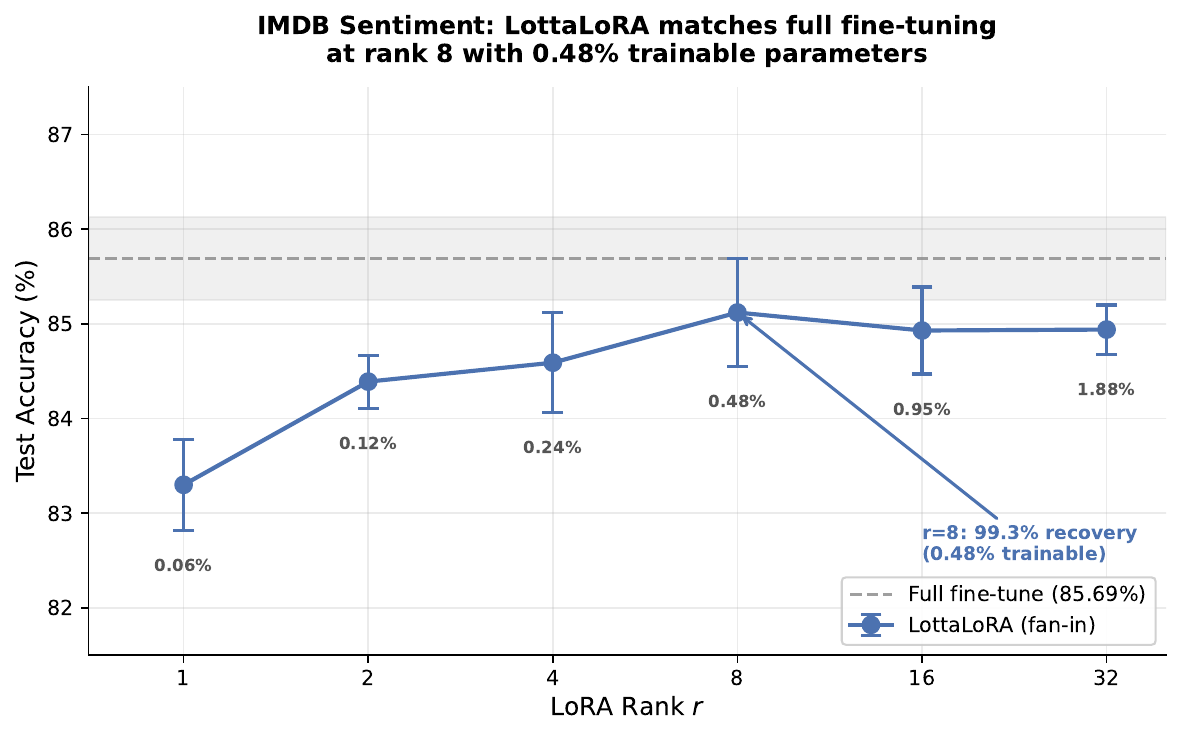}
  \caption{\textbf{LottaLoRA matches full fine-tuning on IMDB sentiment at
    rank~8 with 0.48\% trainable parameters.}  Error bars show $\pm 1$
    standard deviation over 4 seeds.  Dashed line and shaded band indicate
    the full fine-tuning baseline ($85.69 \pm 0.44$\%).  Performance
    saturates at $r{=}8$, with higher ranks providing no additional gain.}
  \label{fig:imdb_rank}
\end{figure}

\paragraph{Results.}
Table~\ref{tab:imdb_full} reports test accuracy across all ranks
(Figure~\ref{fig:imdb_rank}).  LottaLoRA at $r{=}8$ achieves
$85.12 \pm 0.57$\%, recovering \textbf{99.3\%} of the full fine-tuning
baseline ($85.69 \pm 0.44$\%) while training only \textbf{0.48\%} of the
parameters (49{,}681 out of 10.36\,M).  Performance saturates near $r{=}8$;
higher ranks ($r{=}16$, $r{=}32$) yield no additional gain, consistent with
the minimum-rank hypothesis (Section~\ref{sec:hypothesis}).

\paragraph{Baseline context.}
Our full fine-tuning baseline ($85.69$\%) falls several percentage points
below published DistilBERT accuracy on IMDB (${\sim}87$--$88$\%), likely
because our 3-layer architecture is smaller than the full 6-layer DistilBERT
and our training budget is constrained.  The comparison is therefore
protocol-fair: under matched architecture and training conditions, LottaLoRA
at $r{=}8$ essentially matches full fine-tuning at 0.48\% of the parameters.

\begin{table}[ht]
  \caption{IMDB sentiment classification: test accuracy (\%) by LoRA rank
    (4 seeds, fan-in-scaled $\Wseed$, DistilBERT backbone).  LottaLoRA at
    $r{=}8$ recovers 99.3\% of full fine-tuning with 0.48\% trainable
    parameters.}
  \label{tab:imdb_full}
  \centering
  \small
  \begin{tabular}{lccccc}
    \toprule
    Method & Rank & Mean Acc & Std & Trainable & \% Trainable \\
    \midrule
    Baseline (full)       & --- & 85.69\% & 0.44\% & 10{,}315{,}030 & 100\% \\
    \midrule
    LottaLoRA (fan-in)    &  1  & 83.30\% & 0.48\% &  6{,}673       & 0.06\% \\
    LottaLoRA (fan-in)    &  2  & 84.39\% & 0.28\% & 12{,}817       & 0.12\% \\
    LottaLoRA (fan-in)    &  4  & 84.59\% & 0.53\% & 25{,}105       & 0.24\% \\
    LottaLoRA (fan-in)    &  8  & 85.12\% & 0.57\% & 49{,}681       & 0.48\% \\
    LottaLoRA (fan-in)    & 16  & 84.93\% & 0.46\% & 98{,}833       & 0.95\% \\
    LottaLoRA (fan-in)    & 32  & 84.94\% & 0.26\% & 197{,}137      & 1.88\% \\
    \bottomrule
  \end{tabular}
\end{table}

\section{CIFAR-10 Plain-20 CNN}
\label{app:cifar10}

\paragraph{Setup.}
We apply LottaLoRA to a Plain-20 CNN~\cite{he2016deep} on CIFAR-10
(50{,}000 training, 10{,}000 test images, 10 classes). The backbone consists of
20 convolutional layers organized in three stages ($16, 32, 64$ channels)
with a final average-pooling and linear classifier (269{,}722 total parameters).
In the LottaLoRA condition, all convolutional weight matrices are frozen and
replaced with $W_\mathrm{eff} = \beta\,\Wseed + BA$; $\Wseed$ is initialized
with fan-in-scaled Gaussian noise. Training uses AdamW for 164 epochs with
cosine learning rate schedule. Results are averaged over 3 seeds (42, 123, 7).

\paragraph{Results.}
Table~\ref{tab:cifar10} reports test accuracy across LoRA ranks and
configurations. At rank~8, vanilla LottaLoRA (no DoRA, with residual
connections) achieves 86.20\%, recovering 94.9\% of the 90.81\% baseline.
Adding DoRA~\cite{liu2024dora} improves this to 88.12\% (97.0\% recovery).
At rank~32, DoRA+residual reaches 89.66\% (98.7\% recovery) with remarkably
tight variance ($\pm 0.03\%$ across 3 seeds), the best LottaLoRA result on
this benchmark---within 1.15\,pp of the fully trained baseline.
Dropout hurts performance at all ranks, unlike MNIST where it is benign.

\begin{table}[ht]
  \caption{CIFAR-10 Plain-20 test accuracy (\%) by LoRA rank and configuration
    (3 seeds; $r\in\{1,4,8,16\}$: 150 epochs, $r\in\{2,32\}$: 164 epochs). Residual = residual connection around frozen layer.
    DoRA = weight-decomposed low-rank adaptation.
    Baseline (full training): $90.81\% \pm 0.20\%$.}
  \label{tab:cifar10}
  \centering
  \small
  \begin{tabular}{rcccc}
    \toprule
    Rank & Vanilla & +Residual & +DoRA+Res & Trainable ratio \\
    \midrule
    1  & $61.08 \pm 1.42$ & $61.57 \pm 1.67$ & $76.70 \pm 0.54$ &  3.1\% \\
    2  & $75.02 \pm 0.39$ & $76.31 \pm 0.83$ & $82.24 \pm 0.50$ &  5.2\% \\
    4  & $82.40 \pm 0.09$ & $82.59 \pm 0.11$ & $85.43 \pm 0.08$ & 10.1\% \\
    8  & $86.18 \pm 0.41$ & $86.20 \pm 0.25$ & $88.12 \pm 0.31$ & 19.5\% \\
    16 & $87.70 \pm 0.26$ & $88.76 \pm 0.09$ & $89.39 \pm 0.17$ & 38.3\% \\
    32 & $88.01 \pm 0.34$ & $88.48 \pm 0.38$ & $89.66 \pm 0.03$ & 43.3\% \\
    \bottomrule
  \end{tabular}
\end{table}

\paragraph{DoRA effect.}
DoRA provides a dramatic improvement at rank~1 ($+$15.6\,pp over vanilla),
where the direction normalization compensates for the extremely limited
capacity of a single-rank adapter. The effect generally diminishes at
higher ranks: $+$7.2\,pp at rank~2, $+$3.0\,pp at rank~4, $+$1.9\,pp at
rank~8, $+$0.6\,pp at rank~16, and $+$1.7\,pp at rank~32.
Residual connections alone provide negligible benefit ($<$1\,pp at all ranks),
suggesting that on this architecture the adapter path, not the skip
connection, is the primary mechanism.

\section{Decision Transformer (HalfCheetah-Expert)}
\label{app:dt}

\paragraph{Setup.}
We apply LottaLoRA to a Decision Transformer~\cite{chen2021decision}
trained on the HalfCheetah-expert-v2 offline RL dataset at two scales:
\textbf{small} (embedding dimension 128, 3 layers, 1 attention head, MLP
ratio 4$\times$) and \textbf{large} (embedding dimension 1024, 12 layers,
8 heads, MLP ratio 4$\times$). Context length is $K{=}20$ timesteps
(state--action--reward triples, yielding 60 tokens per sequence).
Return-to-go values are scaled by 6{,}000. Training uses AdamW (learning
rate $10^{-4}$, weight decay $10^{-4}$, batch size 64, 120 epochs) with
5\% linear warmup and early stopping (patience 15). In the LottaLoRA
condition, all attention projections are frozen and replaced with LoRA
adapters ($\alpha{=}r$, dropout 0.05); $\Wseed$ uses fan-in-scaled
Gaussian initialization. Results are averaged over 5 seeds.

\paragraph{Extended rank sweep (small scale).}
We extend the small-scale analysis with a rank sweep
$r \in \{1, 2, 4, 8, 16, 32\}$ under two scaffold conditions:
\emph{static} (frozen $\Wseed$, denoted \texttt{noise\_lora}) and
\emph{resampled} ($\Wseed$ redrawn from its initialization distribution at
the start of each epoch, denoted \texttt{meta\_lora}). All other
hyperparameters match the static condition. Online evaluation uses
D4RL-normalized returns from HalfCheetah-v4 rollouts (1{,}000 timesteps,
10 episodes per seed).
The large-scale result ($r{=}8$, 1024-d / 12-layer) is from the original
sweep and does not include online evaluation.

\begin{table}[ht]
  \caption{Decision Transformer on HalfCheetah-expert-v2 (5 seeds).
    Val MSE is offline prediction error (lower is better);
    D4RL~(\%) is normalized online return from HalfCheetah-v4 rollouts
    (higher is better; 10 episodes $\times$ 1{,}000 timesteps per seed).
    \emph{Static} = frozen $\Wseed$; \emph{Resampled} = $\Wseed$ redrawn
    each epoch.
    Large-scale baseline and LottaLoRA from the original sweep ($r{=}8$ only).}
  \label{tab:dt}
  \centering
  \footnotesize
  \begin{tabular}{lrcccr}
    \toprule
    Method & $r$ & Val MSE & Std & D4RL (\%) & Ratio \\
    \midrule
    \multicolumn{6}{l}{\textit{Small scale (128-d, 3-layer)}} \\
    Baseline (full)     & --- & 0.02718 & 0.00173 & $91.86 \pm 0.55$ & 100.00\% \\
    \midrule
    \multicolumn{6}{l}{\textit{Static scaffold (\texttt{noise\_lora})}} \\
    LottaLoRA &  1 & 0.03482 & 0.00169 & $61.10 \pm 21.46$ & 19.23\% \\
    LottaLoRA &  2 & 0.03221 & 0.00169 & $84.29 \pm 5.74$  & 19.98\% \\
    LottaLoRA &  4 & 0.02985 & 0.00173 & $89.13 \pm 3.29$  & 21.44\% \\
    LottaLoRA &  8 & 0.02832 & 0.00179 & $91.75 \pm 2.35$  & 24.22\% \\
    LottaLoRA & 16 & 0.02761 & 0.00173 & $93.47 \pm 0.94$  & 29.22\% \\
    LottaLoRA & 32 & 0.02713 & 0.00171 & $94.33 \pm 0.83$  & 37.47\% \\
    \midrule
    \multicolumn{6}{l}{\textit{Resampled scaffold (\texttt{meta\_lora})}} \\
    LottaLoRA &  1 & 0.05116 & 0.00338 & $4.38 \pm 0.59$   & 19.23\% \\
    LottaLoRA &  2 & 0.04421 & 0.00248 & $13.26 \pm 11.26$ & 19.98\% \\
    LottaLoRA &  4 & 0.03580 & 0.00166 & $51.72 \pm 17.18$ & 21.44\% \\
    LottaLoRA &  8 & 0.02974 & 0.00180 & $91.00 \pm 1.23$  & 24.22\% \\
    LottaLoRA & 16 & 0.02804 & 0.00183 & $91.75 \pm 1.65$  & 29.22\% \\
    LottaLoRA & 32 & 0.02731 & 0.00179 & $93.34 \pm 2.68$  & 37.47\% \\
    \midrule
    \multicolumn{6}{l}{\textit{Large scale (1024-d, 12-layer)}} \\
    Baseline (full)     & --- & 0.02812 & 0.00188 & ---               &  99.97\% \\
    LottaLoRA (static)  &   8 & 0.02847 & 0.00149 & ---               &   1.87\% \\
    \bottomrule
  \end{tabular}
\end{table}

\begin{figure}[ht]
  \centering
  \includegraphics[width=0.90\linewidth]{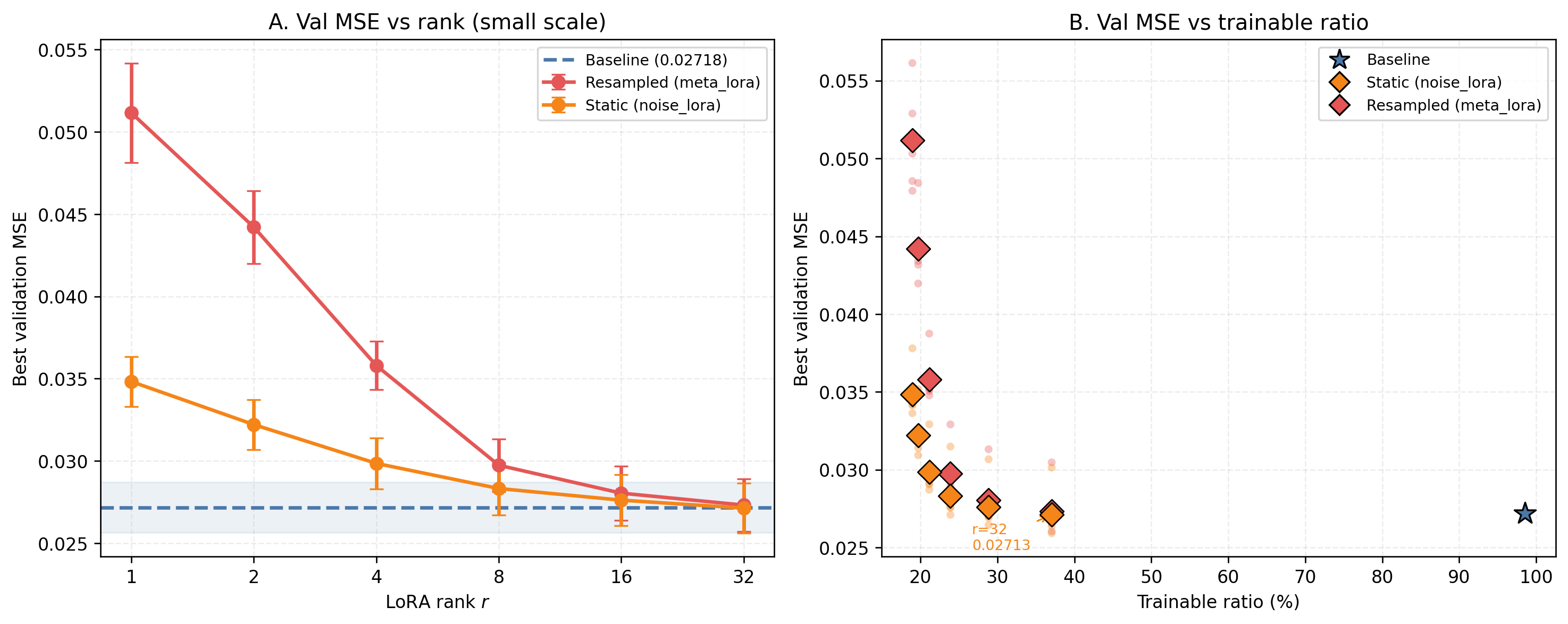}
  \caption{\textbf{LottaLoRA matches the fully trained Decision Transformer at sufficient rank.}
    Left: validation MSE by method and rank at small scale (128-d, 3-layer).
    The static scaffold (\texttt{noise\_lora}) closes the gap monotonically
    with rank; at $r{=}32$ the difference is not statistically significant.
    Right: val MSE vs trainable parameter ratio.  The large-scale $r{=}8$
    result (1.87\% trainable) trails the baseline by 1.2\% relative MSE;
    the small-scale $r{=}32$ result (37.47\% trainable) matches the baseline.}
  \label{fig:dt}
\end{figure}

\paragraph{Rank progression at small scale.}
Under the static scaffold, validation MSE decreases monotonically with rank
from 0.03482 ($r{=}1$) to 0.02713 ($r{=}32$).  At $r{=}32$ the paired
mean difference versus the baseline is $-$0.00004 (0.15\% relative), which
is not statistically significant---LottaLoRA matches the fully trained
baseline while training 37.47\% of the parameters
(Table~\ref{tab:dt}, Figure~\ref{fig:dt}).
On the online D4RL evaluation, static-scaffold LottaLoRA at $r \geq 16$
exceeds the baseline score (93.47\% vs 91.86\% at $r{=}16$; 94.33\% vs
91.86\% at $r{=}32$), suggesting that the frozen scaffold may provide a
regularization benefit during offline-to-online transfer.

\paragraph{Resampling collapse confirms Section~\ref{sec:q3}.}
The \texttt{meta\_lora} condition replicates the resampling collapse
observed on MNIST (Table~\ref{tab:meta_ablation}).
At $r{=}1$, resampling drops D4RL score from 61.1\% (static) to 4.4\%
(resampled)---a near-complete collapse.
Recovery follows the same rank-dependent pattern: by $r{=}8$,
\texttt{meta\_lora} recovers to D4RL 91.0\% (vs 91.75\% static), and by
$r{=}32$ the gap narrows to 1.0\,pp (93.34\% vs 94.33\%).
This confirms across a second architecture and domain that scaffold
stability is critical at low rank but becomes less important as adapter
capacity increases, consistent with the compression interpretation in
Section~\ref{sec:q3}.
The learned $\beta$ scalars confirm this mechanistically: under the static
scaffold $\beta$ averages ${\approx}0.48$ across layers---strictly positive
in all seeds, though lower than the ${\approx}0.91$--$0.99$ range seen in
Transformers and ViT (Section~\ref{sec:q3})---while under resampled
scaffolds $\beta$ collapses to $|{\beta}| < 0.001$, confirming backbone
silencing across this second architecture and domain.

\paragraph{Large-scale result.}
At large scale (1024-d / 12-layer), LottaLoRA ($r{=}8$) trails the matched
baseline by $+$0.00035 MSE (1.2\% relative) while training only 1.87\% of
parameters---a $53\times$ reduction.  A paired $t$-test across 5 seeds
yields $t(4) = 4.29$, $p = 0.013$ (Cohen's $d = 1.92$): the gap is
statistically significant but small, representing the extreme-compression
regime where a tiny adapter steers a large frozen backbone.

\section{Engineering: LayerNorm, Fan-In Scaling, $\beta$ Distribution}
\label{app:engineering}

\paragraph{LayerNorm on combined path.}
A controlled 2$\times$2 experiment (16 points, fan-in vs explicit scaling
$\times$ LayerNorm on/off) shows that applying LayerNorm to $\Wseed x + \Delta W x$
before the nonlinearity adds $+$1.01\,pp over fan-in scaling alone
(95.0\% vs 94.0\%).
Enabling LayerNorm introduces a trainable gain $\gamma \in \mathbb{R}^{d}$
and bias $b \in \mathbb{R}^{d}$ per normalized layer, adding $2d$ trainable
parameters. This overhead is negligible relative to the LoRA factors ($2rd$
per layer) at any rank $r \geq 2$, but is included in all reported trainable
counts for experiments that enable it. Configurations without LayerNorm
(e.g., the \textbf{no-LN} ablations from prior NCP experiments)
omit these parameters entirely.

\paragraph{Fan-in $\Wseed$ scaling.}
Setting $\sigma_{\Wseed} = 1/\sqrt{d_\mathrm{in}}$ is necessary for stable
training in deep/wide networks. Without it, activation variance compounds
geometrically across layers, causing loss explosion (observed on deep GNN
training at $L = 8$, $d = 1024$: initial loss $\approx 1.7 \times 10^{11}$). Fan-in
scaling resolves the instability entirely and is the confirmed default for
all experiments.

\begin{figure}[ht]
  \centering
  \begin{minipage}[t]{0.38\linewidth}
    \centering
    \includegraphics[width=\linewidth]{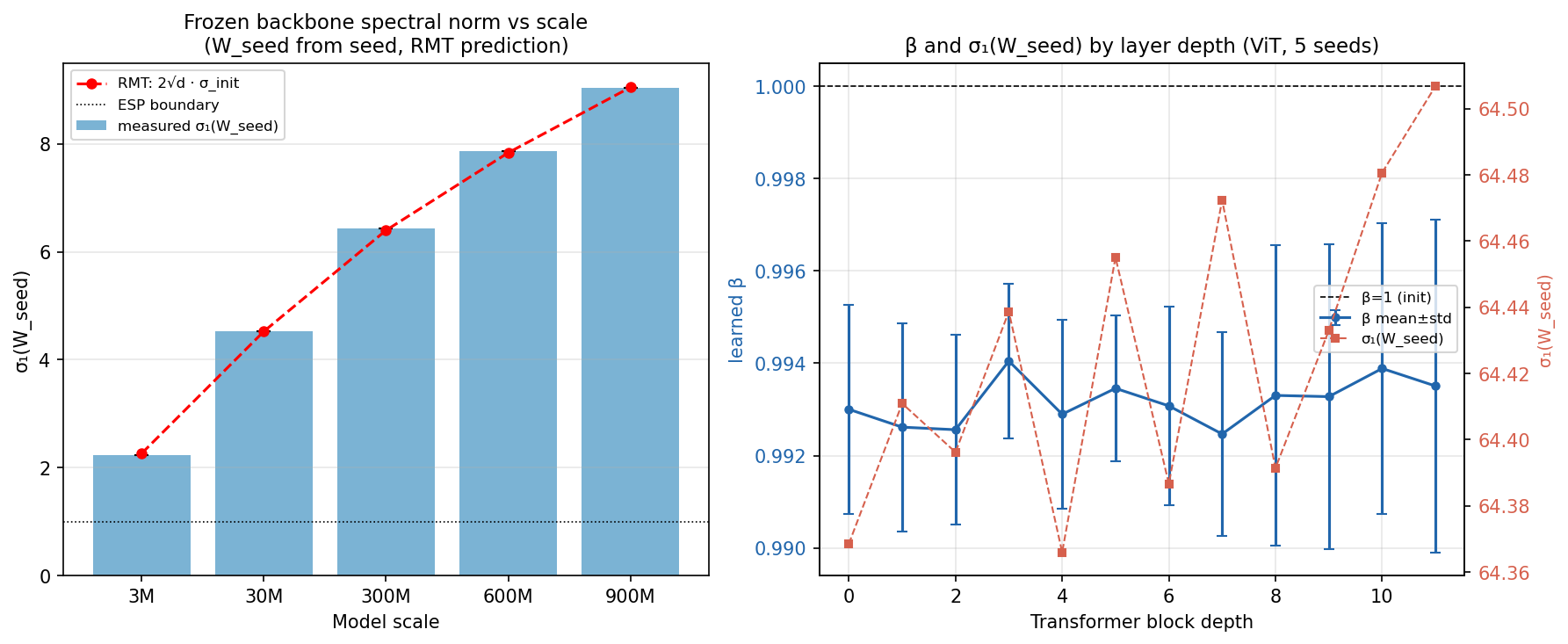}
  \end{minipage}\hfill
  \begin{minipage}[t]{0.58\linewidth}
    \centering
    \includegraphics[width=\linewidth]{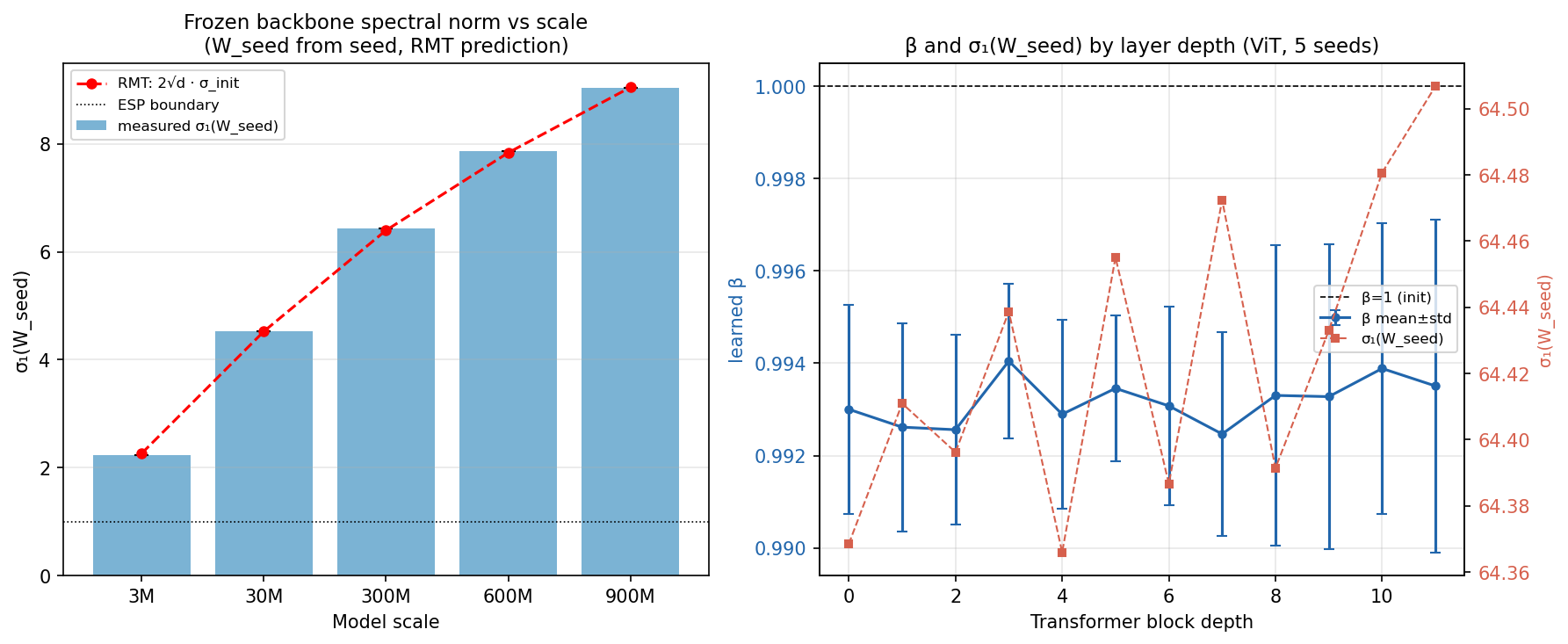}
  \end{minipage}
  \caption{\textbf{The frozen backbone violates the Echo State Property at every layer, yet $\beta$ remains strictly positive (near 1 in ViT, median ${\approx}0.99$; see Section~\ref{sec:q3} for architecture-dependent values).}
    \textbf{Left:} Empirical distribution of learned $\beta$ values from all
    Transformer checkpoints (1,632 values). Mass concentrates below 1.0;
    all values remain positive, confirming the backbone always contributes.
    \textbf{Center:} Spectral norm $\sigma_1(\Wseed)$ by backbone scale
    (bars, reconstructed from seed) versus the Random Matrix Theory
    prediction $2\sqrt{d}\,\sigma_\mathrm{init}$ (red dashed). All scales
    violate the ESP boundary (dotted line at 1).
    \textbf{Right:} Learned $\beta$ (blue, left axis) and
    $\sigma_1(\Wseed)$ (red, right axis) by transformer block depth in ViT
    (5 seeds). $\beta$ is flat across depth and near 1 despite
    $\sigma_1(\Wseed) \approx 60$--$80$, indicating that $\beta$ does not
    compensate for spectral structure---the adapter matrices do.}
  \label{fig:beta_combined}
\end{figure}

\paragraph{$\beta$ distribution.}
As discussed in Section~\ref{sec:method}, $\beta$ is a secondary
amplitude control whose primary role is to modulate the overall backbone
gain; the adapter matrices $A$ and $B$ channel the scaffold's representational capacity toward the task.
Across 1,632 learned $\beta$ values from Transformer experiments (Figure~\ref{fig:beta_combined}, left): mean
0.894, median 0.915, IQR $[0.837, 0.946]$.
Across 360 ViT layer pairs the median is 0.993 (IQR $[0.992, 0.995]$).
In both architectures every $\beta$ value remains strictly positive:
the optimizer consistently attenuates but never zeros or inverts the
frozen backbone contribution.

\paragraph{$\beta$ by layer depth and spectral norm (ViT).}
Figure~\ref{fig:beta_combined} (center and right) shows $\sigma_1(\Wseed)$ across all five backbone scales
and $\beta$ by transformer block depth for ViT (5 seeds, 72 layer pairs per seed).
Two findings are notable.
First, $\sigma_1(\Wseed) \gg 1$ at every layer and every scale---the frozen
backbone violates the Echo State Property by a wide margin (100\% of ViT
layers), yet the system trains stably because the
adapter compensates.
Second, $\beta$ shows no systematic depth trend and remains near 0.993
regardless of layer depth or $\sigma_1(\Wseed)$, indicating that $\beta$ is
not the compensation mechanism: the adapter matrices handle spectral
correction while $\beta$ provides only a mild global attenuation.

\section{General LottaLoRA Algorithm}
\label{app:algorithm}

Algorithm~\ref{alg:lottalora} provides a framework-agnostic pseudocode
description of the LottaLoRA training procedure applicable to any
architecture with linear layers.

\begin{figure*}[ht]
  \centering
  \includegraphics[width=\linewidth]{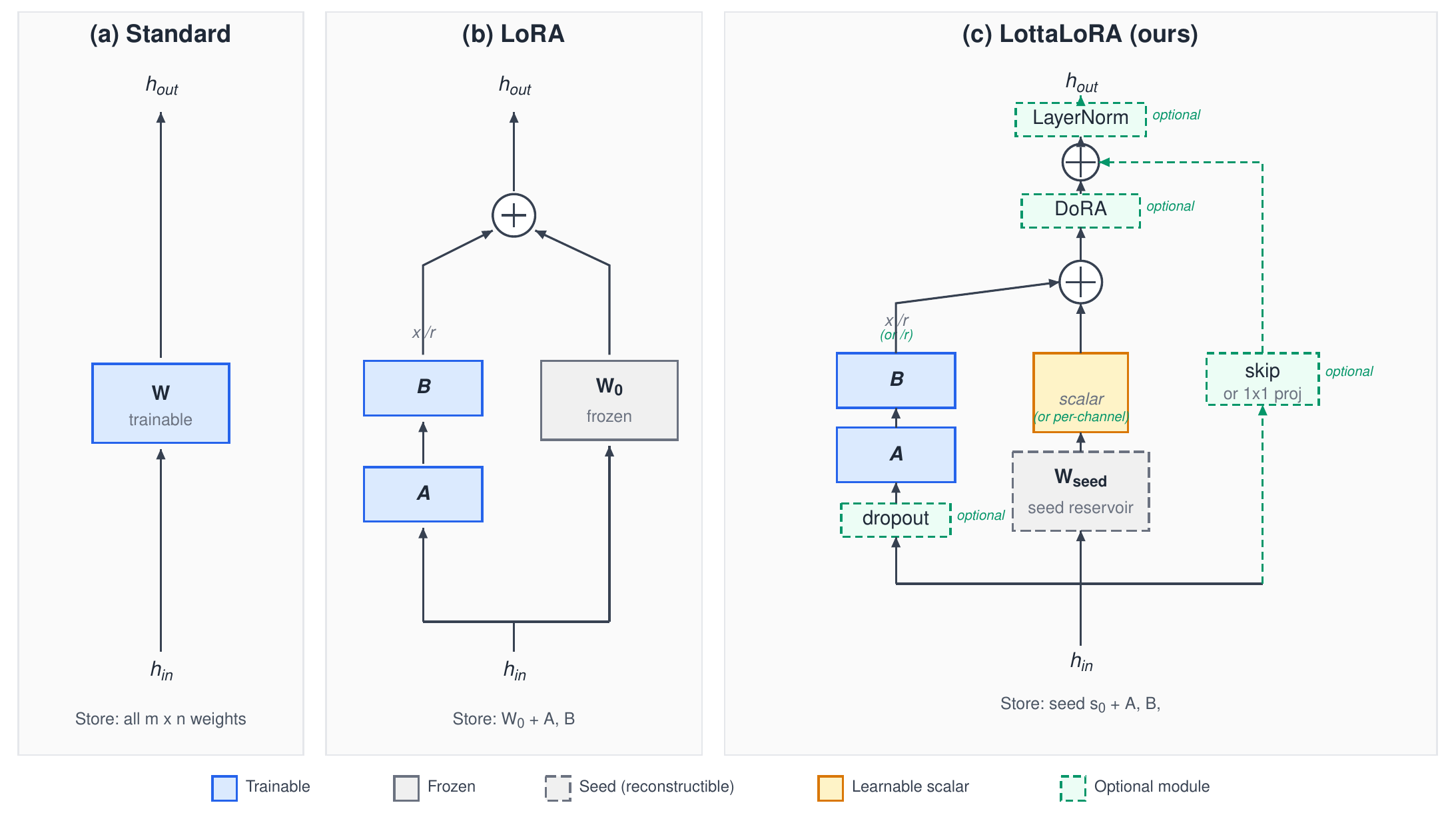}
  \caption{\textbf{LottaLoRA replaces the frozen pre-trained backbone with a seed-reconstructible random scaffold.}
    \textbf{(a)}~Standard training: all weights in~$W$ are trainable and must be stored.
    \textbf{(b)}~LoRA: a frozen pre-trained backbone~$W_0$ is augmented with
    trainable low-rank adapters~$A$ and~$B$; both~$W_0$ and the adapters must
    be stored.
    \textbf{(c)}~LottaLoRA (ours): the backbone~$\Wseed$ is generated from a
    random seed and frozen; a learnable scalar~$\beta$ modulates its amplitude.
    The adapter path (trainable~$A$,~$B$ with $\alpha/r$ or rsLoRA scaling,
    optional dropout) and backbone path merge at a summation node;
    optional DoRA renormalises the combined output, and a residual skip
    connection may be added thereafter.
    An optional LayerNorm follows before the final output.
    Only the seed, adapter weights, and~$\beta$ need to be stored or distributed.}
  \label{fig:layer_diagram}
\end{figure*}

\begin{figure}[ht]
\small
\centering
\fbox{\parbox{0.92\linewidth}{%
\textbf{Algorithm 1:} LottaLoRA Training\\[4pt]
\textbf{Input:} Architecture spec $\mathcal{A}$, random seed $s$,
LoRA rank $r$, scaling hyperparameter $\alpha$, initialization
distribution $\mathcal{D}$ (any; see Section~\ref{sec:q2})\\[2pt]
\textbf{Output:} Trained LoRA state
$\{A_i, B_i, \beta_i\}_{i=1}^{n}$, head parameters $\theta_\mathrm{head}$\\[6pt]
\textit{// Phase 1: Backbone construction}\\
1.\quad Initialize PRNG with seed $s$\\
2.\quad \textbf{for} each linear layer $i$ in $\mathcal{A}$ \textbf{do}\\
3.\quad\quad Draw $\Wseed^{(i)} \sim \mathcal{D}$; freeze
($\texttt{requires\_grad} = \texttt{False}$)\\
4.\quad\quad Initialize $A^{(i)} \in \mathbb{R}^{r \times d_\mathrm{in}}$
with Kaiming uniform\\
5.\quad\quad Initialize $B^{(i)} \in \mathbb{R}^{d_\mathrm{out} \times r}$
with zeros\\
6.\quad\quad Initialize $\beta^{(i)} = 1.0$ (trainable scalar)\\
7.\quad \textbf{end for}\\
8.\quad Initialize task-specific head $\theta_\mathrm{head}$ (embeddings,
classification layer, LayerNorm)\\[6pt]
\textit{// Phase 2: Training}\\
9.\quad $\Theta \leftarrow \{A_i, B_i, \beta_i\}_{i=1}^{n}
\cup \theta_\mathrm{head}$\\
10.\;\, Train $\Theta$ with standard optimizer; $\Wseed^{(i)}$ never updated\\[6pt]
\textit{// Forward pass per layer $i$:}\\
\quad $h_\mathrm{out} = \beta^{(i)}\, \Wseed^{(i)}\, h_\mathrm{in}
+ \frac{\alpha}{r}\, B^{(i)} A^{(i)}\, h_\mathrm{in}$\\[6pt]
\textit{// Phase 3: Distribution}\\
11.\;\, Save: seed $s$, architecture $\mathcal{A}$, distribution
$\mathcal{D}$, PRNG algorithm,
$\{A_i, B_i, \beta_i\}$, $\theta_\mathrm{head}$\\[6pt]
\textit{// Note: $\mathcal{D}$ may be any distribution---Gaussian,}\\
\textit{// binary, sparse, quantized, or spectral-radius-controlled}\\
\textit{// (Section~\ref{sec:q2}). The LoRA adapter compensates for}\\
\textit{// the specific choice; only the seed must be recorded.}
}}
\caption{The LottaLoRA training procedure.}
\label{alg:lottalora}
\end{figure}

\section{Design Choices: Required vs.\ Optional Components}
\label{app:design_choices}

Table~\ref{tab:design_choices} summarizes the architectural components
of LottaLoRA and their necessity, as established by the ablation
experiments across MNIST and Transformer benchmarks.

\begin{table}[ht]
  \caption{Required and optional design choices in LottaLoRA.}
  \label{tab:design_choices}
  \centering
  \small
  \begin{tabular}{llp{0.52\linewidth}}
    \toprule
    \textbf{Component} & \textbf{Status} & \textbf{Notes} \\
    \midrule
    Frozen $\Wseed$            & Required        & Any distribution; values interchangeable but actively exploited when static ($\beta > 0$, magnitude varies by architecture; Sections~\ref{sec:q2},~\ref{sec:q3}) \\
    LoRA factors ($A$, $B$) & Required        & Core trainable parameters \\
    $\beta$ scalar          & Recommended     & Stabilizes training; can be fixed at 1.0 \\
    $\alpha/r$ scaling      & Standard        & Follows LoRA convention~\cite{hu2022lora} \\
    LayerNorm               & Optional        & Adds $2d$ params/layer; helps at low rank \\
    Fan-in scaling          & Req.\ if no LN  & Either fan-in or LN needed; both removed $\to$ divergence \\
    $B$-init $= 0$          & Recommended     & Kaiming hurts at low rank; converges at high rank (Appendix~\ref{app:mnist_binit}) \\
    Trainable head          & Recommended     & LoRA-only head viable at sufficient rank; task-dependent (Appendix~\ref{app:mnist_output}) \\
    \bottomrule
  \end{tabular}
\end{table}

\end{document}